\documentclass[runningheads]{llncs}

\usepackage[mobile]{eccv}

\usepackage{eccvabbrv}

\usepackage{siunitx}
\usepackage{graphicx}
\usepackage{amsmath}
\usepackage{amssymb}
\usepackage{booktabs}

\usepackage[dvipsnames]{xcolor,colortbl}
\usepackage{threeparttable}
\usepackage{tabularx}
\usepackage{array,multirow}
\usepackage{threeparttable}

\usepackage{textcomp}
\usepackage{bm}
\usepackage{mathrsfs}
\usepackage{times}
\usepackage{epsfig}

\usepackage[normalem]{ulem}
\usepackage{outlines}
\usepackage{ifthen}
\usepackage{pifont}

\usepackage{pgfplots}
\pgfplotsset{compat=1.18}

\usepackage{caption}
\usepackage{setspace}

\usepackage{enumitem}
\setlist[itemize]{topsep=0.1\baselineskip}

\DeclareMathSizes{11}{8}{6}{4} %
\BeforeBeginEnvironment{align}{\setlength{\abovedisplayskip}{0.5\baselineskip}}
\AfterEndEnvironment{align}{\setlength{\belowdisplayskip}{0.5\baselineskip}}

\usepackage[accsupp]{axessibility}  %

\usepackage[pagebackref,breaklinks,colorlinks,citecolor=eccvblue]{hyperref}

\usepackage{orcidlink}

\usepackage[capitalize,noabbrev]{cleveref}

\newcolumntype{s}{>{\centering\arraybackslash}X}
\newcommand*{\ourmodel}{\textsc{DeTra}}
\newcommand*{\ua}{$\uparrow$}
\newcommand*{\da}{$\downarrow$}

\newcommand{\teaserfactor}{0.8}

\makeatletter
\renewcommand\paragraph{\@startsection{paragraph}{4}{\z@}%
                       {-10\p@ \@plus -4\p@ \@minus -4\p@}%
                       {-0.05em \@plus -0.05em \@minus -0.05em}%
                       {\normalfont\normalsize\bfseries}}
\makeatother
\newcommand{\paragraphc}[1]{\paragraph{#1}: }

\makeatletter
\def\@fnsymbol#1{\ensuremath{\ifcase#1\or \star\or \dagger\or \ddagger\or
   \mathsection\or \mathparagraph\or \|\or **\or \dagger\dagger
   \or \ddagger\ddagger \else\@ctrerr\fi}}
\makeatother

\begin{document}

\title{
DeTra: A Unified Model for Object Detection and Trajectory Forecasting
\vspace{-15pt}
}

\author{
\textbf{Sergio Casas$^\star$$^{1,2}$, Ben Agro$^\star$$^{1,2}$, Jiageng Mao\thanks{Denotes equal contribution. $^\dagger$ Work done while at Waabi.}$^\dagger$, \\ Thomas Gilles$^{1}$, Alexander Cui$^\dagger$, Thomas Li$^{1,2}$, Raquel Urtasun$^{1,2}$} \\
}

\authorrunning{S.~Casas, B.~Agro, J.~Mao, T.~Gilles, A.~Cui, T.~Li, R.~Urtasun}

\institute{\vspace{-5pt}Waabi$^{1}$, University of Toronto$^{2}$ \\ \email{\{sergio, bagro, tgilles, tli, urtasun\}@waabi.ai}}

\maketitle

\vspace{-15pt}
\begin{abstract}
    The tasks of object detection and trajectory forecasting play a crucial role in understanding the scene for autonomous driving.
    These tasks are typically executed in a cascading manner, making them prone to compounding errors. 
    Furthermore, there is usually a very thin interface between the two tasks, creating a lossy information bottleneck.
    To address these challenges, our approach formulates the union of the two tasks as a trajectory refinement problem, where the first pose is the detection (current time), and the subsequent poses are the waypoints of the multiple forecasts (future time).
    To tackle this unified task, we design a refinement transformer that infers the presence, pose, and multi-modal future behaviors of objects 
    directly from LiDAR point clouds and high-definition maps.
    We call this model \ourmodel{}, short for \textit{object \underline{De}tection} and \textit{\underline{Tra}jectory forecasting}.
    In our experiments, we observe that \ourmodel{} outperforms the state-of-the-art on Argoverse 2 Sensor and Waymo Open Dataset by a large margin, across a broad range of metrics.
    Last but not least, we perform extensive ablation studies that show the value of refinement for this task, that every proposed component contributes positively to its performance, and that key design choices were made.
    For more information, visit the project website: \href{https://waabi.ai/research/detra}{https://waabi.ai/research/detra}.
\end{abstract}

\vspace{-20pt}
\section{Introduction}

To ensure the safe deployment of autonomous driving technology, self-driving vehicles (SDVs) must be able to perceive their surroundings and anticipate potential outcomes accurately. Object detection and trajectory forecasting tasks fulfill these two capabilities in autonomy stacks. Object detection aims to recognize and localize objects in the environment, while trajectory forecasting anticipates the future behaviors of those objects in the form of trajectories. Subsequently, a planner ingests the detected objects and their future trajectories to produce a safe maneuver for the ego vehicle. 

\begin{figure}[t]
    \centering
    \includegraphics[width=\teaserfactor\linewidth]{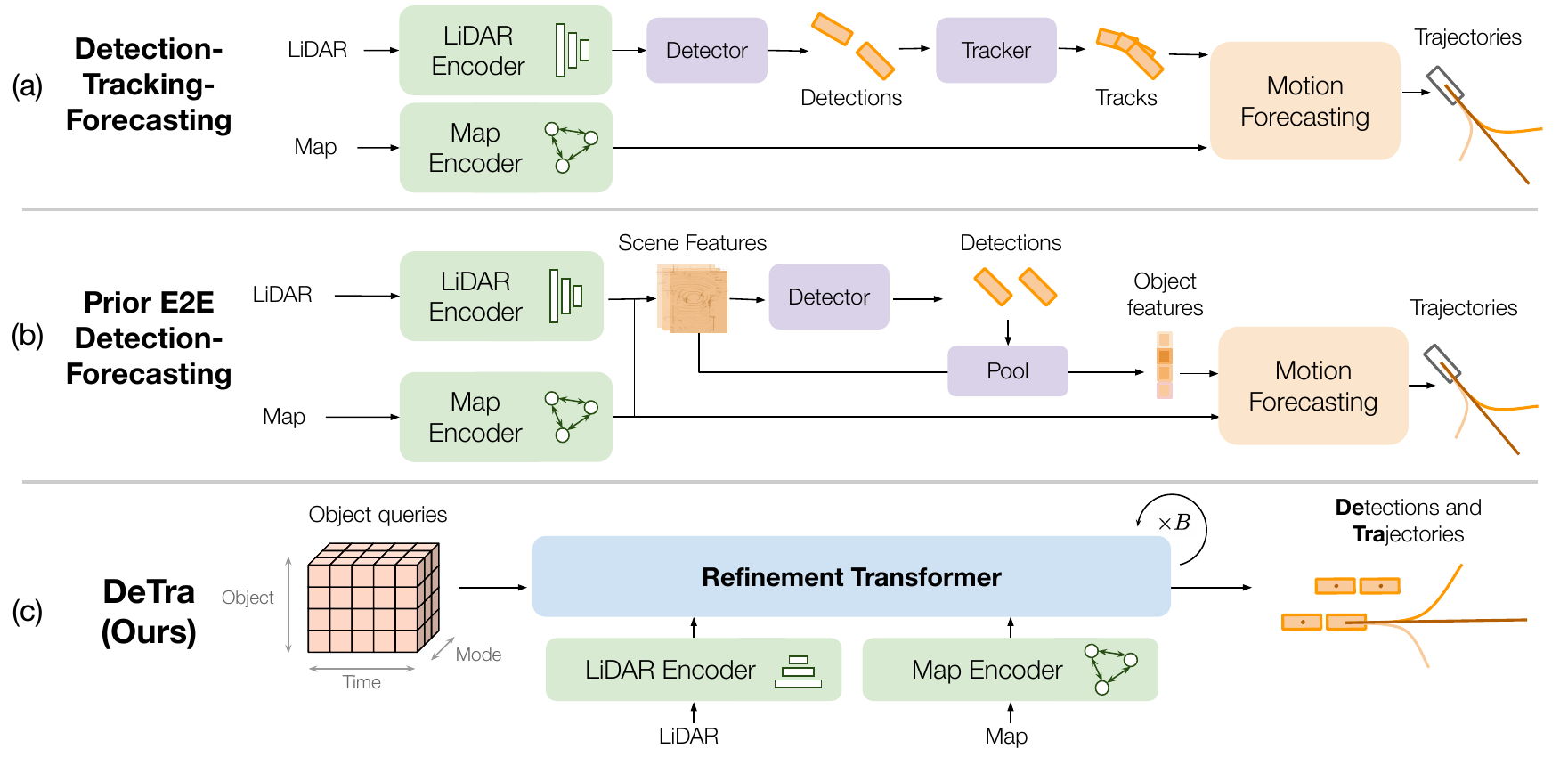}
    \vspace{-10pt}
    \caption{Modular detection-tracking-forecasting methods use narrow interfaces, hindering their performance due to compounding errors and a lossy information bottleneck. Prior end-to-end methods widen the interfaces and train jointly but still utilize a cascading structure, thus suffering from compounding errors. Our method tackles the problem as a unified trajectory refinement task.}
    \label{fig:hook}
\end{figure}

Traditional autonomous systems tackle object detection and trajectory forecasting as separate tasks, connecting them in a cascaded fashion through tracking (\cref{fig:hook}\textcolor{red}{a}). Tracking is a narrow information bottleneck, only providing downstream forecasting with noisy past trajectories describing the motion of each object. Albeit conceptually simple, this cascading decomposition suffers from compounding errors. For instance, a small estimation error in the detection heading might result in a forecasted trajectory that occupies the wrong lane, potentially causing a dangerous maneuver by the SDV, such as a hard brake. The gravity of these compounded errors has motivated research to measure and incorporate perception uncertainty into forecasting \cite{ivanovic2022propagating}. Furthermore, tracking errors such as miss-associations can have catastrophic consequences, inspiring work to maintain multiple tracking hypotheses \cite{weng2022mtp} throughout the entire pipeline.

Another line of work \cite{faf, intentnet, pnpnet, spagnn, ilvm} has proposed a more fundamental way to address these issues by shifting to an end-to-end detection and forecasting paradigm where the two tasks share high-dimensional scene features and are jointly optimized (\cref{fig:hook}\textcolor{red}{b}). Compared to traditional approaches where forecasting relies purely on object tracks, this paradigm can better propagate uncertainty downstream with a wider interface and intermediate representations optimized for both tasks. Nevertheless, these methods rely on a cascading inference where detection errors can still propagate to forecasting. 

Our method circumvents the established cascading approach by reformulating the detection and forecasting steps into a single, more general trajectory refinement task in Bird's-Eye-View (BEV), as shown in \cref{fig:hook}\textcolor{red}{c}. In most self-driving pipelines, the planner receives a set of object trajectories over time, making no distinction between the current time (detection) and the future (forecasting). Thus, we formulate our output to be a set of object trajectories, where each trajectory represents an object's pose in BEV from the current time into the future. In this formulation, detection is simply the particular case of the pose at the present time. 

To tackle object detection and trajectory forecasting as a pose refinement problem, we utilize multi-object, multi-hypothesis and temporal learnable queries refined by attending to LiDAR point clouds and high-definition (HD) maps.
However, naive global cross-attention to LiDAR and maps and self-attention between all queries are prohibitively expensive and difficult to optimize.
To make cross-attention practical, we pair each of the queries with a pose that represents our belief about the BEV position of a particular \emph{object}, at a specific \emph{time}, for an individual future behavior \emph{mode}, and perform local attention in a neighborhood instead.
To make self-attention practical with three-dimensional queries, we factorize it into object, time, and mode attention.

We demonstrate the effectiveness of our approach in two popular self-driving datasets, Argoverse 2 Sensor and Waymo Open, where \ourmodel{} outperforms the state-of-the-art in a broad range of detection, forecasting, and joint metrics. Importantly, through ablation studies we also show that our proposed refinement mechanism is powerful, that every proposed component contributes positively to the overall method, and that key design choices were made across multiple components.

\section{Related Work}
\label{sec:related_work}

\paragraphc{Detection}
Image-based object detectors traditionally leverage convolutions to extract features efficiently. These CNN-based approaches fall into two main classes: single-stage methods \cite{Redmon2015YouOL, Liu2015SSDSS, Zhou2019ObjectsAP} that make predictions w.r.t a grid of possible centers, and two-stage detectors \cite{Ren2015FasterRT, Cai2019CascadeRH} that have different networks first to generate proposals and then refine them. LiDAR-based detectors also leverage similar architectures for Bird's-Eye-View (BEV) detection \cite{pixor, Zhou2017VoxelNetEL ,Lang2018PointPillarsFE, Zhou2019EndtoEndMF}. More recently, DETR \cite{detr} redefines detection as a set prediction problem and uses a Transformer \cite{transformer} architecture coupled with a bipartite matching loss. Deformable-DETR \cite{deformabledetr} leverages reference points for each of the objects so that the attention layers focus on the local context around these. DAB-DETR \cite{liu2022dabdetr} extends this framework by feeding a DETR-based model with geometry-based anchor proposals, updating these spatial queries after each attention layer. We build on top of these works by encoding LiDAR point clouds through a CNN backbone \cite{pixor} and leveraging set-based transformers \cite{detr} with deformable attention for efficient geometric priors \cite{deformabledetr}.

\paragraphc{Forecasting}
High-definition (HD) maps and object past trajectories (tracks) are usually employed to forecast multiple BEV trajectories per object. The inputs can either be represented as a raster and processed with a CNN \cite{Cui2018MultimodalTP, multipath, PhanMinh2019CoverNetMB} or in vectorized form and processed with Graph Neural Networks (GNN) \cite{lanegcn,Deo2021MultimodalTP, Mo2022MultiAgentTP, cui2022gorela} or Transformers \cite{Gao2020VectorNetEH,girgis2021latent, gilles2021thomas,Mo2022MultiAgentTP, Zhou2022HiVTHV, agentformer, scenetransformer, latent,wayformer}. 
Most prior works decode multiple futures for each object by first encoding these inputs into object features. 
AutoBots \cite{girgis2021latent} and SceneTransformer \cite{scenetransformer} pioneered the use of mode queries: learnable parameters that separate distinct futures earlier in the network so the futures modes can interact with each other. QCNet \cite{zhou2023query} recently took this one step further, exploiting mode queries from the beginning of the network to predict an initial set of anchor trajectories, which they leverage to gather local information and refine once. However, these mode queries are stripped of temporal information, limiting their expressivity.
Inspired by this trend, we leverage a volume of learnable queries with object, mode, and time dimensions from the start of the network and generalize the refinement so that it can be done iteratively without any necessary changes to the design of subsequent blocks.

\paragraphc{End-to-end (E2E) Detection and Forecasting} Early attempts adopted a single-stage method by adding a lightweight forecasting header to an existing detection framework \cite{faf, intentnet}. More recent E2E detection and forecasting works use two-stage architectures \cite{pnpnet, interactiontransformer, spagnn, LiRaNet, multixnet, laserflow, ilvm, hu2023_uniad}, where the second stage performs relational reasoning between objects and the map. While their details differ, these models share a common framework. First, an encoder extracts shared features from raw sensor data. Next, a perception module predicts a set of candidate detections from the shared features. Then, a feature vector representing the object detections is extracted from the shared features by a pooling operation. Finally, a forecasting module takes the detections and their feature vectors and predicts future trajectories. While these two-stage architectures are more powerful and can propagate uncertainty via end-to-end training, they also partially reintroduce the compounding errors from the modular pipeline by cascading the detection and forecasting stages. In contrast, our approach tackles detection and forecasting as a single trajectory refinement task by decoding detections and forecasts in parallel.

\begin{figure}[t]
    \centering
    \includegraphics[width=\textwidth]{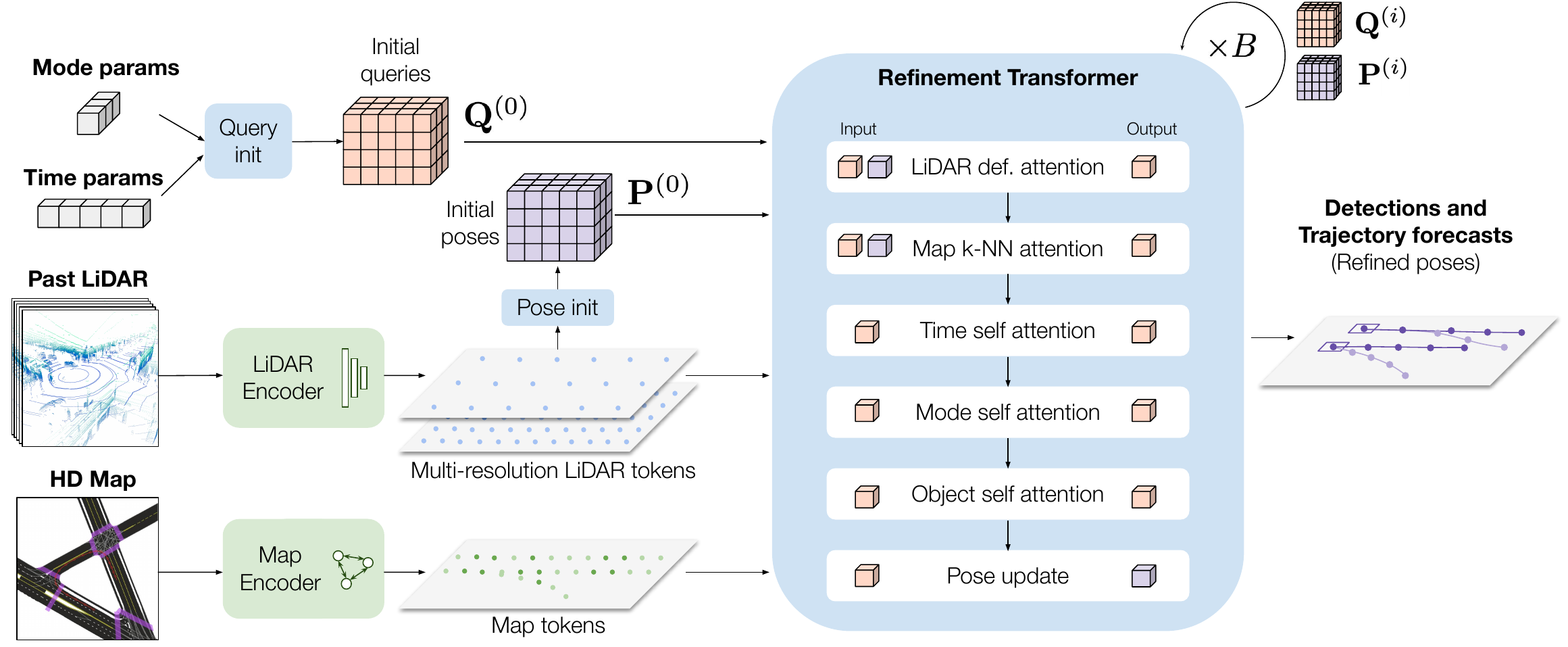}
    \caption{
        \ourmodel{} uses object queries and poses to represent objects' trajectories from the present time into the future. \ourmodel{} refines the initial estimates of the object queries and poses over $B$ blocks. Each refinement transformer block consists of several attention layers followed by a pose update. The set of poses at the end of the $B$-th block are the final detections and forecasts.
        }
    \label{fig:model}
\end{figure}
\section{Unifying Object Detection and Trajectory Forecasting}
\label{sec:method}

For a self-driving vehicle (SDV) to plan a safe and explainable maneuver, it must attain a thorough scene understanding, including where other traffic participants are now and where they will be in the future. In other words, a self-driving vehicle must understand object trajectories over time.
Because roads are reasonably flat, Bird's-Eye-View (BEV) is the de-facto representation for downstream planning in self-driving \cite{fan2018baidu,rhinehart2018deep,zeng2019end,sadat2019jointly,casas2021mp3}.
Thus, we design \ourmodel{}, an end-to-end model that turns raw observations into a set of BEV pose trajectories $\mathbf{P}$ for $N$ objects, $F$ future hypothesis per object, and $T$ time steps. Each pose here is a triplet $(x, y, \theta)$ describing the object's centroid and heading (yaw). 
In addition to the poses, it also predicts the confidence of the object's presence, its dimensions, and the probability of each future hypothesis.

\Cref{fig:model} shows an overview of \ourmodel{}. 
It is composed of two parts: \textcolor{LimeGreen}{extracting BEV scene representations}, and \textcolor{CornflowerBlue}{refining object trajectories}. 
In \Cref{method:input}, we explain how to extract BEV representations of the scene from LiDAR point clouds and HD maps. While this is not our main contribution, the reader should understand these representations as they are essential to the task. 
Then, \Cref{method:object_transformer} explains our core contribution: a refinement transformer that refines poses for multiple objects and hypotheses over time via attention.
Finally, \Cref{method:learning} describes the learning objective used to train our method end-to-end in a single stage.

\subsection{Extracting Scene Representations from LiDAR and HD maps} \label{method:input}

To understand the environment the SDV is in, we use LiDAR as our sensory input, as it has established itself as the primary sensor in most SDV's platforms \cite{argoverse2,waymo}.
In addition, our model can optionally utilize high-definition (HD) maps as an offline source of information.
In contrast to previous end-to-end methods, our approach does not require the fusion of these two inputs at the scene level, which is challenging as they have a very different structure and typically results in information loss (e.g., by creating a raster of the map to fuse with BEV LiDAR representations \cite{intentnet,spagnn}).
Instead, \ourmodel{} learns to attend to relevant parts of each modality for each object in its refinement transformer (explained in \Cref{method:object_transformer}).

Next, we describe the LiDAR and Map encoders, which extract meaningful features from these two modalities and provide these as tokens with spatial locations associated with the refinement transformer for efficient attention.

\paragraphc{LiDAR encoder} 
To extract rich geometric and motion features from LiDAR, we first voxelize a stack of $H=5$ past LiDAR sweeps (0.5s) in BEV, where all points in all LiDAR sweeps are expressed in the vehicle coordinate frame at the time the last sweep was acquired \cite{pixor,faf}.
To obtain the voxel features, we use a PointNet \cite{qi2017pointnet} for the points residing inside each voxel, as done in VoxelNet \cite{Zhou2017VoxelNetEL}. We use a voxel size of 10 cm.
Then, a 2D ResNet backbone \cite{he2016deep} encodes these voxelized features to generate multi-scale BEV feature maps with resolutions of 0.4, 0.8, \SI{1.6}{m}. The backbone also leverages dynamic convolution \cite{chen2020dynamic},
squeeze-excitation \cite{hu2018squeeze}, and multi-scale deformable attention \cite{deformabledetr}, the details of which are included in the 
supplementary.
Each multi-resolution feature map can then be used as a sequence of LiDAR tokens for cross-attention in our refinement transformer. We note that each of the tokens has an associated centroid corresponding to the position of its BEV pixel, which will be important for efficient attention.

\paragraphc{Map encoder}
To summarize the heterogeneous information in the HD map into map tokens, we use a LaneGCN \cite{lanegcn} with GoRela \cite{cui2022gorela} positional encodings.
In summary, we first preprocess HD maps into lane graphs, a representation in which lane centerlines are divided into segments (each \SI{3}{m} in size), and each segment center constitutes a node in the graph. Each node contains information about the length, heading, and curvature of the segment it represents and the left and right lane boundary distances and types.
We connect the nodes with edges according to the topology described in the HD map (e.g., successors, neighbors). Then, a graph convolution network \cite{kipf2016semi} is employed to extract map embeddings for each lane graph node. 
We treat each lane graph node embedding as a map token and store the corresponding coordinates for later use in efficient cross-attention.

\subsection{Detection and Forecasting as Trajectory Refinement} \label{method:object_transformer}

The goal of the refinement transformer is to refine a set of object poses $\mathbf{P} \in \mathbb{R}^{N \times F \times T \times 3}$ with $N$ objects, $F$ future hypothesis per object, and $T$ time steps for each trajectory.
Each trajectory consists of a set of poses $\{(x_t, y_t, \theta_t) \; | \; t \in \{0, \cdots, T-1\}\}$ denoting the centroid and heading of the object in BEV, where $t=0$ is the current object pose and $t>0$ its future poses.
In addition, each future hypothesis also includes the associated mode probability.
For the first time step $t=0$, we also estimate the probability that the object exists $c$ (i.e., object confidence) and the length and width in BEV $(l, w)$, which are assumed constant over the future time horizon.

To jointly model the current and future poses of an object, we utilize spatial-temporal object queries $\mathbf{Q}^{(i)} \in \mathbb{R}^{N \times F \times T \times d}$ that contain $d$-dimensional latent representations (i.e., features) of the objects for the different future hypothesis and time steps.
At the beginning of inference ($i=0$), these object queries are initialized from learned parameters \cite{liu2022dabdetr} to get $\mathbf{Q}^{(0)}$. Our refinement transformer refines them iteratively using $B$ blocks ($i \in 1 \dots B$), each consisting of several interleaved attention layers.
Queries $\mathbf{Q}^{(i)}$ are ``spatially anchored'' with poses $\mathbf{P}^{(i)}$, which are also refined iteratively during the $B$ transformer blocks. The final object poses that constitute the detections and forecasts are simply those after $B$ transformer blocks, i.e., $\mathbf{P} = \mathbf{P}^{(B)}$.

\begin{figure}[t]
    \centering
    \includegraphics[width=\textwidth]{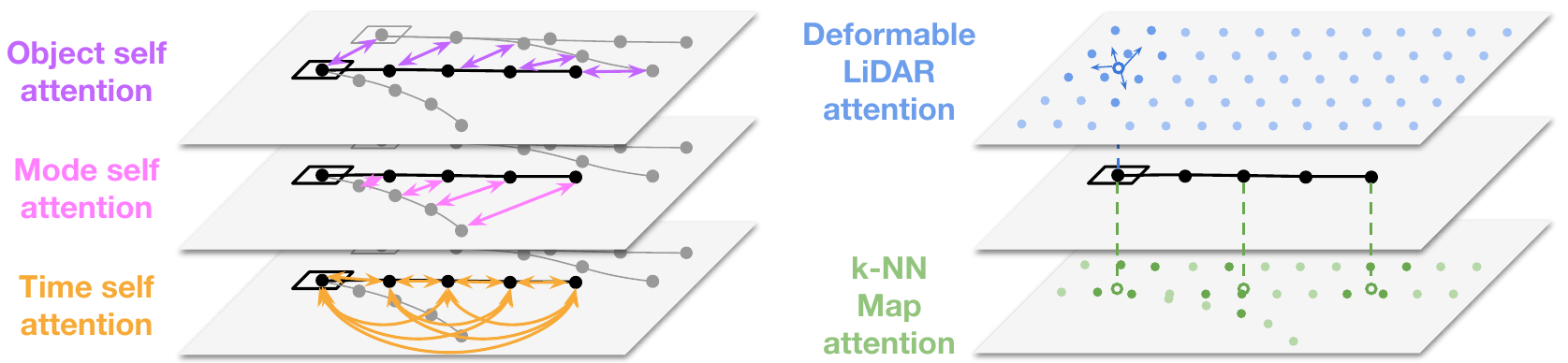}
    \caption{
        A diagram of the attention mechanisms used in \ourmodel{}. We factorize self-attention in our learnable query volume into object, mode, and time axes. For efficiency and ease of optimization, we perform deformable attention to LiDAR and $k$-nearest neighbor attention to the map.
        }
    \label{fig:attention}
\end{figure}

A refinement transformer block contains $A$ attention layers, each tasked with refining object queries by incorporating information from different inputs and other queries.
Each attention layer $\mathcal{A}$ (with index $a$) updates the object queries as follows:
\begin{align}
    \mathbf{Q}^{(i, a)}_{\texttt{att}} &= LN(\mathbf{Q}^{(i, a)} + \mathcal{A}(\mathbf{Q}^{(i, a)}, \cdot)), \\
    \mathbf{Q}^{(i, a+1)} &= LN(\mathbf{Q}_{\texttt{att}}^{(i, a)} + FFN(\mathbf{Q}_{\texttt{att}}^{(i, a)})),
\end{align}
where $LN(\cdot)$ denotes layer normalization and $\mathcal{A}(\mathbf{Q}^{(i, a)}, \cdot)$ indicates one of the specialized attentions illustrated in \Cref{fig:attention}, with $\mathbf{Q}^{(i, a)}$ as queries and any other features as keys/values. 
The updated queries after all the $A$ attention layers in a transformer refinement block are $\mathbf{Q}^{(i+1)}=\mathbf{Q}^{(i, A)}$.
From this point onwards, we drop the double superscript ${(i, a)}$ for brevity and use ${(i)}$ to refer to the queries at any point during the $i$-th block. 

At the end of each transformer block, a pose update function $\mathcal{U}$ updates the poses given the updated queries $\mathbf{Q}^{(i+1)}$ and previous poses $\mathbf{P}^{(i+1)} = \mathcal{U}(\mathbf{Q}^{(i+1)}, \mathbf{P}^{(i)})$.
The updated queries $\mathbf{Q}^{(i+1)}$ and poses $\mathbf{P}^{(i+1)}$ become the input to the next transformer block.

Next, we dive into the details of all the components.

\paragraphc{Query initialization} We start off with learnable \emph{mode parameters} $\mathbf{F} \in \mathbb{R}^{1 \times F \times 1 \times d}$ to consider $F$ possible futures \cite{girgis2021latent,zhou2023query,varadarajan2022multipath++}, and learnable \emph{time parameters} $\mathbf{T} \in \mathbb{R}^{1 \times 1 \times T \times d}$ to consider multiple time steps. 
The intuition is that the attention layers may want to perform different reasoning for different modes or time steps.
To obtain the initial query volume $\mathbf{Q}^{(0)}$ we sum the mode and time parameter tensors (with broadcasting), and expand the first dimension to the number of objects $N$, obtaining a tensor of size $N \times F \times T \times d$. 
Thus, all objects are initialized with the same query features and are only differentiated through their different initial poses $\mathbf{P}^{(0)}$, which will guide the local attention operations and iteratively update each query distinctively.

\paragraphc{Pose initialization} 
Our method relies on an initial estimate of the poses, $\mathbf{P}^{(0)}$.
While we can estimate this in many ways, we found it best to initialize from the output of a lightweight detector header that re-uses the LiDAR tokens at a single resolution of \SI{0.4}{m}.
The detector is a very lightweight convolutional, single-shot, anchorless detector that predicts a score heatmap and bounding box parameters per pixel.
The per-pixel bounding box proposals go through NMS to yield $N_\texttt{init}$ objects. If $N_\texttt{init} > N$, we take the top-$N$ candidates by confidence score.
This detector's goal is to achieve very high recall, as detections will be refined in the refinement transformer to obtain high precision.
These bounding boxes constitute $\mathbf{P}^{(0)}_{t=0}$, and we assume objects are stationary (same pose over time and mode dimensions) to initialize $\mathbf{P}^{(0)}_{t>0}$. This static initialization is not an issue since the pose volume $\mathbf{P}^{(i)}$ is dynamically refined for all time steps over different refinement transformer blocks.

\paragraphc{LiDAR cross-attention} 
Object queries cross-attend to the LiDAR tokens. Since the number of LiDAR tokens $N_{lidar}$ is considerable and fully connected attention brings unaffordable memory consumption at LiDAR feature maps with high resolution, we follow~\cite{deformabledetr} and utilize deformable attention to predict multiple offsets $\mathbf{o} \in \mathbb{R}^{\ell \times 2}$ originating from the object poses $\mathbf{P}^{(i)}$, getting reference points $\mathbf{r}^{(i)} = \mathbf{P}^{(i)} + \mathbf{o}^{(i)}$. Offsets $\mathbf{o}^{(i)}$ are predicted from the object queries $\mathbf{Q}^{(i)}$, and bilinear interpolation is performed around each reference point, thus limiting the attention to $4\ell$ LiDAR tokens around the pose of each object token, which yields a query-to-LiDAR attention map of $N \times 4\ell$ where $\ell \ll N_{lidar}$, significantly reducing the computational cost. For efficiency purposes, in our implementation, we only attend around the poses for the first time step $\mathbf{P}^{(i)}_{t=0}$, which is where the most relevant LiDAR evidence for a particular object lays. 

\paragraphc{Map cross-attention} 
Object queries cross-attend to neighboring map tokens to exploit the information in the lane graph. 
Since the map encoder is a graph convolutional network, each lane-graph node embedding (or map token) already encodes information about the neighborhood. Thus, attending to the map tokens far from the object poses is unnecessary, so we limit the cross-attention to the $k$ nearest map tokens from the object pose, reducing the computation from $O(N N_{map})$ to $O(N k)$ where $k \ll N_{map}$ (around 100 times in practice). We empirically found that limiting the attention to $k$-nearest map tokens greatly aids learning.
Furthermore, since our transformer also has time self-attention, we do not need to attend to the map for all time steps and can update only a specific subset of the time steps in the object queries.
In our final model, we attend to map for $t \in \{0, \lfloor (T-1) / 2\rfloor, T-1\}$.

\paragraphc{Self-attention} 
For efficiency purposes, we factorize self-attention into \textit{time self-attention}, \textit{mode self-attention} and \textit{object self-attention}. 
In \textit{time self-attention}, the queries $\mathbf{Q}^{(i)}$ only attend to other queries from the same object and mode. In other words, information only propagates along the time dimension. More precisely, the sequence dimension is of size $T$, and $N \times F$ constitutes the batch size, allowing the model to output consistent poses over time.
Similarly, in \textit{mode self-attention}, the queries only attend to other queries from the same object and time. The intuition is to enable the model to diversely place the modes to cover possible futures while still being realistic.
Finally, in \textit{object self-attention}, each query can only attend to queries from other objects at the same time step and mode, allowing the queries to gather contextual information about neighboring traffic participants, which helps predict consistent behaviors (e.g., avoid collisions between different objects' trajectories).
Over different transformer blocks, these operations are repeated, enabling \ourmodel{} to learn complex dependencies across the query volume efficiently.

\paragraphc{Pose update} 
At the end of each refinement transformer block, the poses are updated.
The object tokens at current time $\mathbf{Q}_{t=0}$ are used to produce object bounding boxes, and object tokens at the future time $\mathbf{Q}_{t>0}$ are used to generate multi-modal future trajectories.
The detection bounding box parameters $\mathbf{D}=(x, y, l, w, \theta, c)$ are predicted as $\mathbf{D}^{(i+1)} = \mathbf{D}^{(i)} + \text{MLP}(\frac{1}{N}\sum_{f=1}^F \mathbf{Q}^{(i)}_{t=0, f})$, and the current pose $\mathbf{P}^{(i+1)}_{t=0}$ is updated with $(x, y, \theta)$ from the updated detections $\mathbf{D}^{(i+1)}$ for all modes.
To predict the future trajectories, we use a single-layer bi-directional GRU \cite{cho2014learning} to process the temporal features for each mode and object in parallel. For each time step, an MLP predicts four outputs: the location $(\mu_x, \mu_y)$ and the scale $(\sigma_x, \sigma_y)$ of an isotropic Laplacian distribution from the GRU hidden state at that time step. 
The updated future poses $\mathbf{P}^{(i+1)}_{t>0}$ are composed of the Laplacian's $(\mu_x, \mu_y)$, and the heading $\theta$ is computed via temporal finite differences of the waypoints.
TransFuser~\cite{chitta2022transfuser} inspired using the GRU, and we find it yields smoother trajectories empirically.
Finally, we also predict the score of each future as the logits of a categorical distribution with $F$ classes with another MLP that takes the average of the temporal GRU hidden states for each object and mode as input.

\subsection{Learning} \label{method:learning}

All the parameters in \ourmodel{} are trained jointly from scratch in an end-to-end fashion. We stop the gradients of the poses $\mathbf{P}^{(i)}$ from flowing in between refinement transformer blocks, which we found to improve learning. We optimize a multi-task objective that is a combination of an initial pose loss $L_\texttt{init}$, a detection refinement loss $L_\texttt{det}$ and a forecasting refinement loss $L_\texttt{for}$, where the detection and forecasting losses are computed at every block: $L = L_\texttt{init} + \sum_{i=1}^{B} L_\texttt{det}^{(i)} + \alpha L_\texttt{for}^{(i)}$.
\paragraphc{Initial pose loss}
We use a binary focal loss to supervise the score heatmap. The target heatmap is $1$ (positive) at pixels with a ground-truth object centroid and $0$ (negative) elsewhere.
Bounding box parameters are supervised with an IoU loss for the positive pixels in the target heatmap. 
$L_\texttt{init}$ is the sum of these two losses.

\paragraphc{Detection refinement loss}
For the detection task, we use a combination of a binary focal loss~\cite{focalloss} $L_\texttt{det}^\texttt{cls}$ for classification, with an L1 loss $L_\texttt{det}^\texttt{L1}$ and a generalized intersection-over-union (gIoU) loss $L_\texttt{det}^\texttt{giou}$ to regress the parameters of bounding boxes: $L_\texttt{det} = L_\texttt{det}^\texttt{cls} + \beta L_\texttt{det}^\texttt{L1} + \gamma L_\texttt{det}^\texttt{giou}$.
To calculate the targets for these losses, we first match the detections to the ground truth bounding boxes through bipartite matching as proposed in DETR~\cite{detr}.

\paragraphc{Forecasting refinement loss}
We use a simple winner-takes-all (WTA) loss on the mixture of trajectories, where an isotropic Laplacian parameterizes each waypoint and the mode probabilities are the weights of the mixture \cite{Zhou2022HiVTHV,zhou2023query}.
We supervised the mode probabilities with a cross-entropy loss $L_\texttt{for}^\texttt{cls}$, where the \emph{winner mode} is the one in which centroids are the closest to the GT.
We supervise the Laplacian waypoints (location and scale) predicted for the winner mode with negative log-likelihood $L_\texttt{for}^\texttt{nll}$, and the ones for the other modes are left unsupervised. In sum, the forecasting loss is $L_\texttt{for} = L_\texttt{for}^\texttt{nll} + L_\texttt{for}^\texttt{cls}$.
Note that we only supervise the forecasted trajectories from true positive detections during training (defined as those having over $0.5$ IoU with the ground truth).

\begin{table*}[t]
	\centering
    \fontsize{7.5pt}{8.5pt}\selectfont
    \begin{tabularx}{\textwidth}{l|ss|sss|sssssssss}
        \toprule
        \multirow{3}{*}{{Argoverse2}}                  & \multicolumn{2}{c|}{Joint}         & \multicolumn{3}{c|}{Detection}                            & \multicolumn{7}{c}{Forecasting}                                                                                                                                                     \\
                                                         \cmidrule{2-3}         \cmidrule(l{2pt}r{2pt}){4-6}                               \cmidrule(l{2pt}){7-13}
                                                         & \multicolumn{2}{c|}{AP \ua} & \multicolumn{3}{c|}{AP @ IoU \ua}                & \multicolumn{2}{c}{MR \da}  & \multicolumn{2}{c}{ADE \da} &  \multicolumn{2}{c}{FDE \da}            & \multicolumn{1}{c}{bFDE \da}                                          \\  
                                                         \cmidrule(l{2pt}r{2pt}){2-3}                               \cmidrule(l{2pt}r{2pt}){4-6}                               \cmidrule(l{2pt}r{2pt}){7-8} \cmidrule(l{2pt}r{2pt}){9-10} \cmidrule(l{2pt}r{2pt}){11-12}                                      \cmidrule(l{2pt}r{2pt}){13-13}
                                                       & Occ                             & Traj                    & @0.3             & @0.5             & @0.7                    & $K$ = 1          & $K$ = 6          & $K$ = 1          & $K$ = 6          & $K$ = 1          & $K$ = 6                 & $K$ = 6          \\
        \midrule
        MultiPath~\cite{multipath}                     & 62.7 &43.7 &95.5 &91.7 &76.3 &43.3 &32.9 &2.12 &1.03 &4.81 &2.26 &2.87 \\
        LaneGCN~\cite{lanegcn}                         & 64.4 &41.9 &95.5 &91.7 &76.3 &41.2 &27.3 &2.08 &0.86 &4.70 &1.85 &2.30 \\
        SceneTF~\cite{scenetransformer}                & 63.9 &42.9 &95.5 &91.7 &76.3 &40.9 &24.5 &1.98 &0.86 &4.49 &1.85 &2.32 \\
        GoRela~\cite{cui2022gorela}                    & 57.8 &42.9 &95.5 &91.7 &76.3 &39.8 &21.3 &1.84 &1.00 &4.05 &2.10 &2.46 \\
        \midrule
        FaF~\cite{faf}                                 & 64.4 &39.5 &94.7 &90.8 &74.8 &40.1 &22.1 &1.82 &0.84 &4.21 &1.79 &2.27 \\
        IntentNet~\cite{intentnet}                     & 66.6 &40.0 &95.4 &91.3 &75.3 &40.6 &24.5 &1.88 &0.80 &4.36 &1.67 &2.11 \\
        MultiPathE2E~\cite{multipath}                     & 66.1 &44.1 &95.6 &91.7 &76.3 &41.6 &31.5 &1.84 &0.93 &4.25 &2.07 &2.65 \\
        LaneGCNE2E~\cite{lanegcn}                         & 65.9 &39.2 &94.8 &91.0 &75.0 &40.7 &24.5 &1.93 &0.82 &4.51 &1.73 &2.17 \\
        GoRelaE2E~\cite{cui2022gorela}                    & 58.4 &38.3 &94.7 &91.1 &75.5 &41.3 &23.2 &2.36 &1.35 &5.06 &2.74 &3.12 \\
        \midrule                                          
        \ourmodel{}                                    & \textbf{73.0} &\textbf{50.0} &\textbf{95.8} &\textbf{92.9} &\textbf{80.2} &\textbf{37.0} &\textbf{15.4} &\textbf{1.54} &\textbf{0.55} &\textbf{3.59} &\textbf{1.19} &\textbf{1.86} \\
        \bottomrule 
    \end{tabularx}
    \vspace{1mm}
	\caption{Comparing \ourmodel{} to state-of-the-art detection and forecasting models on AV2.}
	\label{tab:av2-comparison}
\end{table*}

\section{Experiments}
\label{sec:experiments}
This section provides a comprehensive analysis of \ourmodel{} from two perspectives.
\begin{itemize}
    \item Comparing our method to a relevant set of state-of-the-art baselines, including modular detection-tracking-forecasting methods and end-to-end detection-forecasting methods. Our results show clear improvements across all detection, forecasting, and joint metrics in two datasets: Argoverse 2 Sensor \cite{argoverse2} and Waymo Open \cite{waymo}.
    \item Understanding the aspects that make our method attain this higher level of performance with thorough ablation studies. Our results show that (i) all presented components are necessary, (ii) refining the poses over multiple transformer blocks is very effective and benefits both detection and forecasting and (iii) we made key design choices in pose initialization as well as LiDAR and map cross attention.
\end{itemize}
Refer to our supplementary for more implementation details, qualitative results, and ablations (number of blocks, attention ordering, dropping time/mode parameters).

\paragraphc{Datasets}
To show the generality of our approach across different LiDAR sensors and maps, we conduct experiments in  Argoverse 2 Sensor~\cite{argoverse2} Waymo Open Dataset\cite{waymo}.
Argoverse 2 Sensor (AV2) contains 850 sequences collected in Austin, Detroit, Miami, Palo Alto, Pittsburgh, and Washington D.C. Each sequence lasts 15 seconds and consists of 150 frames of point clouds from two 32-beam LiDARs that provide an average of 107K points per frame. Argoverse 2 includes HD maps. We use 700 sequences for training and 150 sequences for validation.
Waymo Open Dataset (WOD) \cite{waymo} is made of 1150 sequences collected from 6 different cities in the U.S. Each sequence lasts 20 seconds and contains 200 frames of point clouds from five LiDAR sensors with an average of 177K points per frame. 

\begin{table*}[t]
	\centering
    \fontsize{7.5pt}{8.5pt}\selectfont
    \begin{tabularx}{\textwidth}{l|ss|sss|sssssssss}
        \toprule
        \multirow{3}{*}{{Waymo Open}}                  & \multicolumn{2}{c|}{Joint}         & \multicolumn{3}{c|}{Detection}                            & \multicolumn{7}{c}{Prediction}                                                                                                                                                     \\
                                                         \cmidrule{2-3}         \cmidrule(l{2pt}r{2pt}){4-6}                               \cmidrule(l{2pt}){7-13}
                                                         & \multicolumn{2}{c|}{AP \ua} & \multicolumn{3}{c|}{AP @ IoU \ua}                & \multicolumn{2}{c}{MR \da}  & \multicolumn{2}{c}{ADE \da} &  \multicolumn{2}{c}{FDE \da}            & \multicolumn{1}{c}{bFDE \da}                                          \\  
                                                         \cmidrule(l{2pt}r{2pt}){2-3}                               \cmidrule(l{2pt}r{2pt}){4-6}                               \cmidrule(l{2pt}r{2pt}){7-8} \cmidrule(l{2pt}r{2pt}){9-10} \cmidrule(l{2pt}r{2pt}){11-12}                                      \cmidrule(l{2pt}r{2pt}){13-13}
                                                       & Occ                             & Traj                    & @0.3             & @0.5             & @0.7                    & $K$ = 1          & $K$ = 6          & $K$ = 1          & $K$ = 6          & $K$ = 1          & $K$ = 6                 & $K$ = 6          \\
        \midrule
        MultiPath~\cite{multipath}                     & 60.2 &31.9 &88.6 &85.1 &72.9 &45.4 &37.2 &2.54 &1.49 &5.63 &3.31 &3.93 \\
        LaneGCN~\cite{lanegcn}                         & 63.9 &30.4 &88.6 &85.1 &72.9 &44.5 &31.4 &2.53 &1.03 &5.63 &2.19 &2.63 \\
        SceneTF~\cite{scenetransformer}                & 64.9 &31.4 &88.6 &85.1 &72.9 &44.4 &27.4 &2.44 &1.00 &5.39 &2.16 &2.66 \\
        GoRela~\cite{cui2022gorela}                    & 61.3 &29.6 &88.6 &85.1 &72.9 &41.1 &19.6 &2.02 &1.07 &4.21 &2.07 &2.41 \\
        \midrule
        FaF~\cite{faf}                                 & 62.3 &23.9 &87.3 &82.3 &67.7 &44.0 &27.7 &2.41 &1.03 &5.40 &2.22 &2.58 \\
        IntentNet~\cite{intentnet}                     & 62.9 &24.9 &87.8 &83.2 &68.8 &43.7 &27.7 &2.48 &1.03 &5.53 &2.21 &2.56 \\
        MultiPathE2E~\cite{multipath}                     & 57.6 &32.7 &88.6 &84.7 &71.1 &44.2 &34.9 &2.13 &1.25 &4.87 &2.75 &3.39 \\
        LaneGCNE2E~\cite{lanegcn}                         & 63.7 &25.1 &88.1 &83.6 &70.0 &43.5 &26.4 &2.37 &0.96 &5.33 &2.03 &2.39 \\
        GoRelaE2E~\cite{cui2022gorela}                    & 61.5 &26.4 &88.1 &84.0 &71.1 &42.7 &21.9 &2.37 &1.35 &4.78 &2.50 &2.86 \\
        \midrule                                          
        \ourmodel{}                                    & \textbf{70.4} &\textbf{37.1} &\textbf{89.8} &\textbf{86.5} &\textbf{74.9} &\textbf{40.3} &\textbf{18.0} &\textbf{1.87} &\textbf{0.59} &\textbf{4.29} &\textbf{1.25} &\textbf{1.93} \\
        \bottomrule 
    \end{tabularx}
    \vspace{1mm}
	\caption{Comparing our proposed  \ourmodel{} to the state-of-the-art detection and forecasting models on the Waymo Open Dataset dataset.}
	\label{tab:wod-comparison}
\end{table*}

\paragraphc{Task settings}
Provided with 0.5 seconds of LiDAR history as well as the HD map, models are tasked to detect vehicles inside the region of interest (RoI) with at least one lidar point, and forecast six possible future trajectories at 2Hz, each 5 seconds long. The RoI is a square centered around the ego vehicle of size $80 \times 80$ meters in AV2 and $150 \times 150$ meters in WOD. The rationale for having different RoIs is to keep the task complexity similar, as WOD features a denser and longer-range LiDAR than AV2.

\paragraphc{Metrics}
We consider a comprehensive set of metrics including those widely used in object detection and trajectory forecasting, and new metrics to evaluate both tasks jointly. Next, we explain them at a high level (more details in the supplementary).
\begin{itemize}
    \item Detection: average precision at multiple IoUs (AP@IoU) measures the ability to estimate object existence (low IoUs) and localization accuracy (high IoUs).
    \item Forecasting: we follow the metrics proposed by Argoverse 2 Motion \cite{argoverse2}: miss rate (MR), average displacement error (ADE), final displacement error (FDE), and its Brier variant (bFDE). Errors are reported for the most likely predicted trajectory ($K = 1$) and the best of 6 trajectories ($K = 6$). These metrics are evaluated at a common detection recall point for all models of $80 \%$ at an IoU threshold of 0.5, as proposed in previous works \cite{spagnn,ilvm,cui2021lookout}. To address the fact that most objects in AV2 and WOD datasets are stationary, we bucket the objects in stationary and dynamic, calculate metrics for each bucket, and report the macro-average. 
    \item Joint: these metrics capture the end-to-end performance in a single metric. Drawing inspiration from occupancy forecasting works \cite{mahjourian2022occupancy,agro2023implicit}, we compute Occupancy Average Precision (OccAP) by rendering the object detections and forecasted trajectories considering the confidence about the object's existence and the mode probabilities. In addition, we extend the Trajectory Average Precision (TrajAP) proposed in WOD Motion \cite{ettinger2021large} to also penalize false positive and false negative detections.
\end{itemize}

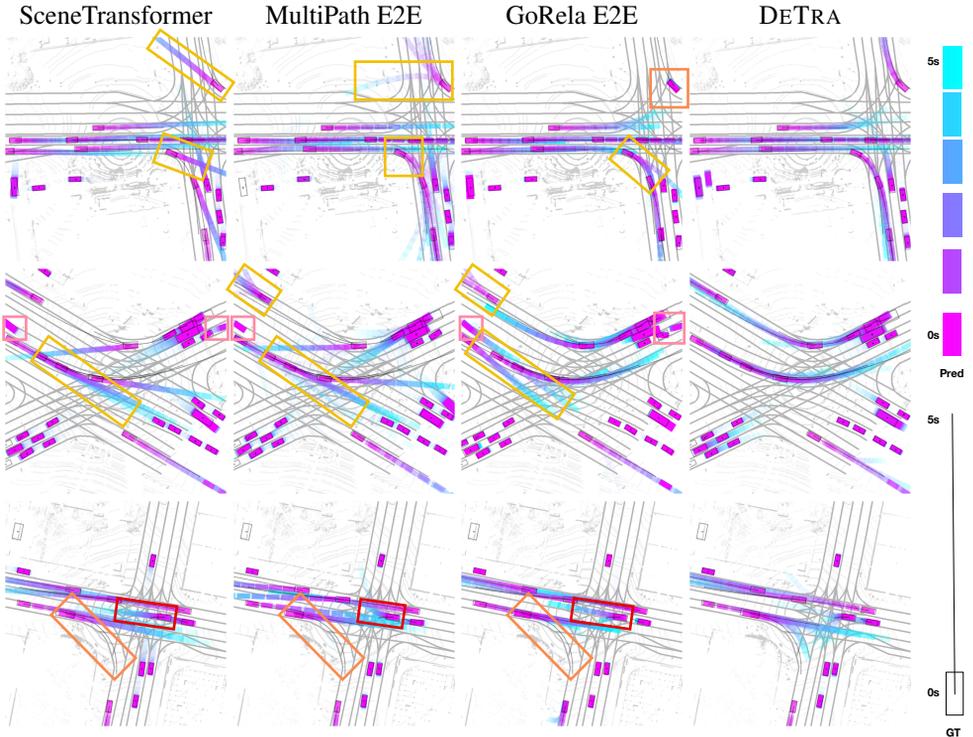
\begin{figure*}[t]
    \centering
    \definecolor{fp-mode}{RGB}{230, 0, 0}
    \definecolor{fn-mode}{RGB}{250, 138, 77}
    \definecolor{off-map}{RGB}{241, 192, 0}
    \definecolor{detection-color}{RGB}{255, 138, 165}
    \begin{tikzpicture}
        \pgfmathsetlengthmacro{\imw}{0.25\textwidth}
        \pgfmathsetlengthmacro{\bw}{1.0pt}

        \node[inner sep=0pt, outer sep=0, anchor=west]  (scene1st)        at (0, 0)                 {\includegraphics[width=\imw]{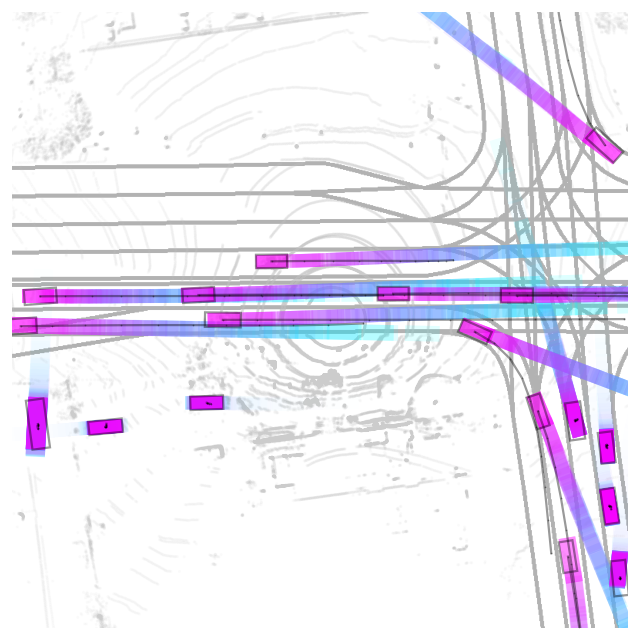}};
        \node[inner sep=0pt, outer sep=0, anchor=west]  (scene1mtp)       at (scene1st.east)        {\includegraphics[width=\imw]{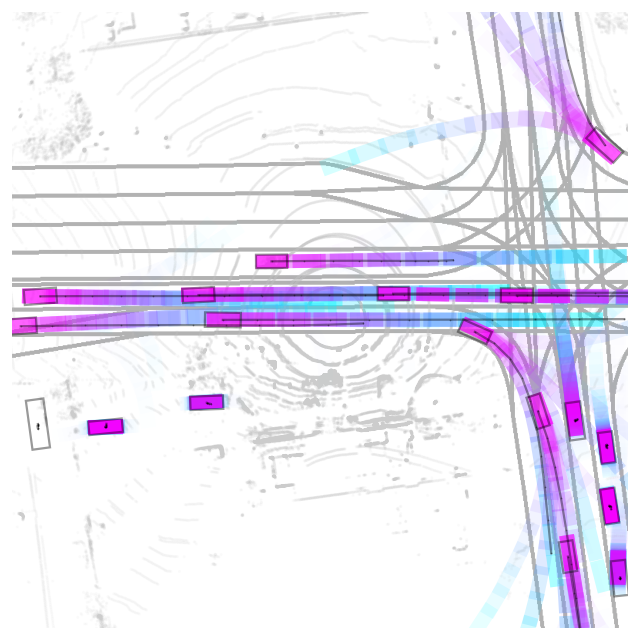}};
        \node[inner sep=0pt, outer sep=0, anchor=west]  (scene1gorela) at (scene1mtp.east)       {\includegraphics[width=\imw]{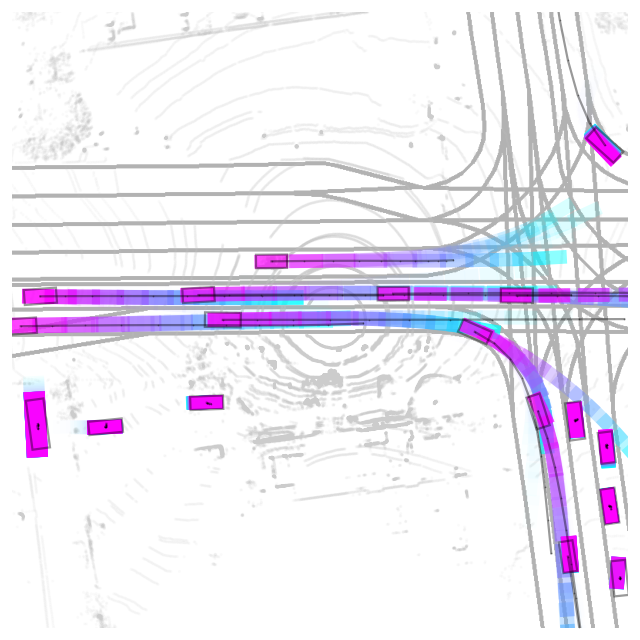}};
        \node[inner sep=0pt, outer sep=0, anchor=west]  (scene1detra)     at (scene1gorela.east) {\includegraphics[width=\imw]{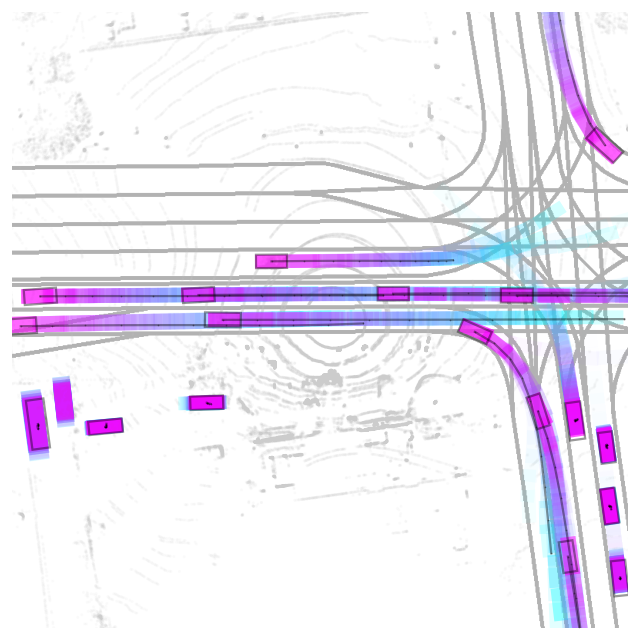}};
        \node[anchor=south] at (scene1st.north) {SceneTransformer};
        \node[anchor=south] at (scene1mtp.north) {MultiPath E2E};
        \node[anchor=south] at (scene1gorela.north) {GoRela E2E};
        \node[anchor=south] at (scene1detra.north) {\ourmodel{}};

        \node[anchor=north west] at (scene1detra.north east) {\includegraphics[height=0.75\textwidth]{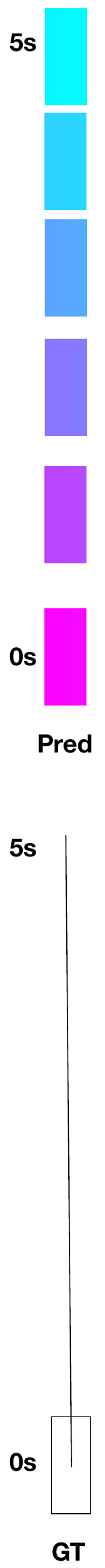}};

        \node[inner sep=0pt, outer sep=0, anchor=north] (scene2st)        at (scene1st.south)       {\includegraphics[width=\imw]{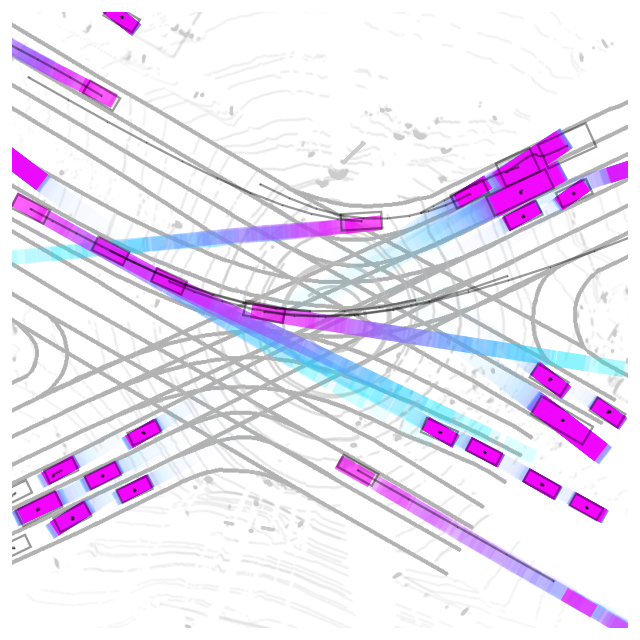}};
        \node[inner sep=0pt, outer sep=0, anchor=west]  (scene2mtp)       at (scene2st.east)        {\includegraphics[width=\imw]{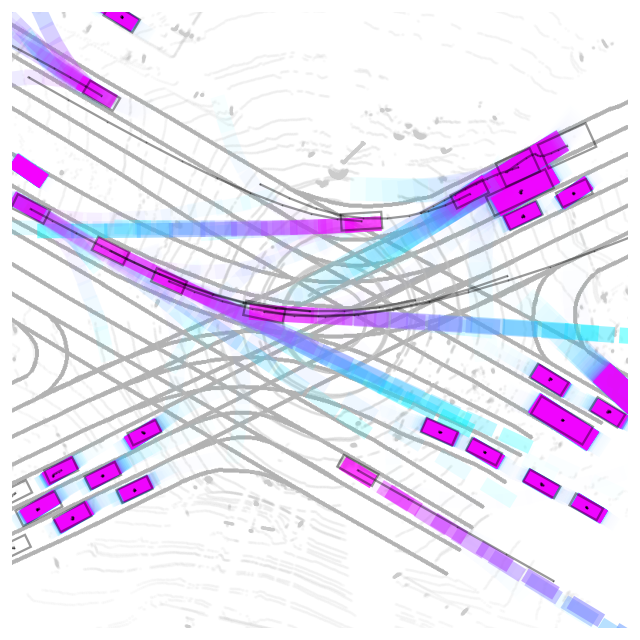}};
        \node[inner sep=0pt, outer sep=0, anchor=west]  (scene2gorela) at (scene2mtp.east)       {\includegraphics[width=\imw]{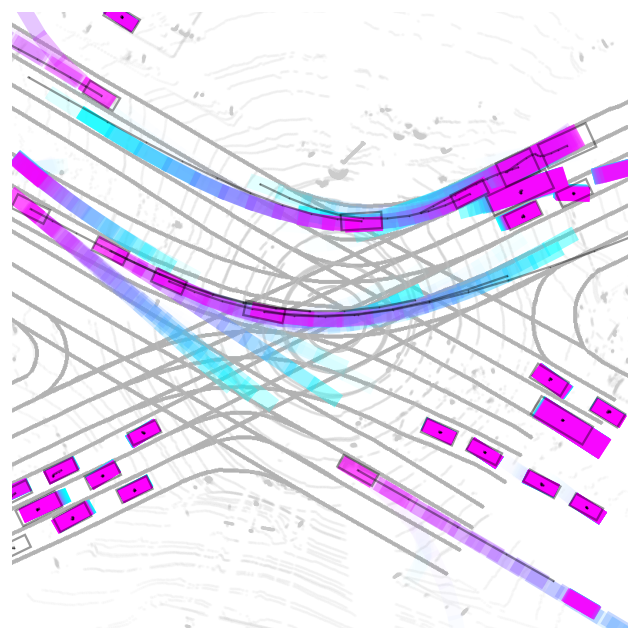}};
        \node[inner sep=0pt, outer sep=0, anchor=west]  (scene2detra)     at (scene2gorela.east) {\includegraphics[width=\imw]{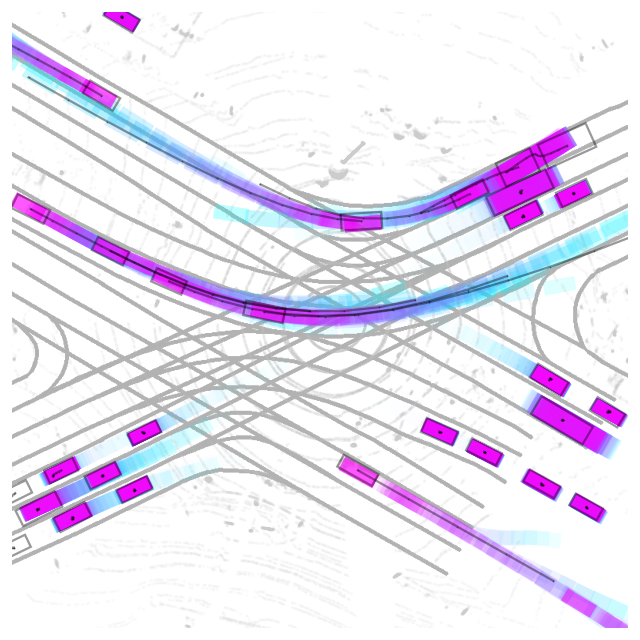}};

        \node[inner sep=0pt, outer sep=0, anchor=north] (scene3st)        at (scene2st.south)       {\includegraphics[width=\imw]{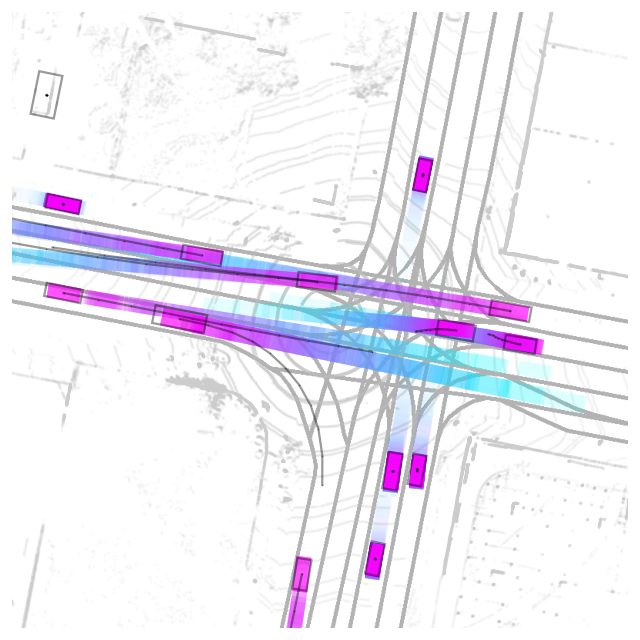}};
        \node[inner sep=0pt, outer sep=0, anchor=west]  (scene3mtp)       at (scene3st.east)        {\includegraphics[width=\imw]{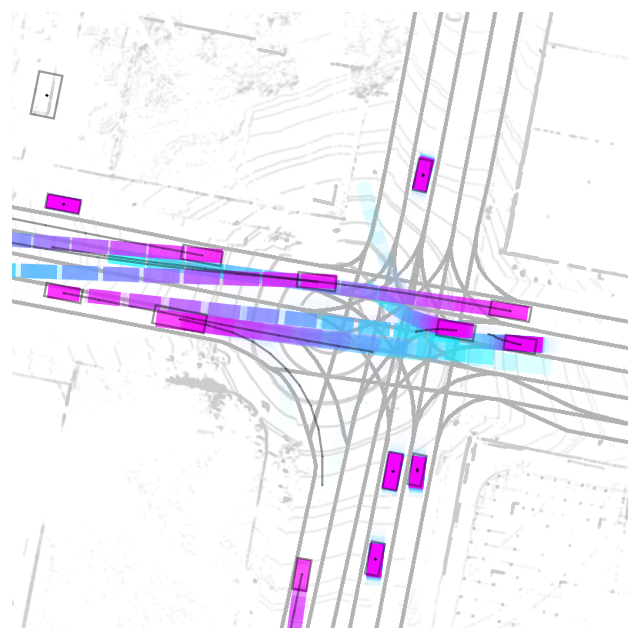}};
        \node[inner sep=0pt, outer sep=0, anchor=west]  (scene3gorela) at (scene3mtp.east)       {\includegraphics[width=\imw]{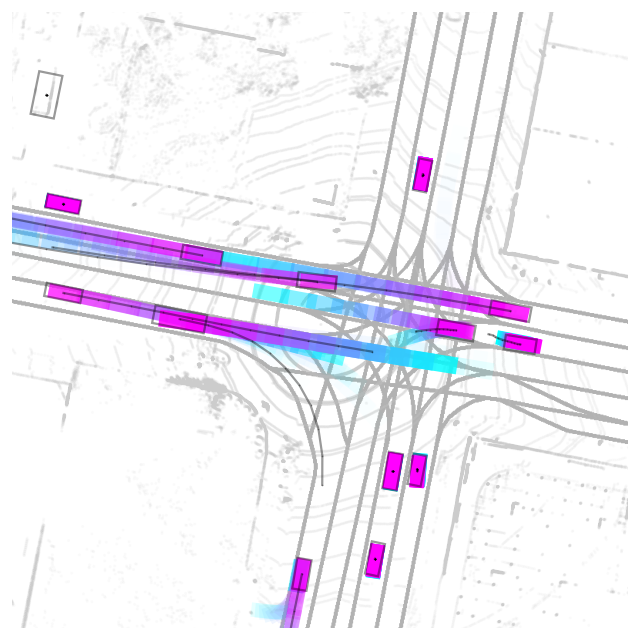}};
        \node[inner sep=0pt, outer sep=0, anchor=west]  (scene3detra)     at (scene3gorela.east) {\includegraphics[width=\imw]{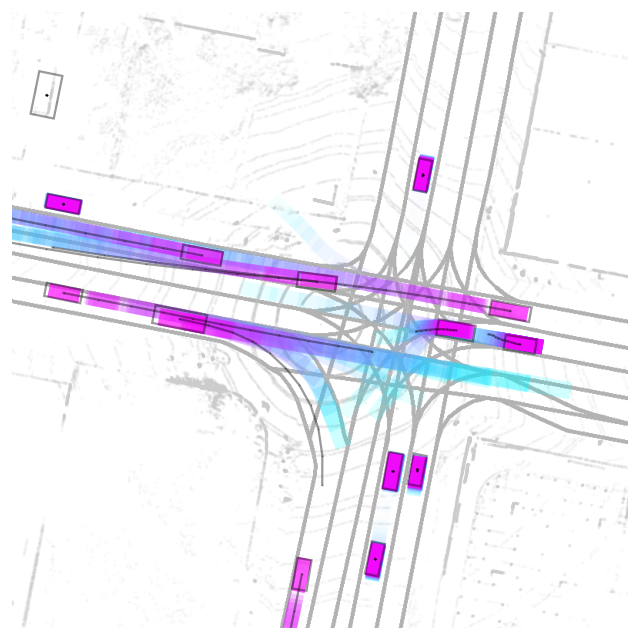}};

        \node at(scene1st)[draw, color=off-map,line width=\bw, minimum width=3mm, minimum height=12mm,yshift=+11mm,xshift=+10mm, rotate=55]{}; 
        \node at(scene1mtp)[draw, color=off-map,line width=\bw, minimum width=13mm, minimum height=5mm,yshift=+9mm,xshift=+8mm, rotate=0]{}; 
        \node at(scene1gorela)[draw, color=fn-mode,line width=\bw, minimum width=5mm, minimum height=5mm,yshift=8mm,xshift=13mm, rotate=0]{}; 

        \node at(scene1gorela)[draw, color=off-map,line width=\bw, minimum width=7mm, minimum height=4mm,yshift=-2mm,xshift=9mm, rotate=-40]{}; 
        \node at(scene1mtp)[draw, color=off-map,line width=\bw, minimum width=5mm, minimum height=5mm,yshift=-1mm,xshift=8mm, rotate=0]{}; 
        \node at(scene1st)[draw, color=off-map,line width=\bw, minimum width=7mm, minimum height=4mm,yshift=-1mm,xshift=9mm, rotate=-20]{}; 

        \node at(scene3gorela)[draw, color=fn-mode,line width=\bw, minimum width=4mm, minimum height=12mm,yshift=-3mm,xshift=-3mm, rotate=45]{}; 
        \node at(scene3mtp)[draw, color=fn-mode,line width=\bw, minimum width=4mm, minimum height=12mm,yshift=-3mm,xshift=-3mm, rotate=45]{}; 
        \node at(scene3st)[draw, color=fn-mode,line width=\bw, minimum width=4mm, minimum height=12mm,yshift=-3mm,xshift=-3mm, rotate=45]{}; 

        \node at(scene3gorela)[draw, color=fp-mode,line width=\bw, minimum width=8mm, minimum height=3mm,yshift=0mm,xshift=4mm, rotate=-8]{}; 
        \node at(scene3mtp)[draw, color=fp-mode,line width=\bw,    minimum width=6mm, minimum height=3mm,yshift=0mm,xshift=5mm, rotate=-8]{}; 
        \node at(scene3st)[draw, color=fp-mode,line width=\bw,     minimum width=8mm, minimum height=3mm,yshift=0mm,xshift=4mm, rotate=-8]{}; 

        \node at(scene2gorela)[draw, color=detection-color,line width=\bw, minimum width=4mm, minimum height=4mm,yshift=7mm,xshift=13mm, rotate=0]{}; 
        \node at(scene2st)[draw, color=detection-color,line width=\bw, minimum width=3mm, minimum height=3mm,yshift=7mm,xshift=13.5mm, rotate=0]{}; 

        \node at(scene2gorela)[draw, color=detection-color,line width=\bw, minimum width=3mm, minimum height=3mm,yshift=7mm,xshift=-13.5mm, rotate=0]{}; 
        \node at(scene2st)[draw, color=detection-color,line width=\bw,     minimum width=3mm, minimum height=3mm,yshift=7mm,xshift=-13.5mm, rotate=0]{}; 
        \node at(scene2mtp)[draw, color=detection-color,line width=\bw,    minimum width=3mm, minimum height=3mm,yshift=7mm,xshift=-13.5mm, rotate=0]{}; 

        \node at(scene2gorela)[draw, color=off-map,line width=\bw, minimum width=15mm, minimum height=4mm,yshift=1mm,xshift=-7mm, rotate=-35]{}; 
        \node at(scene2mtp)[draw, color=off-map,line width=\bw, minimum width=15mm, minimum height=4mm,yshift=0mm,xshift=-4mm, rotate=-35]{}; 
        \node at(scene2st)[draw, color=off-map,line width=\bw, minimum width=15mm, minimum height=4mm,yshift=0mm,xshift=-4mm, rotate=-35]{}; 

        \node at(scene2gorela)[draw, color=off-map,line width=\bw, minimum width=6mm, minimum height=4mm,yshift=12mm,xshift=-12mm, rotate=-35]{}; 
        \node at(scene2mtp)[draw, color=off-map,line width=\bw,    minimum width=6mm, minimum height=4mm,yshift=12mm,xshift=-12mm, rotate=-35]{}; 

    \end{tikzpicture}
    \caption{
        Qualitative results on AV2. 
        We highlight failure modes in the baselines that \ourmodel{} improves: \textcolor{detection-color}{inaccurate detections}, \textcolor{off-map}{off-map predictions}, \textcolor{fn-mode}{FN modes} and \textcolor{fp-mode}{FP modes}.
    }
    \label{fig:baseline-qualitative}    
\end{figure*}

\paragraphc{Baselines} 
We benchmark against the following methods.
\begin{itemize}
    \item Modular detection-tracking-forecasting baselines. We choose~\cite{pixor} as the detector architecture and modify it to predict the position at the previous time step. Following \cite{yang2021auto4d,qi2021offboard,yang2023labelformer}, we utilize a simple online tracker to associate detections over time leveraging the predicted previous position, and refine the tracked trajectory. For trajectory forecasting, we use MultiPath~\cite{multipath}, LaneGCN~\cite{lanegcn}, Scene Transformer~\cite{scenetransformer}, and GoRela~\cite{cui2022gorela}. The forecasting models ingest the (noisy) object tracks from the tracker and the map, and output the future trajectories. 
    \item End-to-end (E2E) detection and forecasting. We implement FaF~\cite{faf} and IntentNet~\cite{intentnet}, in which object bounding boxes and future trajectories are generated in one shot from a single convolutional network~\cite{pixor}. We also extend state-of-the-art forecasting approaches MultiPath~\cite{multipath}, LaneGCN~\cite{lanegcn}, and GoRela~\cite{cui2022gorela} by predicting object bounding boxes from the LiDAR backbone features and then cropping the feature maps with the bounding boxes to generate high-dimensional object embeddings, which inputs to the forecasting part of the end-to-end architecture.
\end{itemize}
Since the LiDAR encoder is not the focus of our work, yet it influences the detection and forecasting quality, we use the same LiDAR encoder architecture in all models (including ours) to ensure a fair comparison.

\paragraphc{Implementation Details} 
We utilize $N=400$ and $N=600$ object queries in AV2 and WOD, respectively. We use more queries in WOD to account for the larger RoI. 
\ourmodel{} uses $B=3$ refinement transformer blocks. The dimensions of all embeddings are 128 ($d=128$). We use 4-nearest map tokens ($k=4$) for map attention.
For the weights in our multi-task objective, we use $\alpha=0.1$, $\beta=0.01$, $\gamma=0.1$.
We train with a batch size of 16 scenes. 
We use AdamW~\cite{adamw} optimizer with a learning rate cosine decay starting at $8\times10^{-4}$ and ending at $0$, and weight decay of $10^{-4}$.
We train for 240K iterations on both datasets.
See the supplementary for more implementation details.

\paragraphc{Comparison against state-of-the-art} \Cref{tab:av2-comparison,tab:wod-comparison} present  detection and forecasting results on AV2 and WOD. Our model outperforms all baselines in all metrics on both datasets.  
On the proposed joint detection and forecasting metrics OccAP and TrajAP, \ourmodel{} outperforms the runner-up method consistently on both datasets, with a relative improvement of $9.6\%$ and $13.4\%$ in AV2, and $8.5\%$ and $13.5\%$ in WOD. Note that the runner-up method varies across datasets and metrics, but the gains over the best baseline on any metric are notable.
On detection metrics, we see substantial improvements ($3.9$ A.P. points in AV2 and $2.0$ A.P. points in WOD) at high IoU (0.7), without sacrificing any performance at low IoU (0.3), showing the power of our refinement transformer for detection. We recall that all baselines use the same LiDAR backbone architecture as \ourmodel{}, which isolates these gains to our proposed refinement transformer.
On forecasting, our model achieves a relative improvement in MissRate with $K=6$ over the best-performing baseline of $27\%$ and $8.2\%$ in AV2 and WOD, respectively. \ourmodel{} also achieves the best metrics when considering only the most likely future, $K=1$.
\Cref{fig:baseline-qualitative} highlights qualitative improvements over the strongest baselines.

\begin{table*}[t]
	\centering
    \fontsize{7.5pt}{8.5pt}\selectfont
    \begin{tabularx}{\textwidth}{l|ss|sss|sssssssss}
        \toprule
                                                      & \multicolumn{2}{c|}{Joint}         & \multicolumn{3}{c|}{Detection}                            & \multicolumn{7}{c}{Forecasting}                                                                                                                                                     \\
                                                         \cmidrule{2-3}         \cmidrule(l{2pt}r{2pt}){4-6}                               \cmidrule(l{2pt}){7-13}
                                                         & \multicolumn{2}{c|}{AP \ua} & \multicolumn{3}{c|}{AP @ IoU \ua}                & \multicolumn{2}{c}{MR \da}  & \multicolumn{2}{c}{ADE \da} &  \multicolumn{2}{c}{FDE \da}            & \multicolumn{1}{c}{bFDE \da}                                          \\  
                                                         \cmidrule(l{2pt}r{2pt}){2-3}                               \cmidrule(l{2pt}r{2pt}){4-6}                               \cmidrule(l{2pt}r{2pt}){7-8} \cmidrule(l{2pt}r{2pt}){9-10} \cmidrule(l{2pt}r{2pt}){11-12}                                      \cmidrule(l{2pt}r{2pt}){13-13}
                                                       & Occ                             & Traj                    & @0.3             & @0.5             & @0.7                    & $K$ = 1          & $K$ = 6          & $K$ = 1          & $K$ = 6          & $K$ = 1          & $K$ = 6                 & $K$ = 6          \\
        \midrule
        \ourmodel{} $i=0$                                              &42.7 &38.7 &95.3 &91.8 &76.7 &44.7 &44.7 &7.39 &7.39 &14.8 &14.8 &15.5 \\
        \ourmodel{} $i=1$                                              & 71.1 &48.0 &95.7 &92.5 &79.2 &38.2 &17.5 &1.67 &0.61 &3.94 &1.34 &2.01 \\
        \ourmodel{} $i=2$                                              & 72.8 &\textbf{50.2} &95.8 &92.8 &80.1 &37.4 &15.7 &1.54 &0.57 &\textbf{3.58} &1.21 &1.88 \\
        \ourmodel{} $i=3$                                  & \textbf{73.0} &50.0 &\textbf{95.8} &\textbf{92.9} &\textbf{80.2} &\textbf{37.0} &\textbf{15.4} &\textbf{1.54} &\textbf{0.55} &3.59 &\textbf{1.19} &\textbf{1.86} \\
        \bottomrule 
    \end{tabularx}
    \vspace{1mm}
	\caption{
        \ourmodel{}'s self-improvement over refinement blocks on AV2.
        Each row evaluates the intermediate predictions at the end of the $i$-th block ($i=0$ being pose initialization).
    }
	\label{tab:intermediate}
\end{table*}

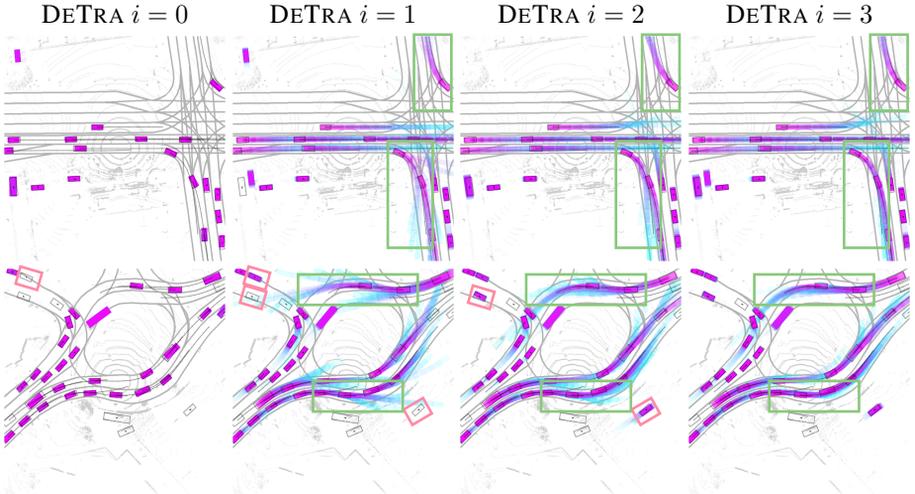
\begin{figure*}[t]
    \centering
    \definecolor{detection-color}{RGB}{255, 138, 165}
    \definecolor{traj-acc-color}{RGB}{138, 200, 124} 
    \begin{tikzpicture}
        \pgfmathsetlengthmacro{\imw}{0.25\textwidth}
        \pgfmathsetlengthmacro{\bw}{1.0pt}

        \node[inner sep=0pt, outer sep=0, anchor=west]  (scene1lvl0)    at (0, 0)                {\includegraphics[width=\imw]{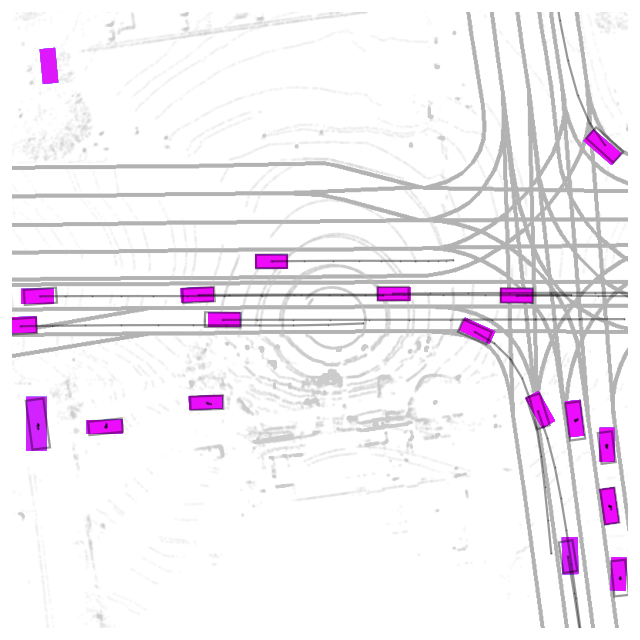}};
        \node[inner sep=0pt, outer sep=0, anchor=west]  (scene1lvl1)    at (scene1lvl0.east)     {\includegraphics[width=\imw]{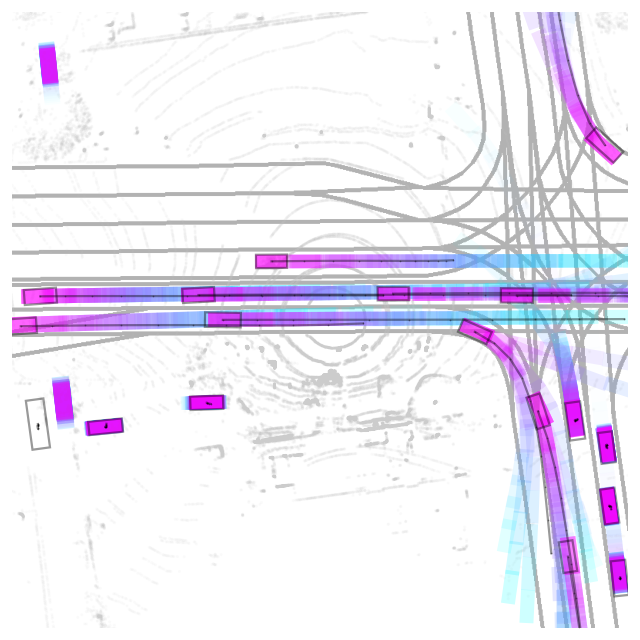}};
        \node[inner sep=0pt, outer sep=0, anchor=west]  (scene1lvl2)    at (scene1lvl1.east)     {\includegraphics[width=\imw]{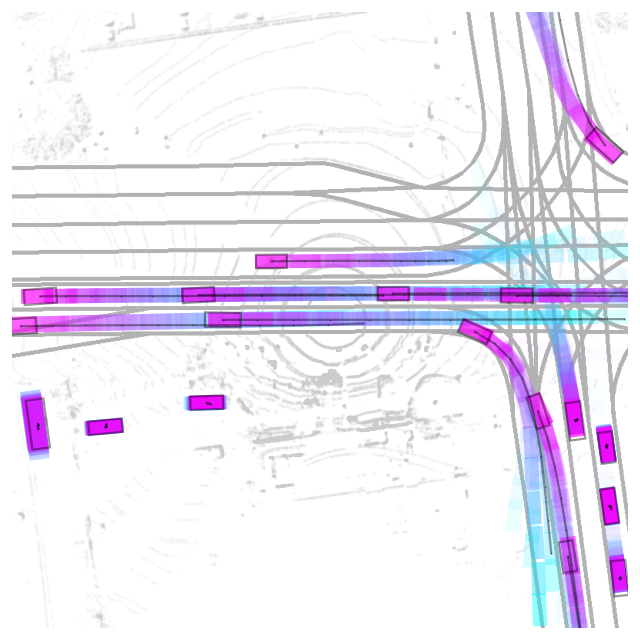}};
        \node[inner sep=0pt, outer sep=0, anchor=west]  (scene1lvl3)    at (scene1lvl2.east)     {\includegraphics[width=\imw]{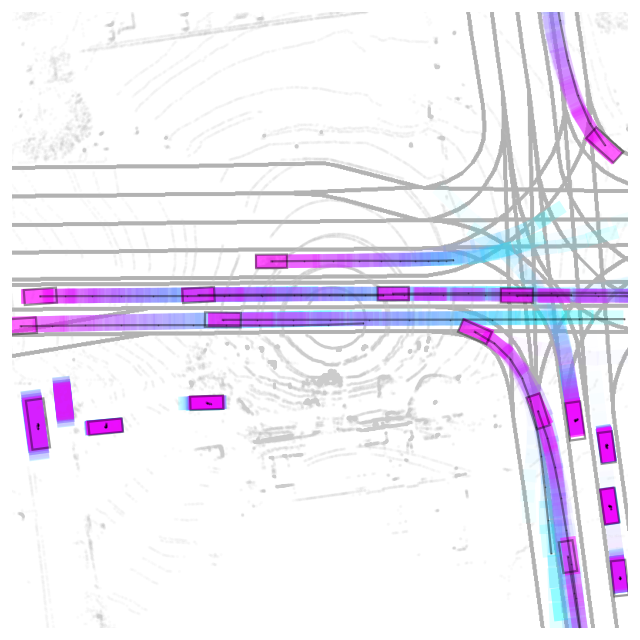}};
        \node[anchor=south] at (scene1lvl0.north) {\ourmodel{} $i=0$};
        \node[anchor=south] at (scene1lvl1.north) {\ourmodel{} $i=1$};
        \node[anchor=south] at (scene1lvl2.north) {\ourmodel{} $i=2$};
        \node[anchor=south] at (scene1lvl3.north) {\ourmodel{} $i=3$};

        \node[inner sep=0pt, outer sep=0, anchor=north]  (scene3lvl0)    at (scene1lvl0.south)    {\includegraphics[width=\imw]{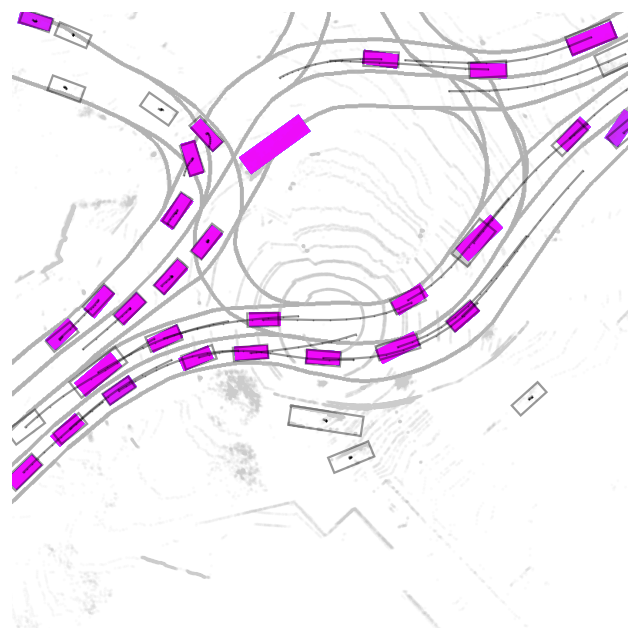}};
        \node[inner sep=0pt, outer sep=0, anchor=west]   (scene3lvl1)    at (scene3lvl0.east)     {\includegraphics[width=\imw]{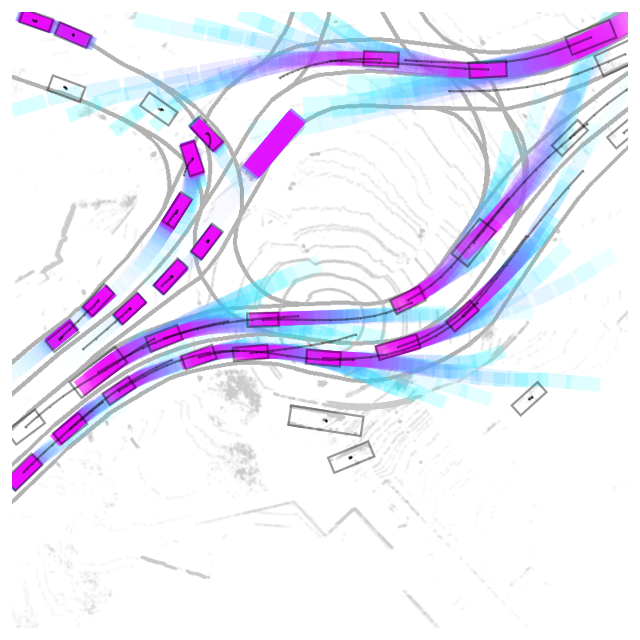}};
        \node[inner sep=0pt, outer sep=0, anchor=west]   (scene3lvl2)    at (scene3lvl1.east)     {\includegraphics[width=\imw]{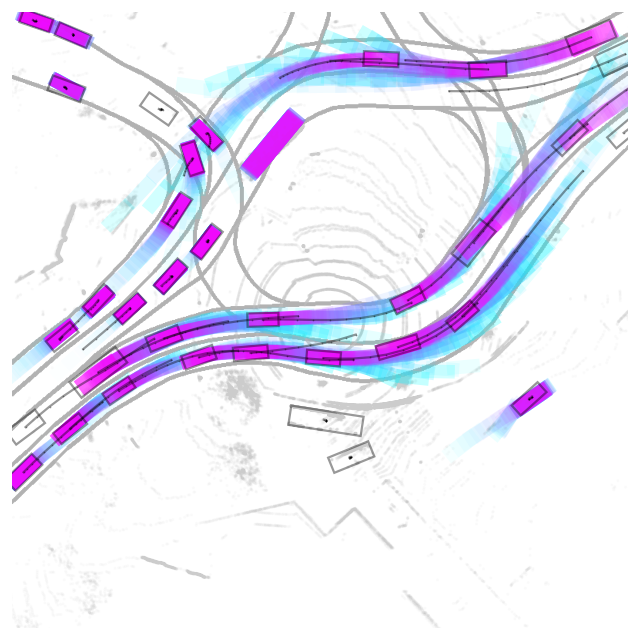}};
        \node[inner sep=0pt, outer sep=0, anchor=west]   (scene3lvl3)    at (scene3lvl2.east)     {\includegraphics[width=\imw]{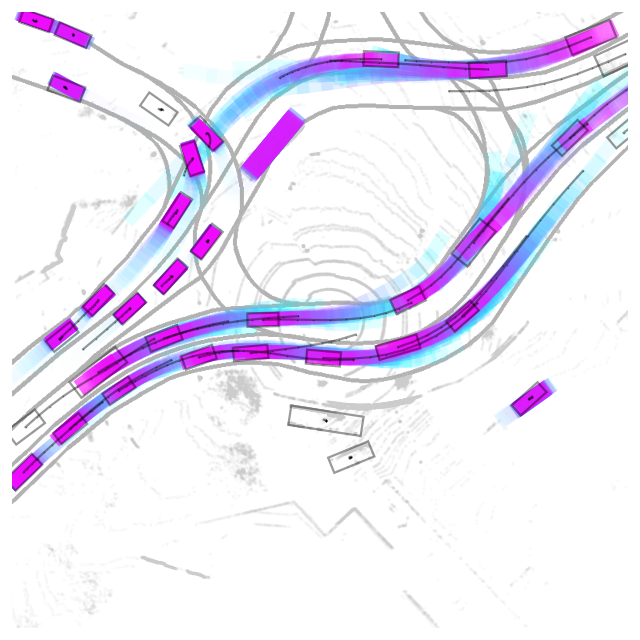}};

        \node at(scene1lvl3)[draw, color=traj-acc-color,line width=\bw, minimum width=5mm, minimum height=10mm,yshift=+10mm,xshift=+12mm]{}; 
        \node at(scene1lvl2)[draw, color=traj-acc-color,line width=\bw, minimum width=5mm, minimum height=10mm,yshift=+10mm,xshift=+12mm]{}; 
        \node at(scene1lvl1)[draw, color=traj-acc-color,line width=\bw, minimum width=5mm, minimum height=10mm,yshift=+10mm,xshift=+12mm]{}; 

        \node at(scene1lvl3)[draw, color=traj-acc-color,line width=\bw, minimum width=6mm, minimum height=14mm,yshift=-6mm,xshift=+9mm]{}; 
        \node at(scene1lvl2)[draw, color=traj-acc-color,line width=\bw, minimum width=6mm, minimum height=14mm,yshift=-6mm,xshift=+9mm]{}; 
        \node at(scene1lvl1)[draw, color=traj-acc-color,line width=\bw, minimum width=6mm, minimum height=14mm,yshift=-6mm,xshift=+9mm]{};

        \node at(scene3lvl1)[draw, color=detection-color,line width=\bw, minimum width=3mm, minimum height=2mm,yshift=13.5mm,xshift=-11.5mm, rotate=-16]{}; 
        \node at(scene3lvl0)[draw, color=detection-color,line width=\bw, minimum width=3mm, minimum height=2mm,yshift=13.5mm,xshift=-11.5mm, rotate=-16]{}; 

        \node at(scene3lvl1)[draw, color=detection-color,line width=\bw, minimum width=3mm, minimum height=2mm,yshift=11mm,xshift=-12mm, rotate=-16]{}; 
        \node at(scene3lvl2)[draw, color=detection-color,line width=\bw, minimum width=3mm, minimum height=2mm,yshift=11mm,xshift=-12mm, rotate=-16]{}; 

        \node at(scene3lvl1)[draw, color=detection-color,line width=\bw, minimum width=3mm, minimum height=2mm,yshift=-4mm,xshift=10mm, rotate=30]{}; 
        \node at(scene3lvl2)[draw, color=detection-color,line width=\bw, minimum width=3mm, minimum height=2mm,yshift=-4mm,xshift=10mm, rotate=30]{}; 

        \node at(scene3lvl3)[draw, color=traj-acc-color,line width=\bw, minimum width=16mm, minimum height=4mm,yshift=12mm,xshift=2mm, rotate=0]{}; 
        \node at(scene3lvl2)[draw, color=traj-acc-color,line width=\bw, minimum width=16mm, minimum height=4mm,yshift=12mm,xshift=2mm, rotate=0]{}; 
        \node at(scene3lvl1)[draw, color=traj-acc-color,line width=\bw, minimum width=16mm, minimum height=4mm,yshift=12mm,xshift=2mm, rotate=0]{}; 

        \node at(scene3lvl3)[draw, color=traj-acc-color,line width=\bw, minimum width=12mm, minimum height=4mm,yshift=-2mm,xshift=2mm, rotate=0]{}; 
        \node at(scene3lvl2)[draw, color=traj-acc-color,line width=\bw, minimum width=12mm, minimum height=4mm,yshift=-2mm,xshift=2mm, rotate=0]{}; 
        \node at(scene3lvl1)[draw, color=traj-acc-color,line width=\bw, minimum width=12mm, minimum height=4mm,yshift=-2mm,xshift=2mm, rotate=0]{}; 

    \end{tikzpicture}
    \caption{Visualizing the \ourmodel{}'s self improvement over refinement blocks on AV2. We highlight \textcolor{traj-acc-color}{improved forecasting accuracy} and \textcolor{detection-color}{improved detection accuracy} cases.}
    \label{fig:intermediate-qualitative}    
\end{figure*}

\paragraphc{Self-improvement across refinement blocks} 
\Cref{tab:intermediate} and \Cref{fig:intermediate-qualitative} shows how \ourmodel{}'s predictions ---both detections and forecasts--- improve over multiple rounds of our refinement transformer. 
Pose initialization ($ i=0 $) has a reasonably high detection performance at low IoUs (AP@IoU0.3) but lacks precise localization of the bounding boxes, as shown by the lower AP@IoU0.7. Forecasting results are weak, as expected since we assume objects to be stationary at initialization. After the first refinement block ($i=1$), detection AP@IoU0.7 is notably boosted (+2.5 points), and forecasting starts predicting the correct future motion (much lower ADEs and FDEs). Yet, it still lacks map understanding as shown by the qualitative results and Miss Rate. The second refinement block brings further detection gains, and the forecasts better follow the lanes. Finally, while the third refinement block still provides some improvements, the returns are diminished.
An attractive by-product of framing detection and forecasting as a unified trajectory refinement model is the ability to use it as an anytime predictor \cite{grubb2012speedboost,hu2019learning,kumbhar2020anytime}, i.e., a network with multiple exits where early outputs can be used if there is a need to reduce the latency/reaction time in specific situations.

\begin{table*}[t]
	\centering
    \fontsize{6.3pt}{7pt}\selectfont
    \begin{tabularx}{\textwidth}{l|ccccccc|ss|sss|sssssssss}
        \toprule
        & \multicolumn{7}{c|}{Components}                                                                                                 &   \multicolumn{2}{c|}{Joint}         & \multicolumn{3}{c|}{Detection}                            & \multicolumn{7}{c}{Forecasting}                                                                                                                                                     \\
          \cmidrule(l{2pt}r{2pt}){2-8}                                                                                                       \cmidrule{9-10}         \cmidrule(l{2pt}r{2pt}){11-13}                               \cmidrule(l{2pt}){14-20}
                      &  \multicolumn{2}{c}{Pose}         & \multicolumn{2}{c}{Cross-attn}           & \multicolumn{3}{c|}{Self-attn}               & \multicolumn{2}{c|}{AP \ua} & \multicolumn{3}{c|}{AP @ IoU \ua}                & \multicolumn{2}{c}{MR \da}  & \multicolumn{2}{c}{ADE \da} &  \multicolumn{2}{c}{FDE \da}            & \multicolumn{1}{c}{bFDE \da}                                          \\  
                      \cmidrule(l{2pt}r{2pt}){2-3} \cmidrule(l{2pt}r{2pt}){4-5} \cmidrule(l{2pt}r{2pt}){6-8}  \cmidrule(l{2pt}r{2pt}){9-10}              \cmidrule(l{2pt}r{2pt}){11-13} \cmidrule(l{2pt}r{2pt}){14-15} \cmidrule(l{2pt}r{2pt}){16-17}              \cmidrule(l{2pt}r{2pt}){18-19} \cmidrule(l{2pt}r{2pt}){20-20}   
                      &  Init       & Rfn         &  Ldr          & Map           & Time          & Mode          & Obj               & Occ              & Traj          & @0.3             & @0.5             & @0.7                    & $K$=1          & $K$=6          & $K$=1          & $K$=6          & $K$=1          & $K$=6                  & $K$=6          \\
        \midrule
        $V_1$         & G          & \checkmark   & \checkmark    & \checkmark    & \checkmark    &  \checkmark   & \checkmark        & 70.8          &46.8          &94.6          &91.4          &78.2            &37.7          &16.1          &1.61          &0.56          &3.79                &1.20          &1.87 \\
        $V_2$         & L          & \checkmark   & \checkmark    & \checkmark    & \checkmark    &  \checkmark   & \checkmark        & 68.5          &46.1          &92.6          &89.3          &76.0            &37.8          &16.5          &1.66          &0.57          &3.88                &1.21          &1.88 \\
        \midrule                     
        $V_{3}$       & \checkmark & $\times$     & \checkmark    & \checkmark    & \checkmark    &  \checkmark   & \checkmark        & 70.0          &44.6          &95.4          &91.5          &75.9            &39.8          &21.6          &1.87          &0.69          &4.31                &1.52          &2.18 \\
        \midrule
        $V_4$         & \checkmark & \checkmark   & $\times$      & \checkmark    & \checkmark    &  \checkmark   & \checkmark        & 42.5          &14.6          &70.0          &52.2          &35.6            & -            & -            & -            & -            & -                  & -            & -   \\
        $V_5$         & \checkmark & \checkmark   & Glb           & \checkmark    & \checkmark    &  \checkmark   & \checkmark        & 48.1          &16.6          &77.5          &52.9          &31.9            & -            & -            & -            & -            & -                  & -            & -   \\
        \midrule
        $V_6$         & \checkmark & \checkmark   & \checkmark    & $\times$      & \checkmark    &  \checkmark   & \checkmark        & 71.2          &47.9          &95.4          &92.5          &79.7            &38.7          &20.2          &1.71          &0.66          &4.01                &1.47          &2.14 \\
        $V_7$         & \checkmark & \checkmark   & \checkmark    & Glb           & \checkmark    &  \checkmark   & \checkmark        & 71.3          &46.9          &95.4          &92.6          &79.6            &39.5          &20.3          &1.80          &0.65          &4.25                &1.46          &2.13 \\
        \midrule
        $V_8$         & \checkmark & \checkmark   & \checkmark    & \checkmark    & $\times$      &  \checkmark   & \checkmark        & 62.4          &40.2          &95.4          &92.4          &79.5            &43.7          &31.3          &4.59          &1.01          &9.23                &2.06          &2.69 \\
        $V_9$         & \checkmark & \checkmark   & \checkmark    & \checkmark    & \checkmark    &  $\times$     & \checkmark        & \textbf{73.1} &49.7          &95.7          &92.9          &\textbf{80.5}   &37.5          &16.0          &1.57          &0.55          &3.65                &1.18          &1.85 \\
        $V_{10}$      & \checkmark & \checkmark   & \checkmark    & \checkmark    & \checkmark    &  \checkmark   & $\times$          & 73.0          &50.0          &95.7          &92.7          &80.1            &37.6          &15.8          &1.61          &0.55          &3.77                &\textbf{1.17} &\textbf{1.84} \\
        \midrule                               
        \ourmodel{}   & \checkmark            & \checkmark   & \checkmark    & \checkmark    & \checkmark    &  \checkmark   & \checkmark        & 73.0          &\textbf{50.0} &\textbf{95.8} &\textbf{92.9} &80.2            &\textbf{37.0} &\textbf{15.4} &\textbf{1.54} &\textbf{0.55} &\textbf{3.59}       &1.19          &1.86 \\
        \bottomrule 
    \end{tabularx}
    \vspace{1mm}
	\caption{
        \ourmodel{} components ablation on AV2.
        $\checkmark$ means we use what is described in \Cref{sec:method}.
        $\times$ means we drop this component. 
        For pose initialization (Init) we also consider initial poses in a grid (G), and learned offsets on top (L).
        The acronym Rfn stands for refinement. For cross-attention, we consider global attention to all LiDAR/Map tokens (Glb) as an alternative.
    }
	\label{tab:component}
\end{table*}

\paragraphc{Effect of pose initialization}
We compare our proposed pose initialization using a lightweight detector head to a fixed grid (G) as well as grid + learnable offsets (L) in \Cref{tab:component} ($V_1$ and $V_2$ respectively). We use a $20 \times 20$ grid to have $N=400$ queries, the same number we consider in our approach.
For the learned offset experiment, we allow the gradients from the $i$-th block to flow back to $\mathbf{P}^{(i-1)}$. 
We see that (i) detection-based initialization is best by a fair margin and (ii) that learned offsets from a grid are not helpful (we hypothesize this is because there is not a clear responsibility split between these learned offsets and the ones predicted in the pose update).

\paragraphc{Effect of pose refinement}
We consider a version of the model where all the refinement transformer blocks receive the initial poses $\mathbf{P}^{(0)}$ as input ($V_3$ in \Cref{tab:component}).
Since \ourmodel{} and $V_3$ have the same number of parameters and supervision (intermediate outputs at $i < B$ are still supervised), this shows that refinement is a critical piece.

\paragraphc{Effect of LiDAR attention} 
In \Cref{tab:component}, we consider two alternatives to the deformable LiDAR attention proposed in \ourmodel: no attention to LiDAR ($V_4$) and global attention to all the tokens in the LiDAR feature map ($V_5$). Both alternatives perform poorly, with detection metrics well below those of the initial detection proposals (see \ourmodel{} $i=0$ in \Cref{tab:intermediate}). With no attention ($V_4$), the model seems to struggle preserving the original proposals through multiple transformer blocks.
Global attention ($V_5$) not only underperforms, but it is also computationally expensive.
We hypothesize it performs poorly because there are too many LiDAR tokens in the multi-resolution feature maps, which makes it very hard to find signal among the noise.
Forecasting metrics are not reported for these two variants as the detector does not reach the required 80\% target recall.
In summary, including the inductive bias that the important LiDAR features are around the object poses $\mathbf{P}^{(i)}$ via deformable attention is very helpful.

\paragraphc{Effect of Map attention} 
In \Cref{tab:component}, we consider no map attention ($V_6$) and global map attention ($V_7$) instead of \ourmodel{}'s proposed $k$-nearest neighbor attention. The results are similar to those observed in LiDAR: limiting the attention to a neighborhood seems critical. 
Without limiting attention to map tokens in a neighborhood, this component is not effective, as shown by the very similar results of $V_6$ and $V_7$.

\paragraphc{Effect of self-attention} 
$V_8$, $V_9$, and $V_{10}$ in \Cref{tab:component} drop time, mode, and object self-attention, respectively.
The results show that time attention has a large impact, which makes sense since \ourmodel{} only attends to LiDAR at $t=0$ to limit the computation, and the geometric and motion features extracted from it can only reach other time steps via time self-attention.
While detection and joint metrics do not show gains for mode and object attention, most forecasting metrics show that it is beneficial to include these.

\section{Conclusion}
\label{sec:conclusion}
In this paper, we introduced \ourmodel{}, a model to tackle detection and forecasting as a unified trajectory refinement task.
We design a refinement transformer architecture to enable self-improvement via cross attention to heterogeneous inputs as well as factorized self attention across time, mode and object dimension. 
Our experiments on two large-scale self-driving datasets show that our model outperforms state-of-the-art modular and end-to-end approaches. 
Importantly, extensive ablation studies back our design, showing that all the components contribute positively, our proposed pose refinement is critical, and key choices were made to aid learning with geometric priors.

\pagebreak

\bibliographystyle{splncs04}
\bibliography{main}

\begin{thebibliography}{10}
\providecommand{\url}[1]{\texttt{#1}}
\providecommand{\urlprefix}{URL }
\providecommand{\doi}[1]{https://doi.org/#1}

\bibitem{agro2023implicit}
Agro, B., Sykora, Q., Casas, S., Urtasun, R.: Implicit occupancy flow fields for perception and prediction in self-driving. In: CVPR (2023)

\bibitem{Cai2019CascadeRH}
Cai, Z., Vasconcelos, N.: Cascade r-cnn: High quality object detection and instance segmentation. PAMI  (2019)

\bibitem{detr}
Carion, N., Massa, F., Synnaeve, G., Usunier, N., Kirillov, A., Zagoruyko, S.: End-to-end object detection with transformers. In: ECCV (2020)

\bibitem{spagnn}
Casas, S., Gulino, C., Liao, R., Urtasun, R.: Spagnn: Spatially-aware graph neural networks for relational behavior forecasting from sensor data. In: ICRA (2020)

\bibitem{ilvm}
Casas, S., Gulino, C., Suo, S., Luo, K., Liao, R., Urtasun, R.: Implicit latent variable model for scene-consistent motion forecasting. In: ECCV (2020)

\bibitem{intentnet}
Casas, S., Luo, W., Urtasun, R.: Intentnet: Learning to predict intention from raw sensor data. In: CoRL (2018)

\bibitem{casas2021mp3}
Casas, S., Sadat, A., Urtasun, R.: Mp3: A unified model to map, perceive, predict and plan. In: CVPR (2021)

\bibitem{multipath}
Chai, Y., Sapp, B., Bansal, M., Anguelov, D.: Multipath: Multiple probabilistic anchor trajectory hypotheses for behavior prediction. arXiv preprint arXiv:1910.05449  (2019)

\bibitem{chen2020dynamic}
Chen, Y., Dai, X., Liu, M., Chen, D., Yuan, L., Liu, Z.: Dynamic convolution: Attention over convolution kernels. In: Proceedings of the IEEE/CVF conference on computer vision and pattern recognition. pp. 11030--11039 (2020)

\bibitem{chitta2022transfuser}
Chitta, K., Prakash, A., Jaeger, B., Yu, Z., Renz, K., Geiger, A.: Transfuser: Imitation with transformer-based sensor fusion for autonomous driving. PAMI  (2022)

\bibitem{cho2014learning}
Cho, K., Van~Merri{\"e}nboer, B., Gulcehre, C., Bahdanau, D., Bougares, F., Schwenk, H., Bengio, Y.: Learning phrase representations using rnn encoder-decoder for statistical machine translation. arXiv preprint arXiv:1406.1078  (2014)

\bibitem{cui2021lookout}
Cui, A., Casas, S., Sadat, A., Liao, R., Urtasun, R.: Lookout: Diverse multi-future prediction and planning for self-driving. In: ICCV (2021)

\bibitem{cui2022gorela}
Cui, A., Casas, S., Wong, K., Suo, S., Urtasun, R.: Gorela: Go relative for viewpoint-invariant motion forecasting. arXiv preprint arXiv:2211.02545  (2022)

\bibitem{Cui2018MultimodalTP}
Cui, H., Radosavljevic, V., Chou, F.C., Lin, T.H., Nguyen, T., Huang, T.K., Schneider, J.G., Djuric, N.: Multimodal trajectory predictions for autonomous driving using deep convolutional networks. 2019 ICRA pp. 2090--2096 (2018), \url{https://api.semanticscholar.org/CorpusID:52891221}

\bibitem{Deo2021MultimodalTP}
Deo, N., Wolff, E.M., Beijbom, O.: Multimodal trajectory prediction conditioned on lane-graph traversals. In: CoRL (2021)

\bibitem{multixnet}
Djuric, N., Cui, H., Su, Z., Wu, S., Wang, H., Chou, F.C., Martin, L.S., Feng, S., Hu, R., Xu, Y., Dayan, A., Zhang, S., Becker, B.C., Meyer, G.P., Vallespi-Gonzalez, C., Wellington, C.K.: Multixnet: Multiclass multistage multimodal motion prediction. IV  (2021)

\bibitem{ettinger2021large}
Ettinger, S., Cheng, S., Caine, B., Liu, C., Zhao, H., Pradhan, S., Chai, Y., Sapp, B., Qi, C.R., Zhou, Y., et~al.: Large scale interactive motion forecasting for autonomous driving: The waymo open motion dataset. In: ICCV (2021)

\bibitem{fan2018baidu}
Fan, H., Zhu, F., Liu, C., Zhang, L., Zhuang, L., Li, D., Zhu, W., Hu, J., Li, H., Kong, Q.: Baidu apollo em motion planner. arXiv preprint arXiv:1807.08048  (2018)

\bibitem{Gao2020VectorNetEH}
Gao, J., Sun, C., Zhao, H., Shen, Y., Anguelov, D., Li, C., Schmid, C.: Vectornet: Encoding hd maps and agent dynamics from vectorized representation. CVPR  (2020)

\bibitem{gilles2021thomas}
Gilles, T., Sabatini, S., Tsishkou, D., Stanciulescu, B., Moutarde, F.: Thomas: Trajectory heatmap output with learned multi-agent sampling. arXiv preprint arXiv:2110.06607  (2021)

\bibitem{girgis2021latent}
Girgis, R., Golemo, F., Codevilla, F., Weiss, M., D'Souza, J.A., Kahou, S.E., Heide, F., Pal, C.: Latent variable sequential set transformers for joint multi-agent motion prediction. arXiv preprint arXiv:2104.00563  (2021)

\bibitem{latent}
Girgis, R., Golemo, F., Codevilla, F., Weiss, M., D'Souza, J.A., Kahou, S.E., Heide, F., Pal, C.: Latent variable sequential set transformers for joint multi-agent motion prediction. arXiv preprint arXiv:2104.00563  (2021)

\bibitem{grubb2012speedboost}
Grubb, A., Bagnell, D.: Speedboost: Anytime prediction with uniform near-optimality. In: Artificial Intelligence and Statistics. pp. 458--466. PMLR (2012)

\bibitem{he2016deep}
He, K., Zhang, X., Ren, S., Sun, J.: Deep residual learning for image recognition. In: CVPR. pp. 770--778 (2016)

\bibitem{hu2019learning}
Hu, H., Dey, D., Hebert, M., Bagnell, J.A.: Learning anytime predictions in neural networks via adaptive loss balancing. In: Proceedings of the AAAI Conference on Artificial Intelligence (2019)

\bibitem{hu2018squeeze}
Hu, J., Shen, L., Sun, G.: Squeeze-and-excitation networks. In: Proceedings of the IEEE conference on computer vision and pattern recognition. pp. 7132--7141 (2018)

\bibitem{hu2023_uniad}
Hu, Y., Yang, J., Chen, L., Li, K., Sima, C., Zhu, X., Chai, S., Du, S., Lin, T., Wang, W., Lu, L., Jia, X., Liu, Q., Dai, J., Qiao, Y., Li, H.: Planning-oriented autonomous driving. In: CVPR (2023)

\bibitem{ivanovic2022propagating}
Ivanovic, B., Lin, Y., Shrivastava, S., Chakravarty, P., Pavone, M.: Propagating state uncertainty through trajectory forecasting. In: ICRA (2022)

\bibitem{kipf2016semi}
Kipf, T.N., Welling, M.: Semi-supervised classification with graph convolutional networks. arXiv preprint arXiv:1609.02907  (2016)

\bibitem{kumbhar2020anytime}
Kumbhar, O., Sizikova, E., Majaj, N., Pelli, D.G.: Anytime prediction as a model of human reaction time. arXiv preprint arXiv:2011.12859  (2020)

\bibitem{Lang2018PointPillarsFE}
Lang, A.H., Vora, S., Caesar, H., Zhou, L., Yang, J., Beijbom, O.: Pointpillars: Fast encoders for object detection from point clouds. CVPR  (2018)

\bibitem{interactiontransformer}
Li, L.L., Yang, B., Liang, M., Zeng, W., Ren, M., Segal, S., Urtasun, R.: End-to-end contextual perception and prediction with interaction transformer. IROS  (2020)

\bibitem{lanegcn}
Liang, M., Yang, B., Hu, R., Chen, Y., Liao, R., Feng, S., Urtasun, R.: Learning lane graph representations for motion forecasting. In: ECCV. Springer (2020)

\bibitem{pnpnet}
Liang, M., Yang, B., Zeng, W., Chen, Y., Hu, R., Casas, S., Urtasun, R.: Pnpnet: End-to-end perception and prediction with tracking in the loop. In: CVPR (2020)

\bibitem{lin2017feature}
Lin, T.Y., Doll{\'a}r, P., Girshick, R., He, K., Hariharan, B., Belongie, S.: Feature pyramid networks for object detection. In: Proceedings of the IEEE conference on computer vision and pattern recognition. pp. 2117--2125 (2017)

\bibitem{focalloss}
Lin, T.Y., Goyal, P., Girshick, R., He, K., Doll{\'a}r, P.: Focal loss for dense object detection. In: ICCV (2017)

\bibitem{liu2022dabdetr}
Liu, S., Li, F., Zhang, H., Yang, X., Qi, X., Su, H., Zhu, J., Zhang, L.: {DAB}-{DETR}: Dynamic anchor boxes are better queries for {DETR}. In: ICLR (2022)

\bibitem{Liu2015SSDSS}
Liu, W., Anguelov, D., Erhan, D., Szegedy, C., Reed, S.E., Fu, C.Y., Berg, A.C.: Ssd: Single shot multibox detector. In: ECCV (2015)

\bibitem{adamw}
Loshchilov, I., Hutter, F.: Decoupled weight decay regularization. arXiv preprint arXiv:1711.05101  (2017)

\bibitem{faf}
Luo, W., Yang, B., Urtasun, R.: Fast and furious: Real time end-to-end 3d detection, tracking and motion forecasting with a single convolutional net. In: CVPR (2018)

\bibitem{mahjourian2022occupancy}
Mahjourian, R., Kim, J., Chai, Y., Tan, M., Sapp, B., Anguelov, D.: Occupancy flow fields for motion forecasting in autonomous driving. IEEE Robotics and Automation Letters  \textbf{7}(2),  5639--5646 (2022)

\bibitem{laserflow}
Meyer, G.P., Charland, J., Pandey, S., Laddha, A.G., Gautam, S., Vallespi-Gonzalez, C., Wellington, C.K.: Laserflow: Efficient and probabilistic object detection and motion forecasting. IEEE Robotics and Automation Letters  \textbf{6},  526--533 (2021)

\bibitem{Mo2022MultiAgentTP}
Mo, X., Huang, Z., Xing, Y., Lv, C.: Multi-agent trajectory prediction with heterogeneous edge-enhanced graph attention network. IEEE Transactions on Intelligent Transportation Systems  (2022)

\bibitem{wayformer}
Nayakanti, N., Al-Rfou, R., Zhou, A., Goel, K., Refaat, K.S., Sapp, B.: Wayformer: Motion forecasting via simple \& efficient attention networks. arXiv preprint arXiv:2207.05844  (2022)

\bibitem{scenetransformer}
Ngiam, J., Caine, B., Vasudevan, V., Zhang, Z., Chiang, H.T.L., Ling, J., Roelofs, R., Bewley, A., Liu, C., Venugopal, A., et~al.: Scene transformer: A unified architecture for predicting multiple agent trajectories. arXiv preprint arXiv:2106.08417  (2021)

\bibitem{PhanMinh2019CoverNetMB}
Phan-Minh, T., Grigore, E.C., Boulton, F.A., Beijbom, O., Wolff, E.M.: Covernet: Multimodal behavior prediction using trajectory sets. CVPR  (2019)

\bibitem{qi2017pointnet}
Qi, C.R., Su, H., Mo, K., Guibas, L.J.: Pointnet: Deep learning on point sets for 3d classification and segmentation. In: CVPR (2017)

\bibitem{qi2021offboard}
Qi, C.R., Zhou, Y., Najibi, M., Sun, P., Vo, K., Deng, B., Anguelov, D.: Offboard 3d object detection from point cloud sequences. In: CVPR (2021)

\bibitem{Redmon2015YouOL}
Redmon, J., Divvala, S.K., Girshick, R.B., Farhadi, A.: You only look once: Unified, real-time object detection. 2016 IEEE Conference on Computer Vision and Pattern Recognition (CVPR)  (2015)

\bibitem{Ren2015FasterRT}
Ren, S., He, K., Girshick, R.B., Sun, J.: Faster r-cnn: Towards real-time object detection with region proposal networks. PAMI  (2015)

\bibitem{rhinehart2018deep}
Rhinehart, N., McAllister, R., Levine, S.: Deep imitative models for flexible inference, planning, and control. arXiv preprint arXiv:1810.06544  (2018)

\bibitem{sadat2019jointly}
Sadat, A., Ren, M., Pokrovsky, A., Lin, Y.C., Yumer, E., Urtasun, R.: Jointly learnable behavior and trajectory planning for self-driving vehicles. In: 2019 IROS. pp. 3949--3956. IEEE (2019)

\bibitem{LiRaNet}
Shah, M., ling Huang, Z., Laddha, A.G., Langford, M., Barber, B., Zhang, S., Vallespi-Gonzalez, C., Urtasun, R.: Liranet: End-to-end trajectory prediction using spatio-temporal radar fusion. In: CoRL (2020)

\bibitem{waymo}
Sun, P., Kretzschmar, H., Dotiwalla, X., Chouard, A., Patnaik, V., Tsui, P., Guo, J., Zhou, Y., Chai, Y., Caine, B., et~al.: Scalability in perception for autonomous driving: Waymo open dataset. In: CVPR (2020)

\bibitem{waymo2020motion}
Sun, P., Kretzschmar, H., Dotiwalla, X., Chouard, A., Patnaik, V., Tsui, P., Guo, J., Zhou, Y., Chai, Y., Caine, B., et~al.: The waymo open motion dataset. \url{https://waymo.com/open/motion/} (2020)

\bibitem{varadarajan2022multipath++}
Varadarajan, B., Hefny, A., Srivastava, A., Refaat, K.S., Nayakanti, N., Cornman, A., Chen, K., Douillard, B., Lam, C.P., Anguelov, D., et~al.: Multipath++: Efficient information fusion and trajectory aggregation for behavior prediction. In: 2022 ICRA (2022)

\bibitem{transformer}
Vaswani, A., Shazeer, N., Parmar, N., Uszkoreit, J., Jones, L., Gomez, A.N., Kaiser, {\L}., Polosukhin, I.: Attention is all you need. NeurIPS  (2017)

\bibitem{weng2022mtp}
Weng, X., Ivanovic, B., Pavone, M.: Mtp: Multi-hypothesis tracking and prediction for reduced error propagation. In: IV (2022)

\bibitem{argoverse2}
Wilson, B., Qi, W., Agarwal, T., Lambert, J., Singh, J., Khandelwal, S., Pan, B., Kumar, R., Hartnett, A., Pontes, J.K., et~al.: Argoverse 2: Next generation datasets for self-driving perception and forecasting. arXiv preprint arXiv:2301.00493  (2023)

\bibitem{yang2023labelformer}
Yang, A.J., Casas, S., Dvornik, N., Segal, S., Xiong, Y., Hu, J.S.K., Fang, C., Urtasun, R.: Labelformer: Object trajectory refinement for offboard perception from lidar point clouds. In: CoRL. PMLR (2023)

\bibitem{yang2021auto4d}
Yang, B., Bai, M., Liang, M., Zeng, W., Urtasun, R.: Auto4d: Learning to label 4d objects from sequential point clouds. arXiv preprint arXiv:2101.06586  (2021)

\bibitem{pixor}
Yang, B., Luo, W., Urtasun, R.: Pixor: Real-time 3d object detection from point clouds. In: CVPR (2018)

\bibitem{agentformer}
Yuan, Y., Weng, X., Ou, Y., Kitani, K.M.: Agentformer: Agent-aware transformers for socio-temporal multi-agent forecasting. In: ICCV (2021)

\bibitem{zeng2019end}
Zeng, W., Luo, W., Suo, S., Sadat, A., Yang, B., Casas, S., Urtasun, R.: End-to-end interpretable neural motion planner. In: CVPR (2019)

\bibitem{zhang2023towards}
Zhang, L., Yang, A.J., Xiong, Y., Casas, S., Yang, B., Ren, M., Urtasun, R.: Towards unsupervised object detection from lidar point clouds. In: Proceedings of the IEEE/CVF Conference on Computer Vision and Pattern Recognition. pp. 9317--9328 (2023)

\bibitem{Zhou2019ObjectsAP}
Zhou, X., Wang, D., Kr{\"a}henb{\"u}hl, P.: Objects as points. ArXiv  (2019)

\bibitem{Zhou2019EndtoEndMF}
Zhou, Y., Sun, P., Zhang, Y., Anguelov, D., Gao, J., Ouyang, T.Y., Guo, J., Ngiam, J., Vasudevan, V.: End-to-end multi-view fusion for 3d object detection in lidar point clouds. ArXiv  (2019)

\bibitem{Zhou2017VoxelNetEL}
Zhou, Y., Tuzel, O.: Voxelnet: End-to-end learning for point cloud based 3d object detection. 2018 IEEE/CVF Conference on Computer Vision and Pattern Recognition  (2017)

\bibitem{zhou2023query}
Zhou, Z., Wang, J., Li, Y.H., Huang, Y.K.: Query-centric trajectory prediction. In: CVPR (2023)

\bibitem{Zhou2022HiVTHV}
Zhou, Z., Ye, L., Wang, J., Wu, K., Lu, K.: Hivt: Hierarchical vector transformer for multi-agent motion prediction. CVPR  (2022)

\bibitem{deformabledetr}
Zhu, X., Su, W., Lu, L., Li, B., Wang, X., Dai, J.: Deformable detr: Deformable transformers for end-to-end object detection. arXiv preprint arXiv:2010.04159  (2020)

\bibitem{zhu2020deformable}
Zhu, X., Su, W., Lu, L., Li, B., Wang, X., Dai, J.: Deformable detr: Deformable transformers for end-to-end object detection. arXiv preprint arXiv:2010.04159  (2020)

\end{thebibliography}


\begin{thebibliography}{10}
\providecommand{\url}[1]{\texttt{#1}}
\providecommand{\urlprefix}{URL }
\providecommand{\doi}[1]{https://doi.org/#1}

\bibitem{ilvm}
Casas, S., Gulino, C., Suo, S., Luo, K., Liao, R., Urtasun, R.: Implicit latent
  variable model for scene-consistent motion forecasting. In: ECCV (2020)

\bibitem{chen2020dynamic}
Chen, Y., Dai, X., Liu, M., Chen, D., Yuan, L., Liu, Z.: Dynamic convolution:
  Attention over convolution kernels. In: Proceedings of the IEEE/CVF
  conference on computer vision and pattern recognition. pp. 11030--11039
  (2020)

\bibitem{cui2022gorela}
Cui, A., Casas, S., Wong, K., Suo, S., Urtasun, R.: Gorela: Go relative for
  viewpoint-invariant motion forecasting. arXiv preprint arXiv:2211.02545
  (2022)

\bibitem{hu2018squeeze}
Hu, J., Shen, L., Sun, G.: Squeeze-and-excitation networks. In: Proceedings of
  the IEEE conference on computer vision and pattern recognition. pp.
  7132--7141 (2018)

\bibitem{lanegcn}
Liang, M., Yang, B., Hu, R., Chen, Y., Liao, R., Feng, S., Urtasun, R.:
  Learning lane graph representations for motion forecasting. In: ECCV.
  Springer (2020)

\bibitem{pnpnet}
Liang, M., Yang, B., Zeng, W., Chen, Y., Hu, R., Casas, S., Urtasun, R.:
  Pnpnet: End-to-end perception and prediction with tracking in the loop. In:
  CVPR (2020)

\bibitem{lin2017feature}
Lin, T.Y., Doll{\'a}r, P., Girshick, R., He, K., Hariharan, B., Belongie, S.:
  Feature pyramid networks for object detection. In: Proceedings of the IEEE
  conference on computer vision and pattern recognition. pp. 2117--2125 (2017)

\bibitem{scenetransformer}
Ngiam, J., Caine, B., Vasudevan, V., Zhang, Z., Chiang, H.T.L., Ling, J.,
  Roelofs, R., Bewley, A., Liu, C., Venugopal, A., et~al.: Scene transformer: A
  unified architecture for predicting multiple agent trajectories. arXiv
  preprint arXiv:2106.08417  (2021)

\bibitem{waymo2020motion}
Sun, P., Kretzschmar, H., Dotiwalla, X., Chouard, A., Patnaik, V., Tsui, P.,
  Guo, J., Zhou, Y., Chai, Y., Caine, B., et~al.: The waymo open motion
  dataset. \url{https://waymo.com/open/motion/} (2020)

\bibitem{zhang2023towards}
Zhang, L., Yang, A.J., Xiong, Y., Casas, S., Yang, B., Ren, M., Urtasun, R.:
  Towards unsupervised object detection from lidar point clouds. In:
  Proceedings of the IEEE/CVF Conference on Computer Vision and Pattern
  Recognition. pp. 9317--9328 (2023)

\bibitem{zhu2020deformable}
Zhu, X., Su, W., Lu, L., Li, B., Wang, X., Dai, J.: Deformable detr: Deformable
  transformers for end-to-end object detection. arXiv preprint arXiv:2010.04159
   (2020)

\end{thebibliography}

\clearpage
\appendix
{\noindent \Large \textbf{Supplementary Material} \vspace{0.15cm}} \\

\appendix

This appendix describes implementation details, including architecture and training, baselines, and metrics in \Cref{sec:impl-details-supp}. 

Additionally, we perform additional ablation studies in \Cref{sec:supp-ablations} to understand the effect of different numbers of transformer blocks, different orderings of the attention modules of \ourmodel{}, and varying the mode and time dimensions of the learnable query volume.

\Cref{sec:qual-wod} provides visualizations of \ourmodel{} compared to the baselines on the Waymo Open Dataset (WOD).

Finally, \Cref{sec:limitations} describes the limitations of our work and opportunities for future work.

\section{Implementation Details} \label{sec:impl-details-supp}

\subsection{LiDAR Encoder} \label{sec:lidar-encoder}
\begin{figure*}[h]
    \centering
    \includegraphics[width=\textwidth]{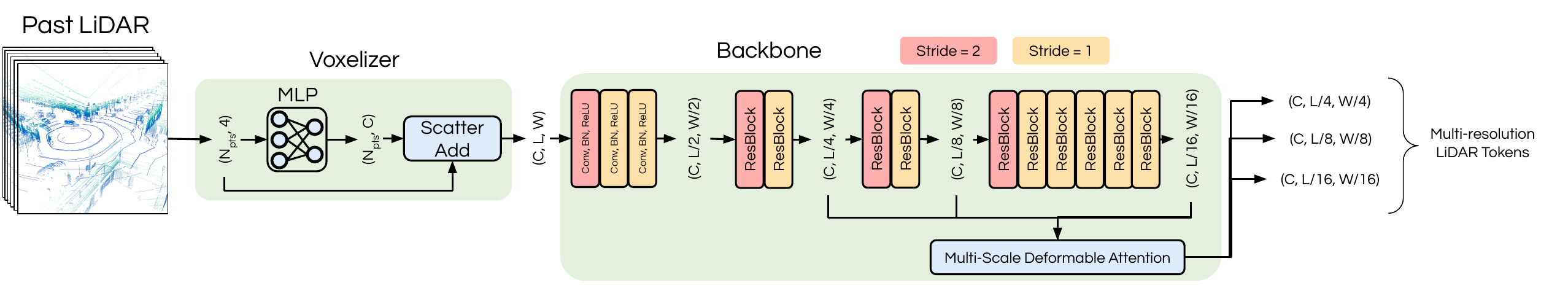}
    \caption{
        The architecture of the LiDAR Encoder. The batch dimension is omitted from the tensor shapes. 
        $N_{pts}$ is the number of points, 
        $C$ is the number of channels, 
        and $L$ and $W$ are the spatial dimensions in BEV.
        The four features of each point are $(x, y, z, t)$.
    }
    \label{fig:encoder}
\end{figure*}

The LiDAR encoder takes $H = 5$ past LiDAR sweeps to produce multi-resolution Birds Eye View (BEV) feature maps 
that serve as the LiDAR tokens for the refinement transformer.
Below, we describe the architecture of the LiDAR encoder, illustrated in \cref{fig:encoder}, where we omit the batch dimension
for simplicity. %

\paragraphc{Voxelization}
Each LiDAR sweep is a set of points with four features $(x, y, z, t)$. 
A small MLP is used to encode these points as feature vectors of size $C = 128$.
The LiDAR points are placed in a 2D BEV grid of shape $(L, W)$ based on their $(x, y)$ coordinates with a simple sum aggregator.
We use an ROI of $[-40, 40]\;\text{m}$ on the $x$ and $y$ dimensions for AV2, and $[-80, 80]\;\text{m}$ on the $x$ and $y$ dimensions for WOD.
We use a voxel size of $\SI{0.1}{m}$, resulting in $L = \frac{80}{0.1} = 800, W = \frac{80}{0.1} = 800$ for AV2, and $L = \frac{160}{0.1} = 1600, W = \frac{160}{0.1} = 1600$ for WOD.
A 2D feature map of shape $(C, L, W)$ is then generated from this BEV grid, where each grid cell has a feature vector
that is the sum of the features from all points in that grid cell.

\paragraphc{Backbone}
This 2D feature map is processed by three convolution layers interleaved with batch-normalization and ReLU layers,
The first convolutional layer has a stride of 2, resulting in a feature map of shape $(C, L/2, W/2)$.
Then, this feature map is processed by a series of ten residual layers, labeled \textit{ResBlock} in \cref{fig:encoder}, which each employ a sequence of dynamic convolution \cite{chen2020dynamic}, batch-normalization, ReLU, dynamic convolution, batch-normalization, squeeze-and-excitation \cite{hu2018squeeze}, and dropout operations. Each residual layer produces a feature map, and layers 0, 2, and 4 down-sample their output by a factor of 2 (using convolutions with stride 2).
We extract three multi-level feature maps from the output of layers 1, 3, and 9 with shapes $(C, L/4, W/4)$, $(C, L/8, W/8)$, and $(C, L/16, W/16)$
respectively, across which information is fused with multiscale deformable attention \cite{zhu2020deformable} to
produce three feature maps with the same shapes as the input. 

\subsection{Map Encoder}
We follow~\cite{lanegcn, cui2022gorela} for lane graph generation and map encoding.
Specifically, our lane graph is constructed by selecting centerlines from HD maps at fixed intervals of \SI{3}{m}.
These centerlines are divided into lane segments, each represented by a node in the graph with features including length, width, curvature, speed limit, and lane boundary type (e.g., solid or dashed).
To establish connections between lane nodes, we employ four types of relationships: successors, predecessors, left neighbors, and right neighbors.
We leverage the graph neural network in~\cite{cui2022gorela} to encode the lane graph nodes with pair-wise relative positional encodings and use the map node embeddings as the map tokens.
 	
\subsection{Initial pose header}  \label{sec:detection-header}
To produce the initial poses $\mathbf{P}^{(0)}$, we use a small Feature Pyramid Network \cite{lin2017feature}
where the three multi-resolution feature maps are fused into a single feature map of shape $(C, L/4, W/4)$. %
Then, four convolution layers interleaved with batch-normalization and ReLU layers predict a heatmap
of bounding-box parameters, and a separate set of four convolutional layers predict a heatmap 
of detection scores. These heatmap outputs are turned into object detections via NMS with an IOU threshold of 0.1.

\subsection{Metrics}

\paragraphc{OccAP} 
To compute this metric, we do the following:
\begin{enumerate}
    \item Generate a ground truth spatio-temporal occupancy grid of shape $(T, L, W)$, where $T$ is the number of forecasted timesteps, and $L$ and $W$ are the spatial dimensions defined in \cref{sec:lidar-encoder}.
    Each grid cell is filled with 0.0 if it is unoccupied and 1.0 if it is occupied, using the ground truth future trajectories for each object to compute this occupancy.
    \item Generate a predicted spatio-temporal occupancy grid of shape $(T, L, W)$, where the values in the grid cells range from $[0.0, 1.0]$ and represent the probability that an object occupies that cell.
    To obtain these probabilities, we rasterize all $F$ modes of each detected object, where each mode has an occupancy probability equal to the product of its detection confidence and mode confidence.
    We deal with overlapping modes appropriately, setting the probability in the grid cell equal to 1 minus the probability that no mode occupies the cell.
    \item Compute the average precision (AP) of the predicted occupancy grid with respect to the ground truth occupancy grid. For 100 evenly spaced threshold values between 0 and 1, we
    compute the number of true positives, false positives, and false negatives and then compute the precision and recall at each threshold. We compute AP as the area under the precision-recall curve.
\end{enumerate}

\paragraphc{TrajAP}
This metric closely follows the Soft AP metric used in the Waymo Motion Prediction Challenge
\cite{waymo2020motion}\footnote{\href{https://waymo.com/open/challenges/2023/motion-prediction/}{https://waymo.com/open/challenges/2023/motion-prediction/}},
but we make the following modifications to adapt it to our detection-trajectory forecasting setting.
First, we threshold all detections at the threshold that attains their maximum $F_1$ score.
We consider a trajectory as a match only if its detection matches with a specific IOU score \textit{and} its forecasted trajectory matches
(using the same definition in the original Soft AP metric).
The final metric we report is a macro average across future time horizons of \SI{3}{s} and \SI{5}{s}, detection IOU thresholds
of 0.5 and 0.7, and static and dynamic actors.

To compute the metric on static (or dynamic) actors only, we set the dynamic (or static) actors to ``ignore",
and any detection or trajectory forecast that matches with an ignore label is removed (not counted).

\subsection{Baselines}

\paragraphc{Detection}
All baselines use the same LiDAR encoder as \ourmodel{} (see \cref{sec:lidar-encoder}) as part of their detection network.
Likewise, their detections are produced with the same header as our prior object detection header described in \cref{sec:detection-header}.
The detection loss is the same as our initial pose loss described in the main paper.

\paragraphc{Tracker}
To associate detections over time and create tracks as input to MultiPath, LaneGCN, SceneTF, and GoRela,
we use the online heuristics tracker described in \cite{zhang2023towards}\footnote{see
the supplementary here \href{https://arxiv.org/pdf/2311.02007.pdf}{https://arxiv.org/pdf/2311.02007.pdf}
for details}. We perform tracking in the forward direction so it can run online and provide
a fair comparison against the end-to-end methods, which also run online.

\paragraphc{Trajectory Forecasting}
We faithfully reproduce the trajectory forecasting models described in each method's paper. 
We can directly use the tracks as input for the modular detection-tracking-forecasting models because the
trajectory forecasting methods were developed to ingest tracks. For these models' End-to-end (E2E) versions, we replace per-actor track features
with per-actor LiDAR features. We applied a rotated ROI pool inside each detected bounding box on the BEV feature map output from the feature-pyramid network to obtain these features.

All methods use the same prediction loss as \ourmodel{}, described in the main paper:
a Laplacian winner-takes-all loss on the mixture of trajectories and a cross-entropy loss on the mode probabilities, where the winner mode is the one
with trajectory waypoints closest to the ground truth, measured with an L1 loss on the Laplacian centroids $(\mu_x, \mu_y)$.

\paragraphc{Training Details}
We train all methods with the same learning rate schedule, optimizer, and batch size as \ourmodel{},
with equal relative loss weighting between the heatmap detection loss and the prediction loss.

\section{Additional Quantitative Results} \label{sec:supp-ablations}

\paragraph{What is the effect of refinement depth? }

\Cref{tab:depth} shows the performance of \ourmodel{} with different numbers of refinement blocks.
We observe that the performance of \ourmodel{} improves from 1 to 3 refinement blocks but then
plateaus. We expect different training recipes, hyper-parameters, and amounts of data would
be necessary to scale \ourmodel{}, which we leave as room for future work.
\begin{table*}[h]
	\centering
    \fontsize{7.5pt}{8.5pt}\selectfont
    \begin{tabularx}{\textwidth}{l|ss|sss|sssssssss}
        \toprule
        \# of refinement blocks                    & \multicolumn{2}{c|}{Joint}         & \multicolumn{3}{c|}{Detection}                            & \multicolumn{7}{c}{Forecasting}                                                                                                                                                     \\
                                                       \cmidrule{2-3}         \cmidrule(l{2pt}r{2pt}){4-6}                               \cmidrule(l{2pt}){7-13}
                                                       & \multicolumn{2}{c|}{AP \ua} & \multicolumn{3}{c|}{AP @ IoU \ua}                & \multicolumn{2}{c}{MR \da}  & \multicolumn{2}{c}{ADE \da} &  \multicolumn{2}{c}{FDE \da}            & \multicolumn{1}{c}{bFDE \da}                                          \\  
                                                       \cmidrule(l{2pt}r{2pt}){2-3}                               \cmidrule(l{2pt}r{2pt}){4-6}                               \cmidrule(l{2pt}r{2pt}){7-8} \cmidrule(l{2pt}r{2pt}){9-10} \cmidrule(l{2pt}r{2pt}){11-12}                                      \cmidrule(l{2pt}r{2pt}){13-13}
                                                       & Occ                             & Traj                    & @0.3             & @0.5             & @0.7                    & $K$ = 1          & $K$ = 6          & $K$ = 1          & $K$ = 6          & $K$ = 1          & $K$ = 6                 & $K$ = 6          \\
        \midrule
        1                                         & 71.2 &46.7 &95.5 &92.3 &78.6 &38.7 &18.7 &1.76 &0.61 &4.07 &1.33 &1.99 \\
        2                                         & 72.5 &48.4 &95.8 &\textbf{93.1} &80.1 &37.7 &16.5 &1.62 &0.57 &3.77 &1.21 &1.88 \\
        3 (\ourmodel{})                           & \textbf{73.0} &\textbf{50.0} &\textbf{95.8} &92.9 &80.2 &\textbf{37.0} &15.4 &\textbf{1.54} &0.55 &\textbf{3.59} &1.19 &1.86 \\
        4                                         & 72.3 &49.8 &95.6 &92.8 &\textbf{80.8} &37.2 &\textbf{18.7} &1.55 &\textbf{0.54} &3.63 &\textbf{1.14} &\textbf{1.81} \\
        5                                         & 72.0 &47.8 &95.8 &92.8 &79.9 &37.8 &16.5 &1.6 &0.55 &3.72 &1.18 &1.84 \\
        \bottomrule 
    \end{tabularx}
    \vspace{1mm}
	\caption{
        \ourmodel{}'s performance using different numbers of refinement blocks on AV2. Note that unlike Table 3 in the main paper,
        where one version of \ourmodel{} was trained with 3 refinement blocks and evaluated after each refinement block,
        here we trained 5 different versions of \ourmodel{} and evaluated the final output.
    }
	\label{tab:depth}
\end{table*}

\paragraph{Does the ordering of the attention layers matter? }

\Cref{tab:attention-ordering} shows the performance of \ourmodel{} with different orderings of the attention layers.
The main finding from this is that it is important for the self-attention modules (time, mode, object) to come after the
cross-attention modules (lidar and map). This result makes sense intuitively because the self-attention modules are meant to
propagate sensor information (obtained in the cross-attention modules) across the query volume, and they
cannot do that if applied before the cross-attention modules.
Besides that, the order within cross-attention (lidar and map) or self-attention (time, future mode, object) matters little. However, we find the selected order of lidar, map, time, future, and object to be the best in most metrics.
\begin{table*}[h]
	\centering
    \fontsize{7.5pt}{8.5pt}\selectfont
    \begin{tabularx}{\textwidth}{l|ss|sss|sssssssss}
        \toprule
        Ordering                                         & \multicolumn{2}{c|}{Joint}         & \multicolumn{3}{c|}{Detection}                            & \multicolumn{7}{c}{Forecasting}                                                                                                                                                     \\
                                                         \cmidrule{2-3}         \cmidrule(l{2pt}r{2pt}){4-6}                               \cmidrule(l{2pt}){7-13}
                                                         & \multicolumn{2}{c|}{AP \ua} & \multicolumn{3}{c|}{AP @ IoU \ua}                & \multicolumn{2}{c}{MR \da}  & \multicolumn{2}{c}{ADE \da} &  \multicolumn{2}{c}{FDE \da}            & \multicolumn{1}{c}{bFDE \da}                                          \\  
                                                         \cmidrule(l{2pt}r{2pt}){2-3}                               \cmidrule(l{2pt}r{2pt}){4-6}                               \cmidrule(l{2pt}r{2pt}){7-8} \cmidrule(l{2pt}r{2pt}){9-10} \cmidrule(l{2pt}r{2pt}){11-12}                                      \cmidrule(l{2pt}r{2pt}){13-13}
                                                       & Occ                             & Traj                    & @0.3             & @0.5             & @0.7                    & $K$ = 1          & $K$ = 6          & $K$ = 1          & $K$ = 6          & $K$ = 1          & $K$ = 6                 & $K$ = 6          \\
        \midrule
        $L,M,O,T,F$                                         & 72.1 &49.9 &95.7 &92.8 &\textbf{80.5} &\textbf{36.9} &\textbf{15.2} &1.54 &0.56 &3.59 &1.19 &1.86 \\
        $T,F,O,L,M$                                         & 71.3 &47.6 &95.7 &92.7 &79.9 &38.1 &17.4 &1.65 &0.58 &3.84 &1.26 &1.92 \\
        $M,L,T,F,O$                                         & 72.7 &\textbf{50.5} &95.5 &92.8 &79.6 &37.1 &16.2 &1.55 &0.55 &3.62 &1.18 &1.85 \\
        \ourmodel{} ($L,M,T,F,O$)                           & \textbf{73.0} &50.0 &\textbf{95.8} &\textbf{92.9} &80.2 &37.0 &15.4 &\textbf{1.54} &\textbf{0.55} &\textbf{3.59} &\textbf{1.19} &1.86 \\
        \bottomrule 
    \end{tabularx}
    \vspace{1mm}
	\caption{
        \ourmodel{}'s performance using different orderings of the attention blocks on AV2.
        Here $L$, $M$, $T$, $F$, and $O$ stand for the \textit{LiDAR}, \textit{Map}, \textit{Time}, \textit{Mode}, and \textit{Object} attention modules, respectively.
    }
	\label{tab:attention-ordering}
\end{table*}

\paragraph{How does \ourmodel{} perform with different learnable query volume dimensions? }

\Cref{tab:query-volume} shows the performance of \ourmodel{} with different query volume dimensions.
Specifically, we experiment with varying the mode dimensionality $F \in \{1, 6\}$ and
time dimensionality $T \in \{1, 10\}$. Note that the output trajectories still have ten timesteps
and six modes regardless of the input query volume dimensions because we adjust the output dimensionality of the MLP in the pose update accordingly.
The time dimension $T = 10$ is crucial for good trajectory forecasting performance, e.g., see the OccAP and TrajAP metrics.
This result makes sense because gathering information from future time queries about the map and interactions with other objects is essential for trajectory forecasting.
The mode dimension $F = 6$ is also important for trajectory forecasting performance, particularly in the forecasting metrics at $K = 6$, which is expected because having the mode dimension from the beginning (as opposed to just as part of the MLP decoder) allows these modes to differentiate more from each other in meaningful ways, enabling learning of more diverse trajectory forecasts that cover multiple possible futures. Detection gets slightly degraded at high IoUs when using multiple modes or time steps, which shows that there may be room for improvement in the mode aggregation over query features and potentially using causal attention in the time dimension.
\begin{table*}[h]
	\centering
    \fontsize{7.5pt}{8.5pt}\selectfont
    \begin{tabularx}{\textwidth}{l|ss|sss|sssssssss}
        \toprule
        Query Volume Dims                              & \multicolumn{2}{c|}{Joint}         & \multicolumn{3}{c|}{Detection}                            & \multicolumn{7}{c}{Forecasting}                                                                                                                                                     \\
                                                         \cmidrule{2-3}         \cmidrule(l{2pt}r{2pt}){4-6}                               \cmidrule(l{2pt}){7-13}
                                                       & \multicolumn{2}{c|}{AP \ua} & \multicolumn{3}{c|}{AP @ IoU \ua}                & \multicolumn{2}{c}{MR \da}  & \multicolumn{2}{c}{ADE \da} &  \multicolumn{2}{c}{FDE \da}            & \multicolumn{1}{c}{bFDE \da}                                          \\  
                                                        \cmidrule(l{2pt}r{2pt}){2-3}                               \cmidrule(l{2pt}r{2pt}){4-6}                               \cmidrule(l{2pt}r{2pt}){7-8} \cmidrule(l{2pt}r{2pt}){9-10} \cmidrule(l{2pt}r{2pt}){11-12}                                      \cmidrule(l{2pt}r{2pt}){13-13}
                                                       & Occ                             & Traj                    & @0.3             & @0.5             & @0.7                    & $K$ = 1          & $K$ = 6          & $K$ = 1          & $K$ = 6          & $K$ = 1          & $K$ = 6                 & $K$ = 6          \\
        \midrule
        $T = 1, F =1$                                  & 50.1 &40.0 &\textbf{95.9} &\textbf{93.0} &\textbf{80.6} &41.8 &40.6 &2.7 &2.28 &5.76 &4.74 &5.1 \\
        $T = 10, F = 1$                                & 71.7 &45.5 &95.8 &92.7 &79.3 &39.3 &19.9 &1.79 &0.64 &4.16 &1.4 &2.05 \\
        $T = 1, F = 6$                                 & 67.1 &45.4 &95.5 &92.3 &78.4 &42.0 &22.9 &2.06 &0.92 &4.47 &2.26 &2.92 \\
        \ourmodel{} ($T = 10, F = 6$)                  & \textbf{73.0} &\textbf{50.0} &95.8 &92.9 &80.2 &\textbf{37.0} &\textbf{15.4} &\textbf{1.54} &\textbf{0.55} &\textbf{3.59} &\textbf{1.19} &\textbf{1.86} \\
        \bottomrule 
    \end{tabularx}
    \vspace{1mm}
	\caption{
        \ourmodel{}'s performance using different query volume dimensions on AV2. 
    }
	\label{tab:query-volume}
\end{table*}

\paragraph{How robust is DeTra to different hyperparameters? }

We have shown that a naive design for LiDAR and map attention (e.g., global attention) is prohibitively expensive and does not learn well. 
Even though our specialized attention mechanisms introduce new hyperparameters (number of LiDAR deformable attention heads $\ell$, and number of map neighbors $k$), \Cref{tab:hyperparam} shows that \ourmodel{} is fairly robust to the choice of the specific hyperparameters  across multiple datasets.
In fact, \ourmodel{} still achieves SOTA performance on both AV2 and WOD for multiple hyperparameters choices, highlighting its robustness to variation in these hyperparameters. 

\begin{table*}[h]
	\centering
    \fontsize{7.5pt}{8.5pt}\selectfont
    \begin{tabularx}{\linewidth}{l|ssss}
        \toprule
        Hyperparameters                          & \tiny AV2 OccAP        & \tiny AV2 TrajAP       & \tiny WOD OccAP        & \tiny WOD TrajAP             \\ \midrule
        $k = 4, \ell = 4$ \tiny{(proposed)}      & \textbf{73.0}    & 50.0             & 70.4             & 37.1             \\
        $k = 4, \ell = 2$                        & 71.6             & \textbf{50.0}    & \textbf{70.9}    & \textbf{38.0}    \\
        $k = 4, \ell = 8$                        & 72.2             & 47.4             & 69.9             & 36.7             \\
        $k = 2, \ell = 4$                        & 72.7             & 50.0             & 70.2             & 37.0             \\
        $k = 8, \ell = 4$                        & 72.8             & 48.6             & 70.7             & 36.9             \\ \bottomrule
        \end{tabularx}
    \vspace{1mm}
	\caption{
        \ourmodel{}'s performance using different hyperparameters for our specialized LiDAR and map attention mechanisms (number of LiDAR deformable attention heads $\ell$, and number of map neighbors $k$).
    }
	\label{tab:hyperparam}
\end{table*}

\section{Additional Qualitative Results} \label{sec:qual-wod}

\begin{figure*}[t]
    \centering
    \begin{tikzpicture}
        \pgfmathsetlengthmacro{\imw}{0.25\textwidth}
        \pgfmathsetlengthmacro{\bw}{1.0pt}

        \node[inner sep=0pt, outer sep=0, anchor=west]  (scene1st)        at (0, 0)                 {\includegraphics[width=\imw]{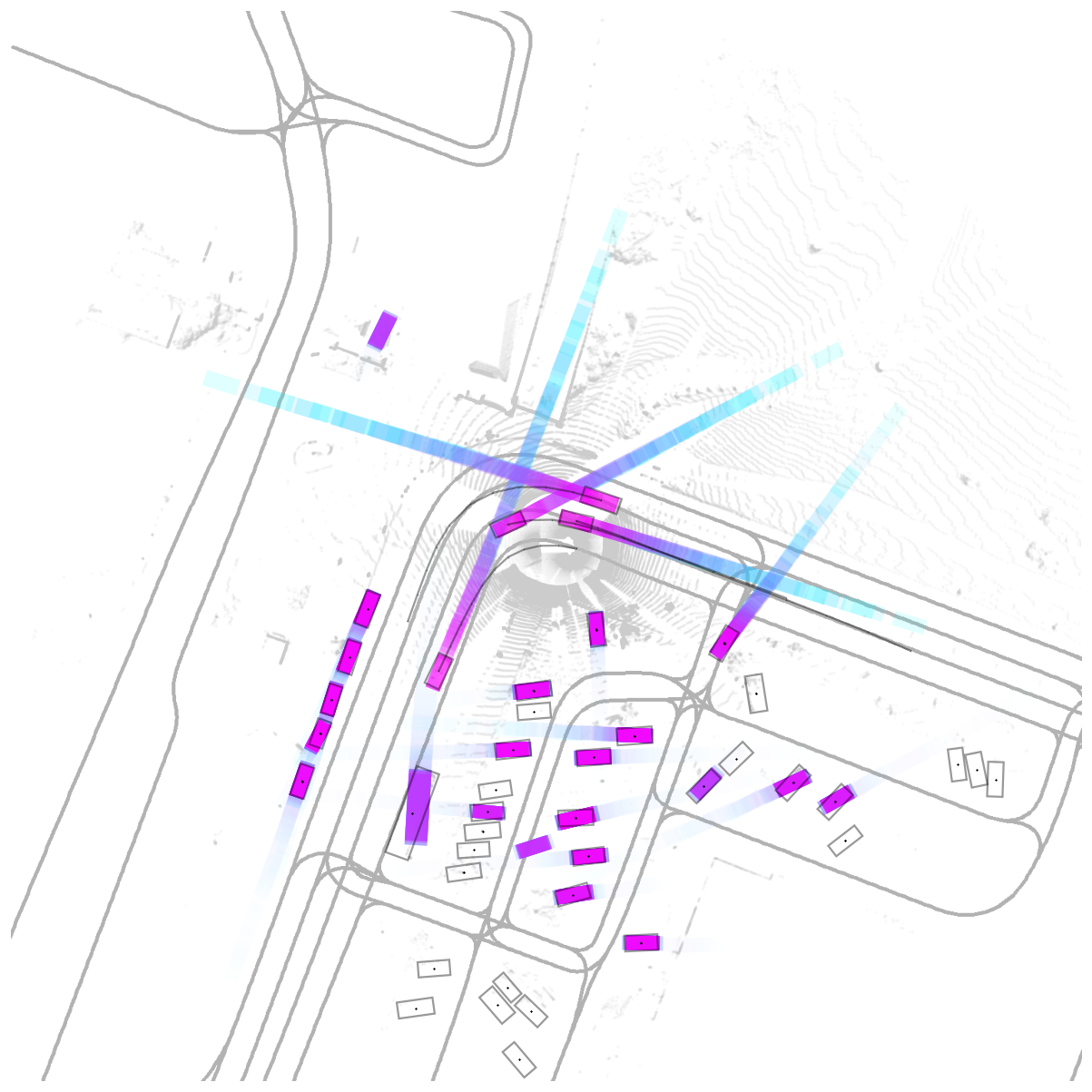}};
        \node[inner sep=0pt, outer sep=0, anchor=west]  (scene1mtp)       at (scene1st.east)        {\includegraphics[width=\imw]{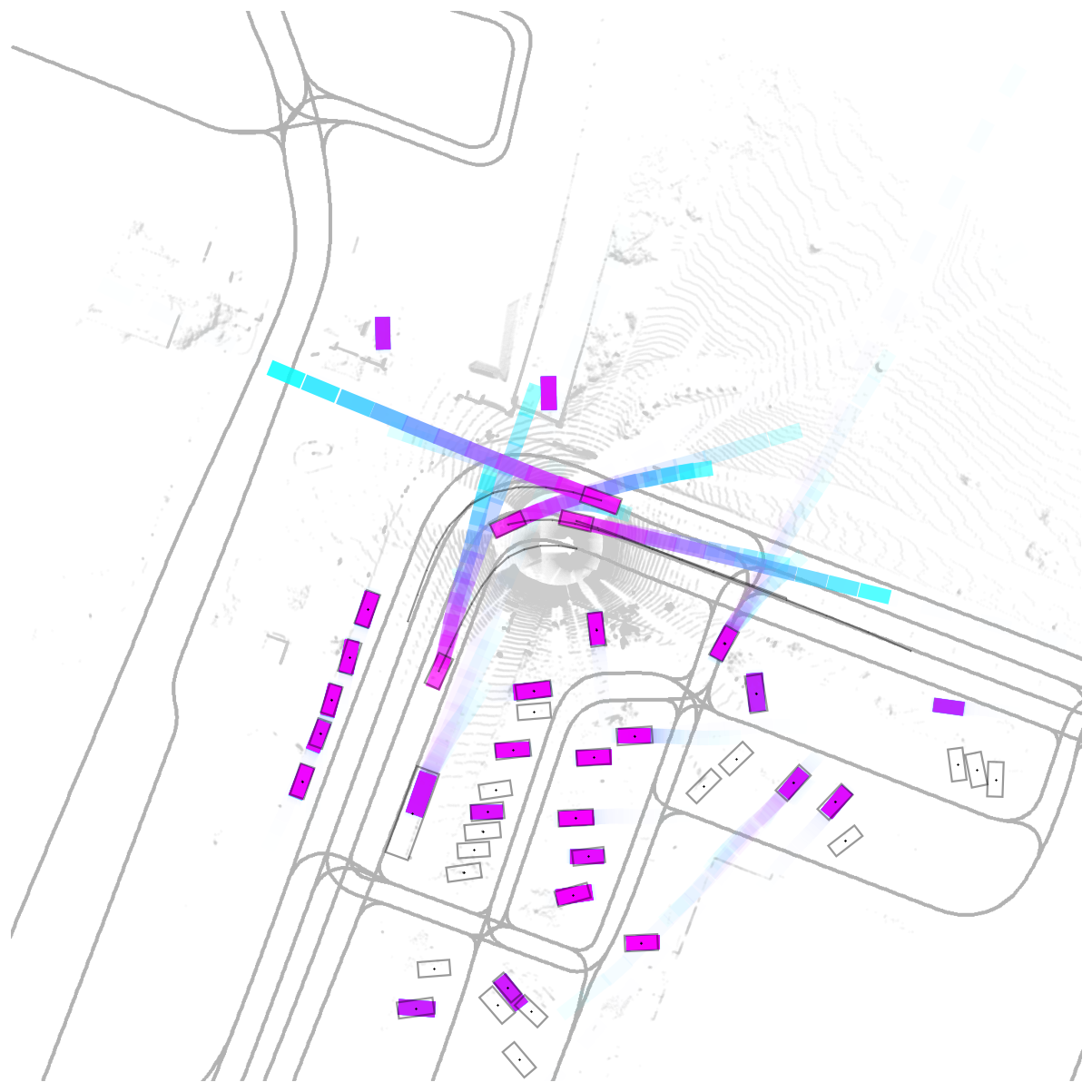}};
        \node[inner sep=0pt, outer sep=0, anchor=west]  (scene1gorela) at (scene1mtp.east)          {\includegraphics[width=\imw]{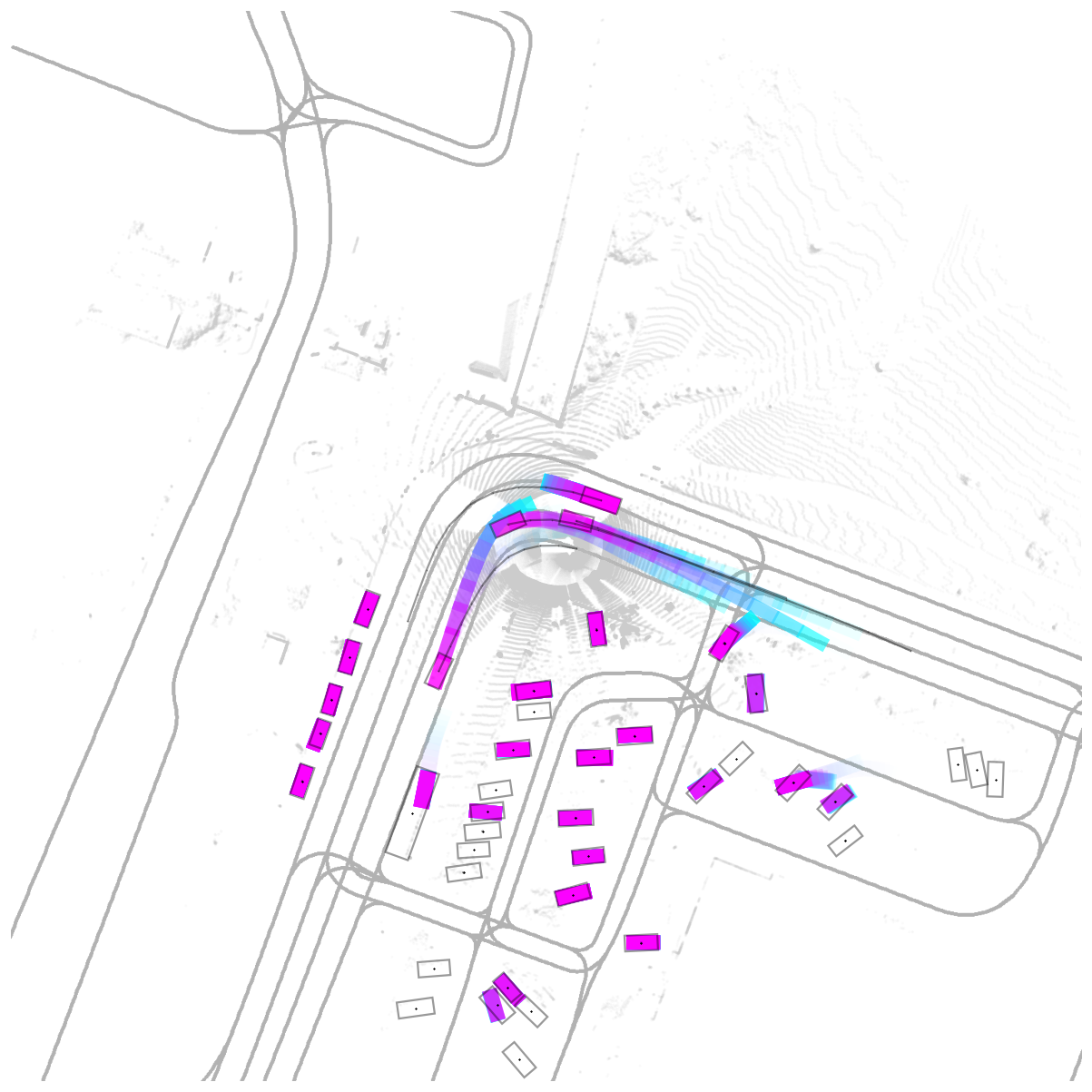}};
        \node[inner sep=0pt, outer sep=0, anchor=west]  (scene1detra)     at (scene1gorela.east)    {\includegraphics[width=\imw]{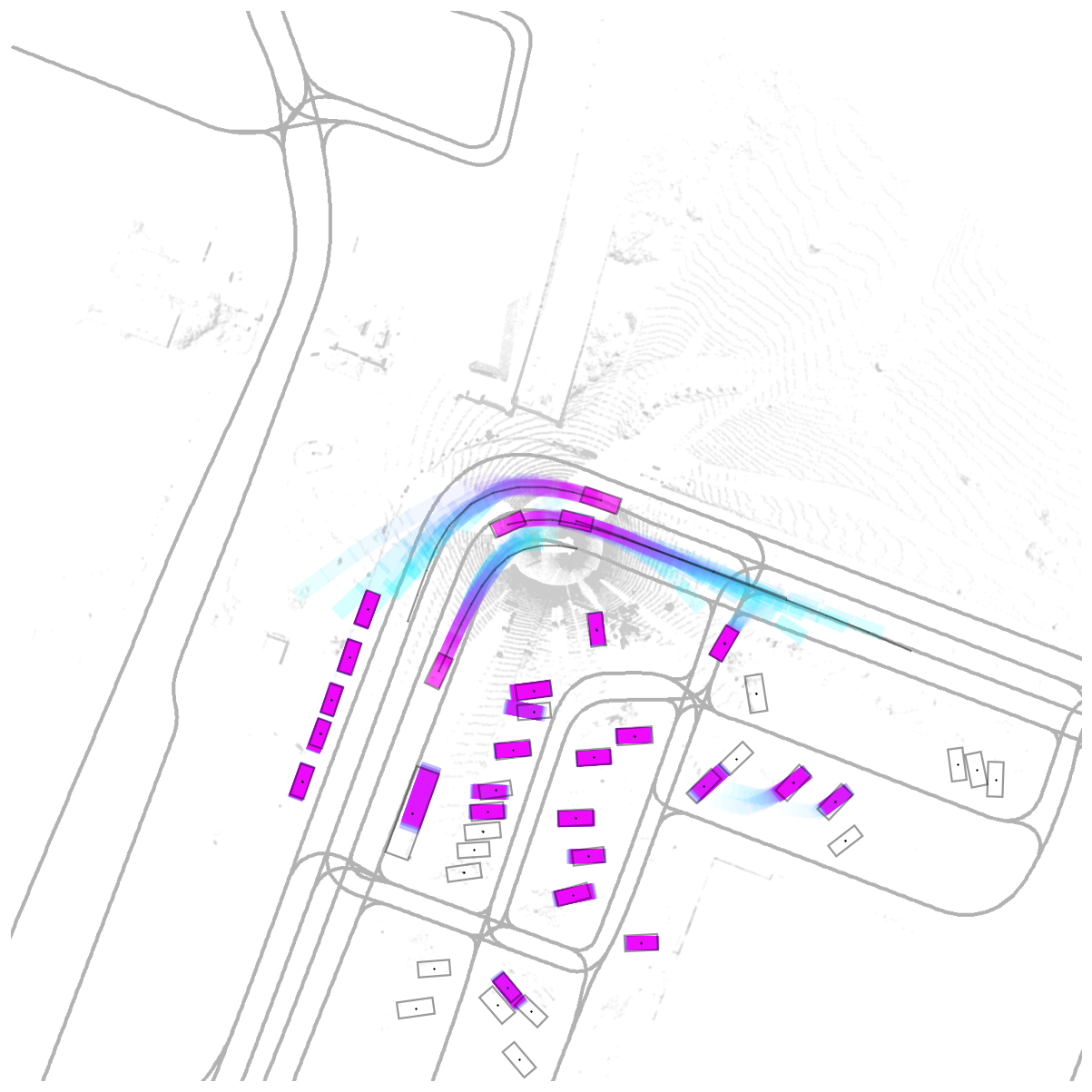}};
        \node[anchor=south] at (scene1st.north) {SceneTransformer};
        \node[anchor=south] at (scene1mtp.north) {MultiPath E2E};
        \node[anchor=south] at (scene1gorela.north) {GoRela E2E};
        \node[anchor=south] at (scene1detra.north) {\ourmodel{}};

        \node[anchor=north west] at (scene1detra.north east) {\includegraphics[height=0.75\textwidth]{images/legend.pdf}};

        \node[inner sep=0pt, outer sep=0, anchor=north] (scene2st)        at (scene1st.south)       {\includegraphics[width=\imw]{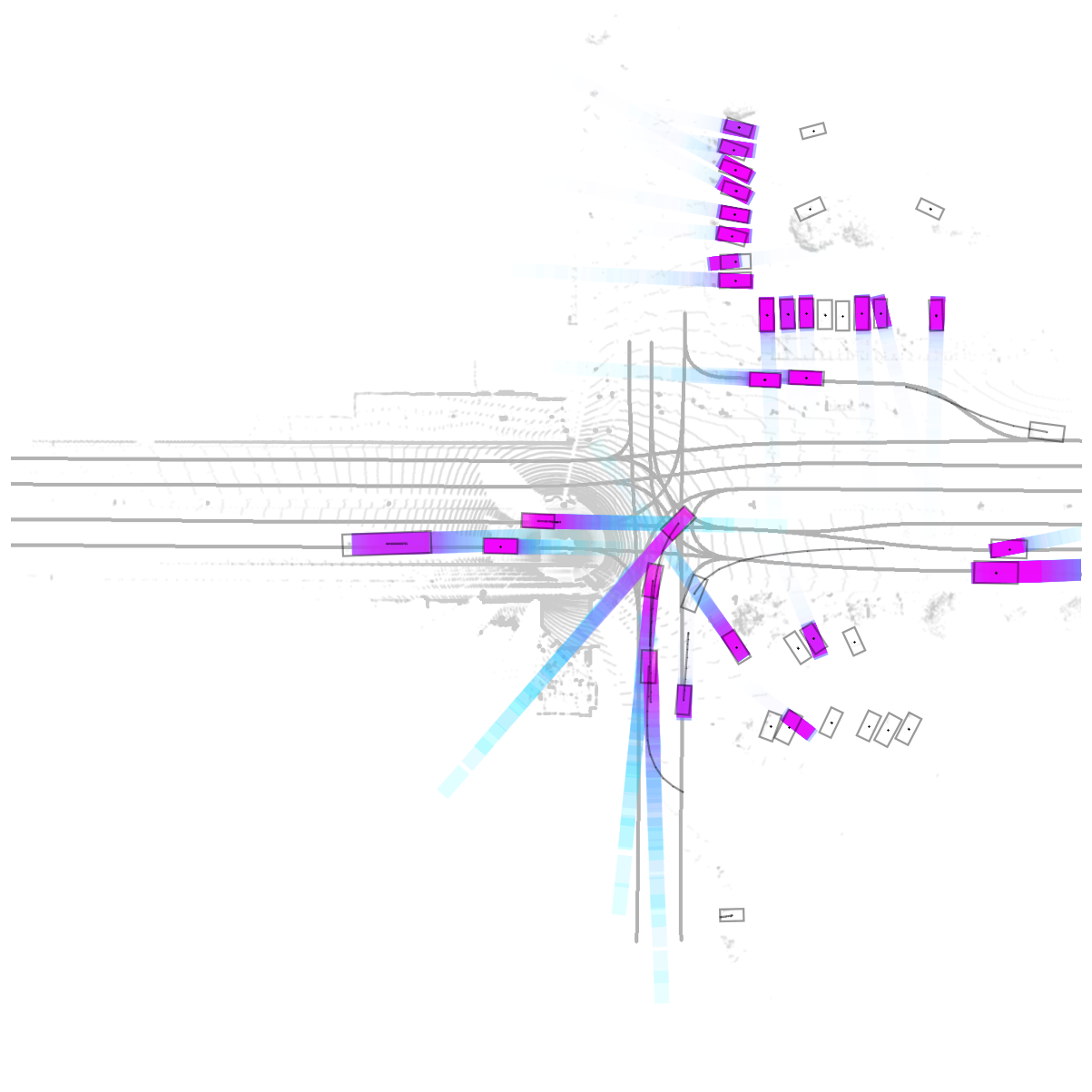}};
        \node[inner sep=0pt, outer sep=0, anchor=west]  (scene2mtp)       at (scene2st.east)        {\includegraphics[width=\imw]{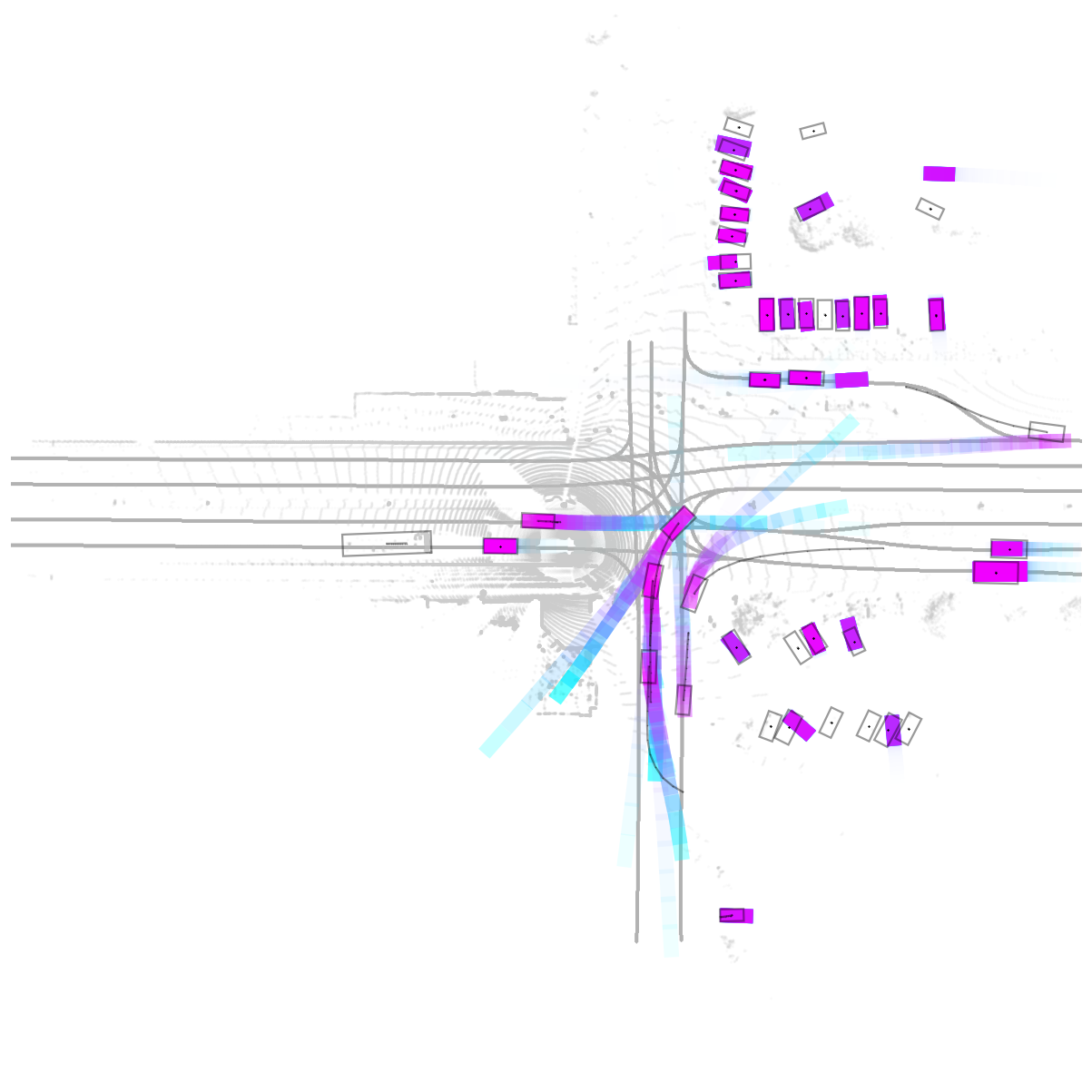}};
        \node[inner sep=0pt, outer sep=0, anchor=west]  (scene2gorela) at (scene2mtp.east)          {\includegraphics[width=\imw]{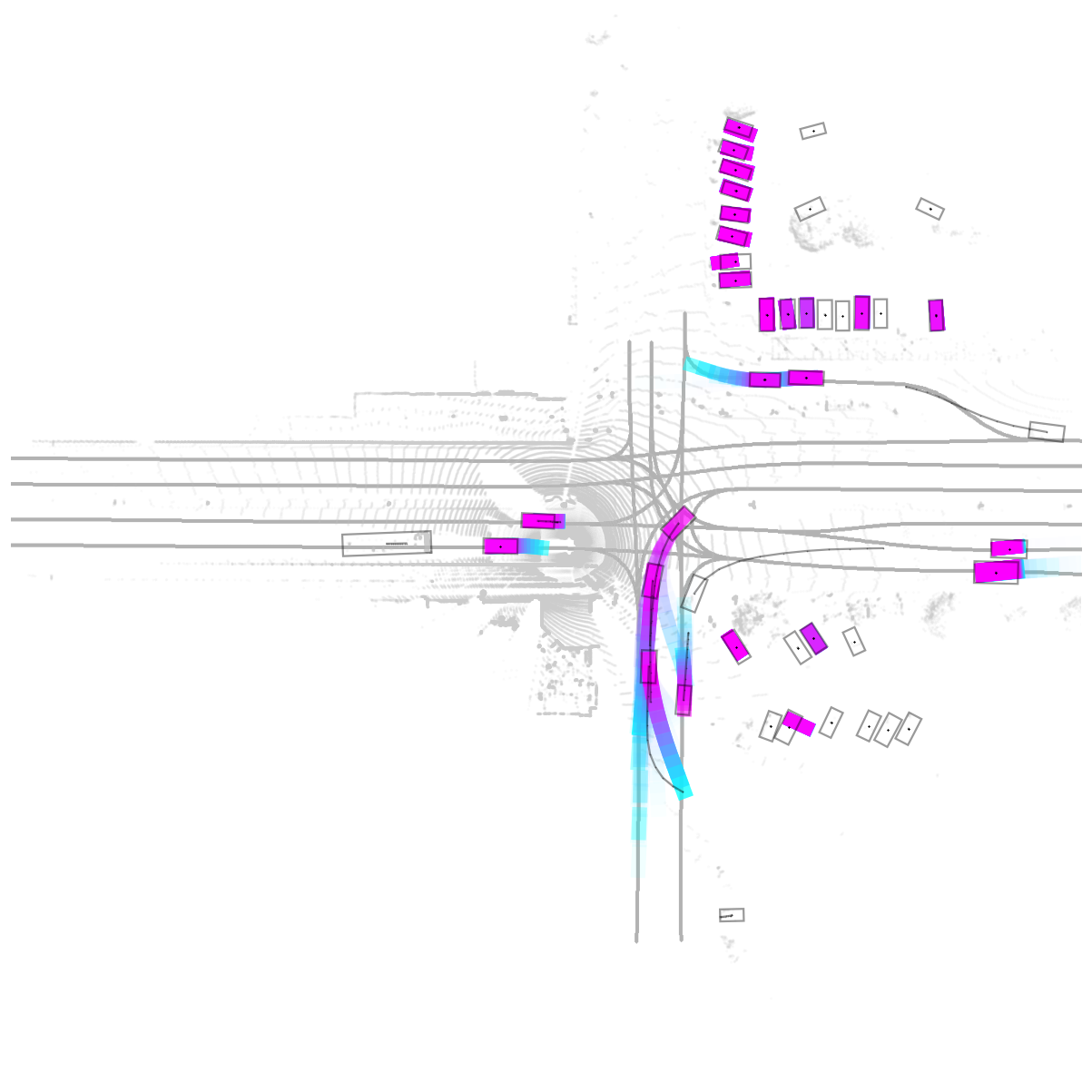}};
        \node[inner sep=0pt, outer sep=0, anchor=west]  (scene2detra)     at (scene2gorela.east)    {\includegraphics[width=\imw]{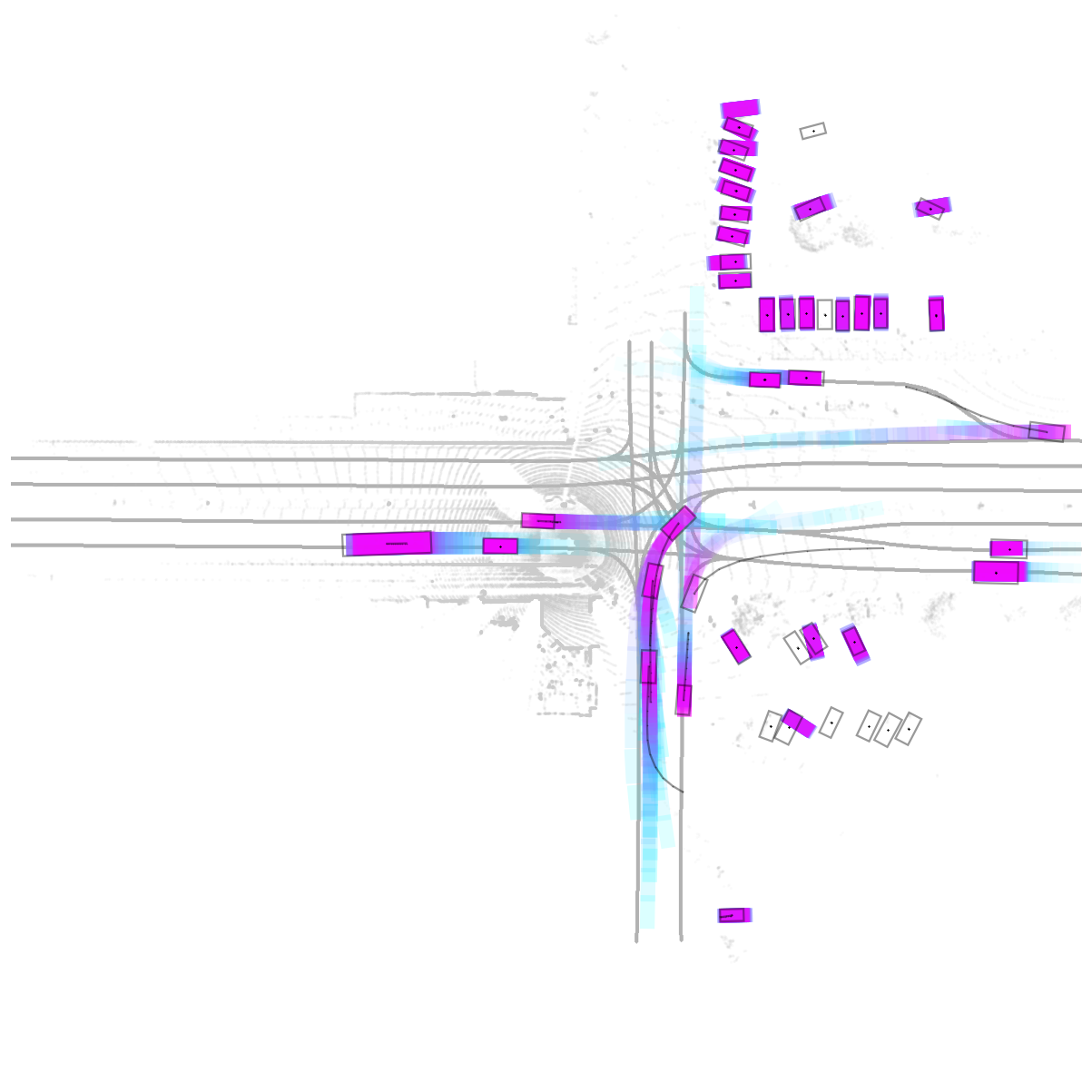}};

        \node[inner sep=0pt, outer sep=0, anchor=north] (scene3st)        at (scene2st.south)       {\includegraphics[width=\imw]{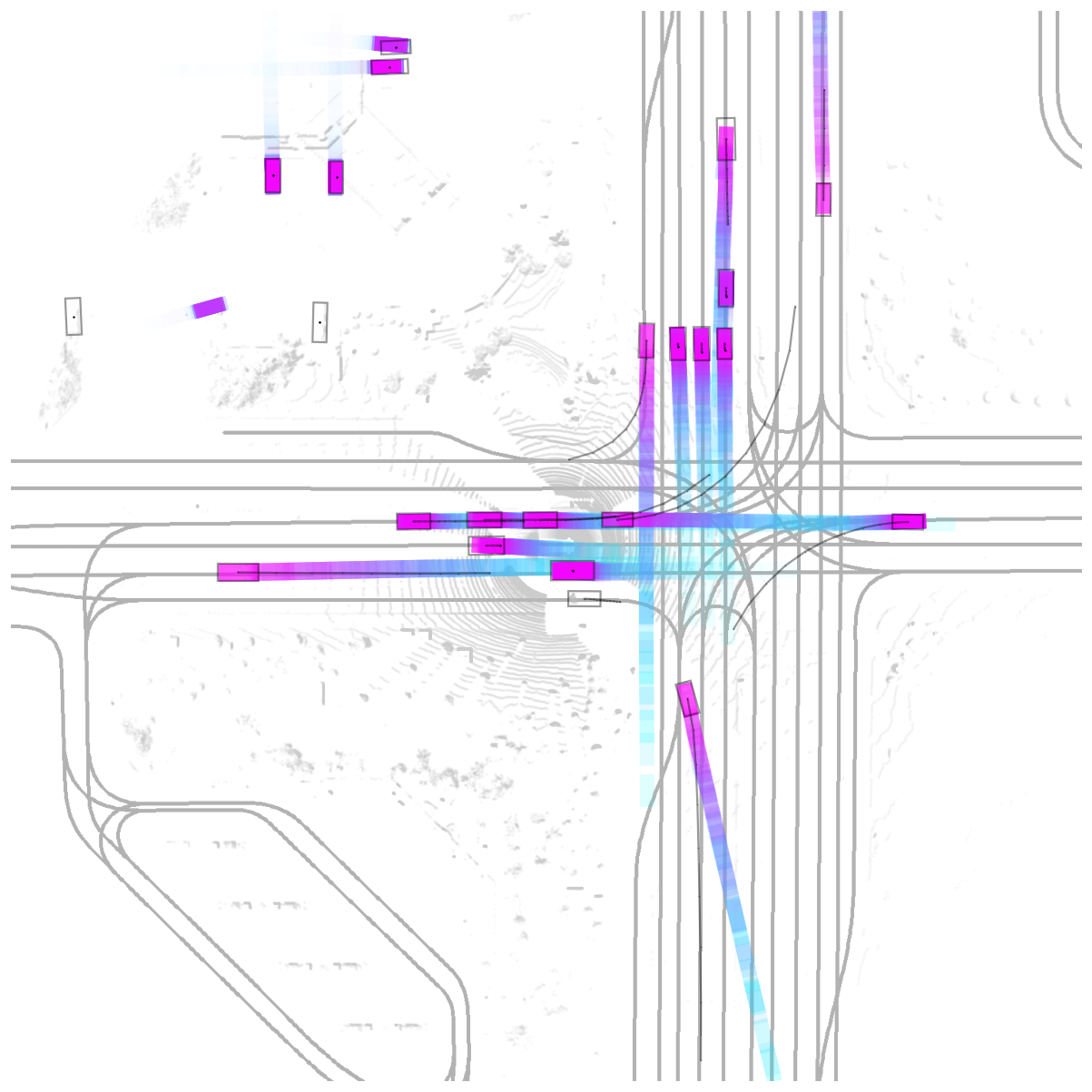}};
        \node[inner sep=0pt, outer sep=0, anchor=west]  (scene3mtp)       at (scene3st.east)        {\includegraphics[width=\imw]{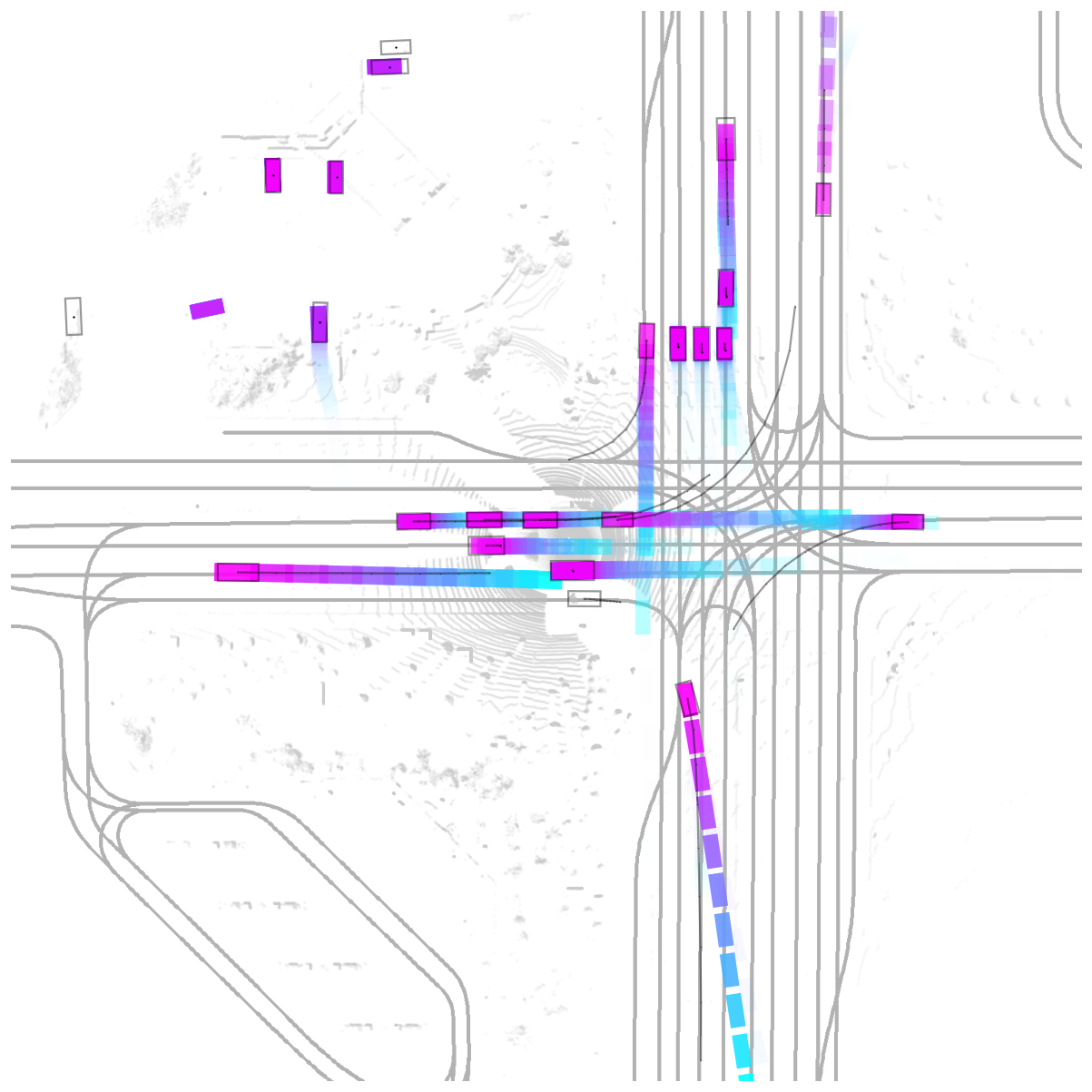}};
        \node[inner sep=0pt, outer sep=0, anchor=west]  (scene3gorela) at (scene3mtp.east)          {\includegraphics[width=\imw]{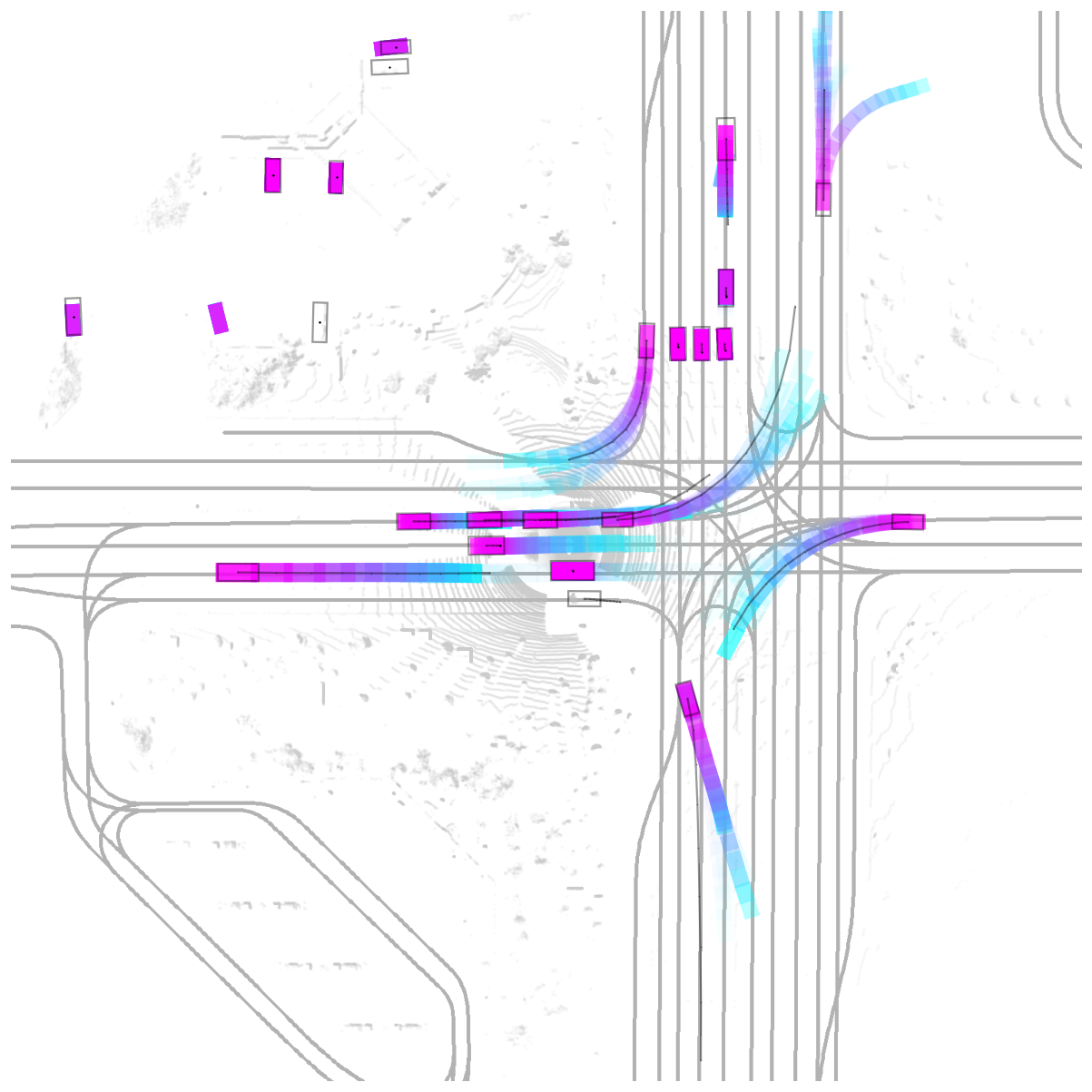}};
        \node[inner sep=0pt, outer sep=0, anchor=west]  (scene3detra)     at (scene3gorela.east)    {\includegraphics[width=\imw]{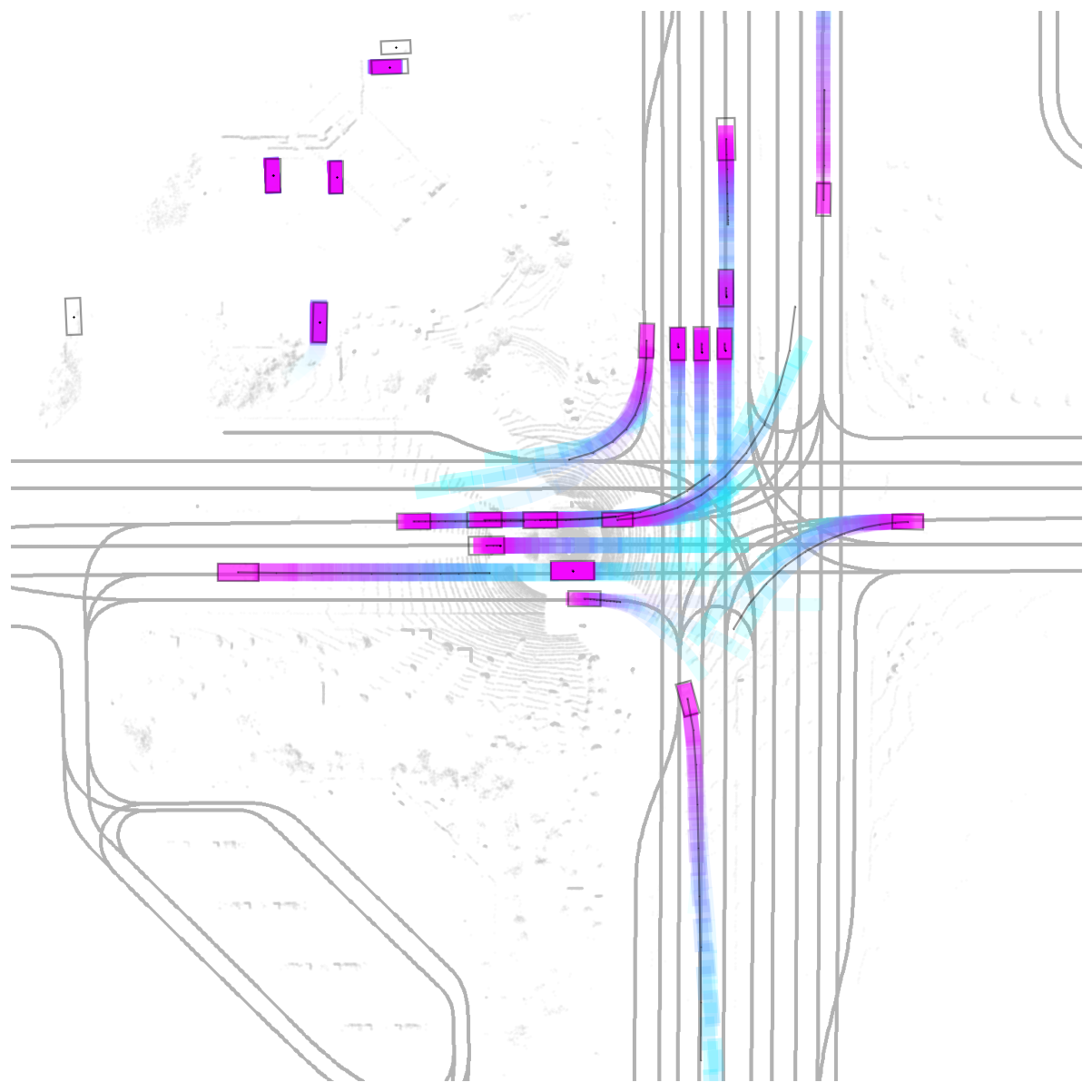}};

        \node[inner sep=0pt, outer sep=0, anchor=north] (scene4st)        at (scene3st.south)       {\includegraphics[width=\imw]{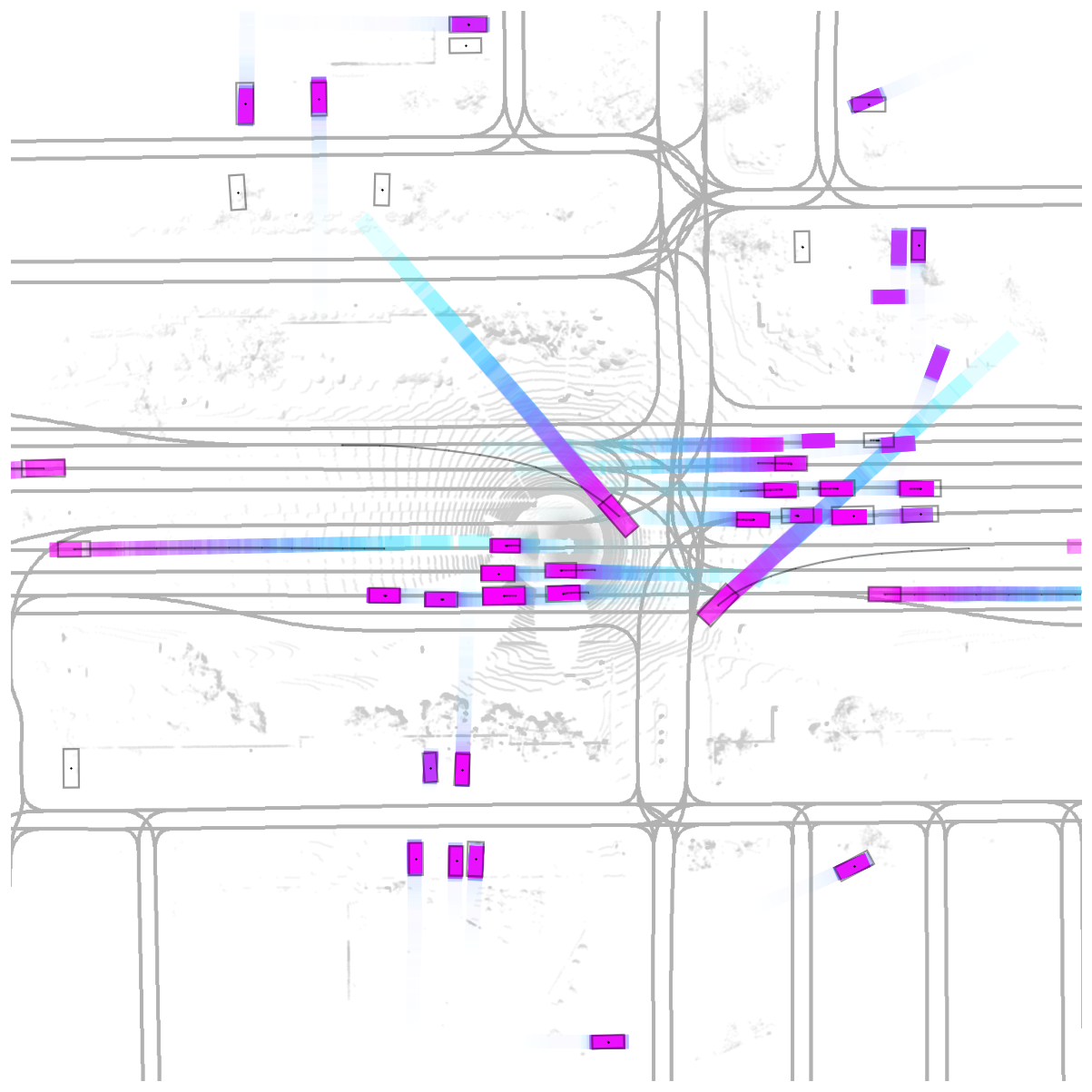}};
        \node[inner sep=0pt, outer sep=0, anchor=west]  (scene4mtp)       at (scene4st.east)        {\includegraphics[width=\imw]{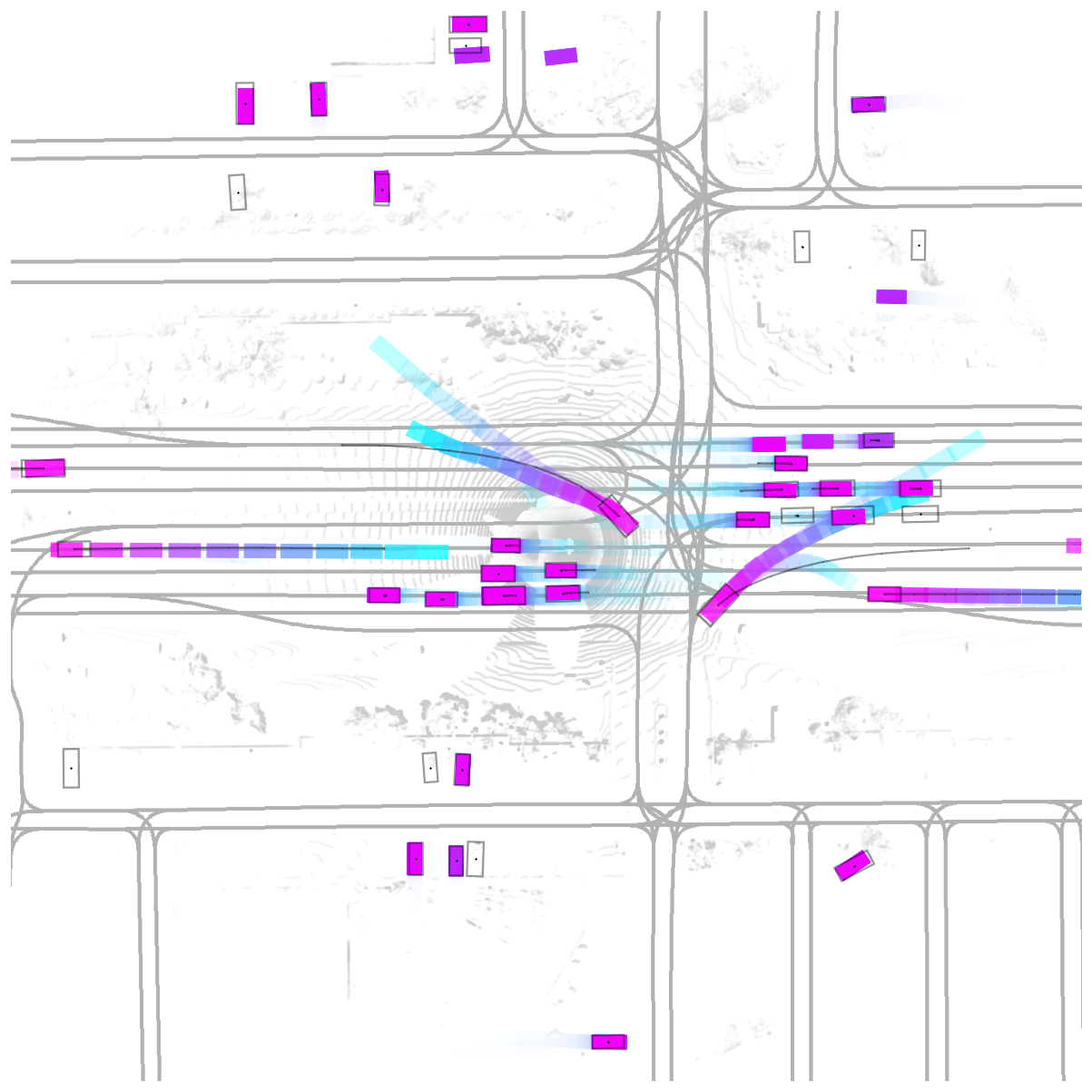}};
        \node[inner sep=0pt, outer sep=0, anchor=west]  (scene4gorela)    at (scene4mtp.east)          {\includegraphics[width=\imw]{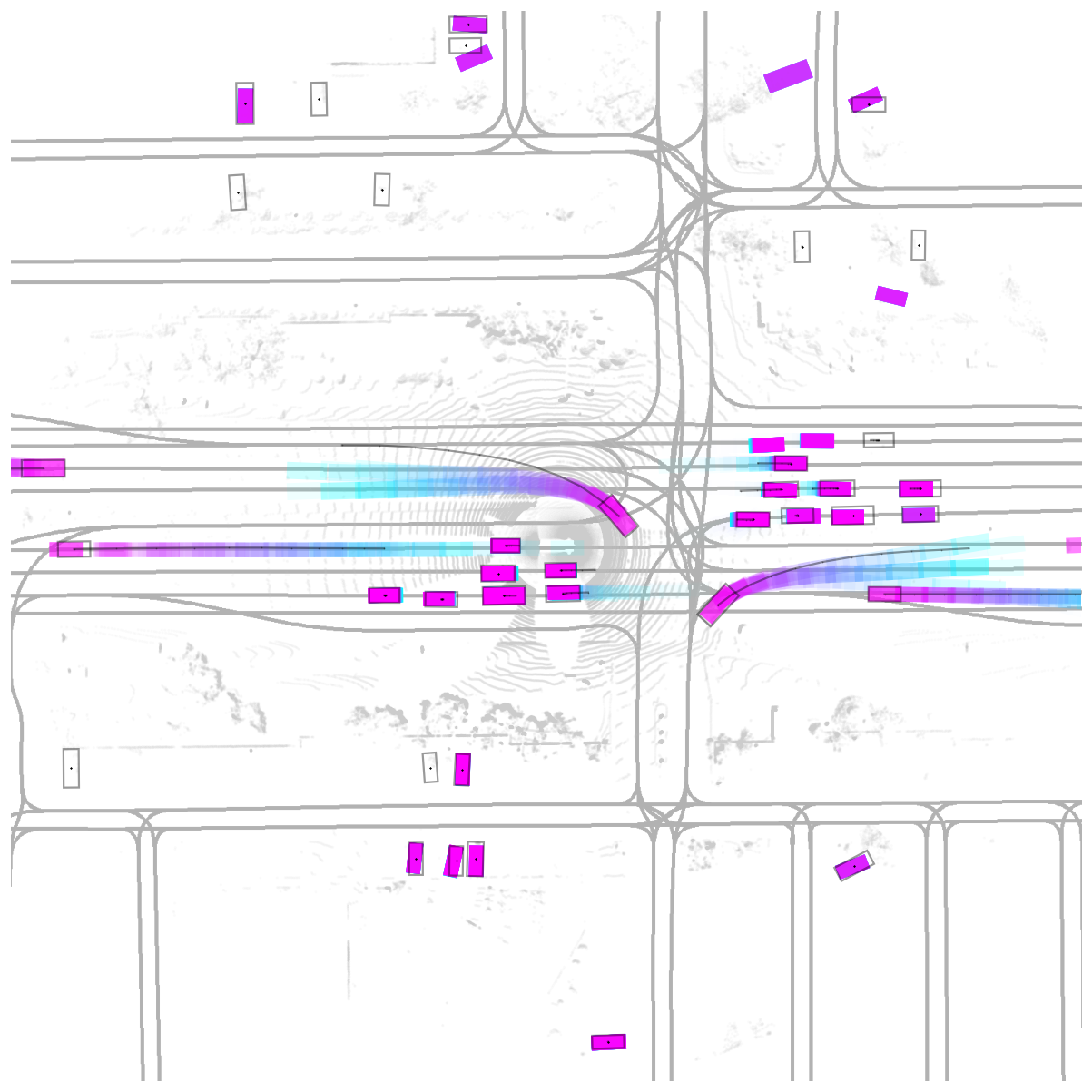}};
        \node[inner sep=0pt, outer sep=0, anchor=west]  (scene4detra)     at (scene4gorela.east)    {\includegraphics[width=\imw]{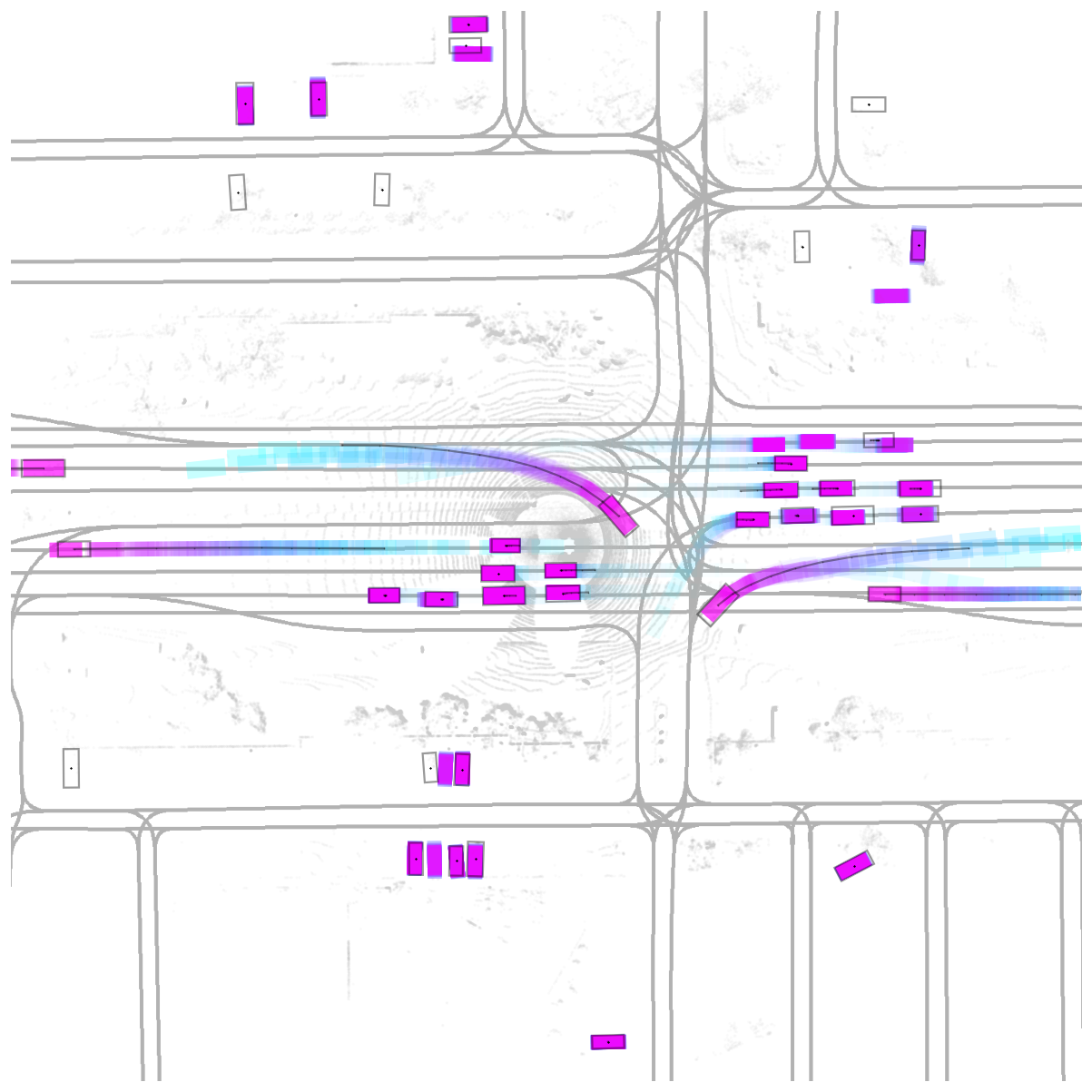}};

    \end{tikzpicture}
    \caption{
        Qualitative results on WOD. 
    }
    \label{fig:baseline-qualitative-wod}    
\end{figure*}

\Cref{fig:baseline-qualitative-wod} shows qualitative results on WOD for \ourmodel{}
and the baselines. Similarly to the results on AV2 shown in the main paper,
\ourmodel{} exhibits more accurate trajectory forecasts and detections than the baselines.

\begin{figure*}[t]
    \centering
    \begin{tikzpicture}
        \pgfmathsetlengthmacro{\imw}{0.25\textwidth}
        \pgfmathsetlengthmacro{\bw}{1.0pt}

        \node[inner sep=0pt, outer sep=0, anchor=west]  (scene1lvl0)    at (0, 0)                {\includegraphics[width=\imw]{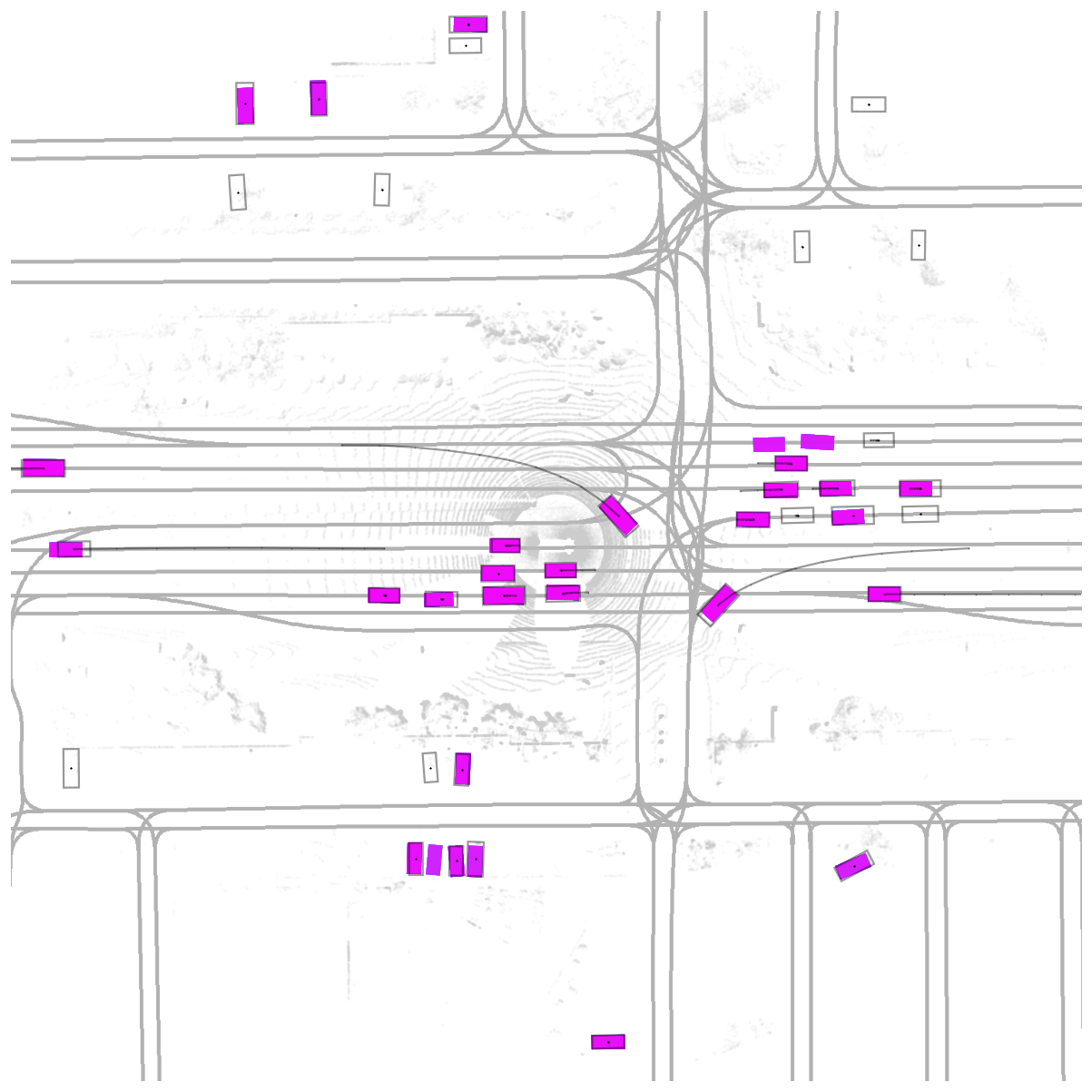}};
        \node[inner sep=0pt, outer sep=0, anchor=west]  (scene1lvl1)    at (scene1lvl0.east)     {\includegraphics[width=\imw]{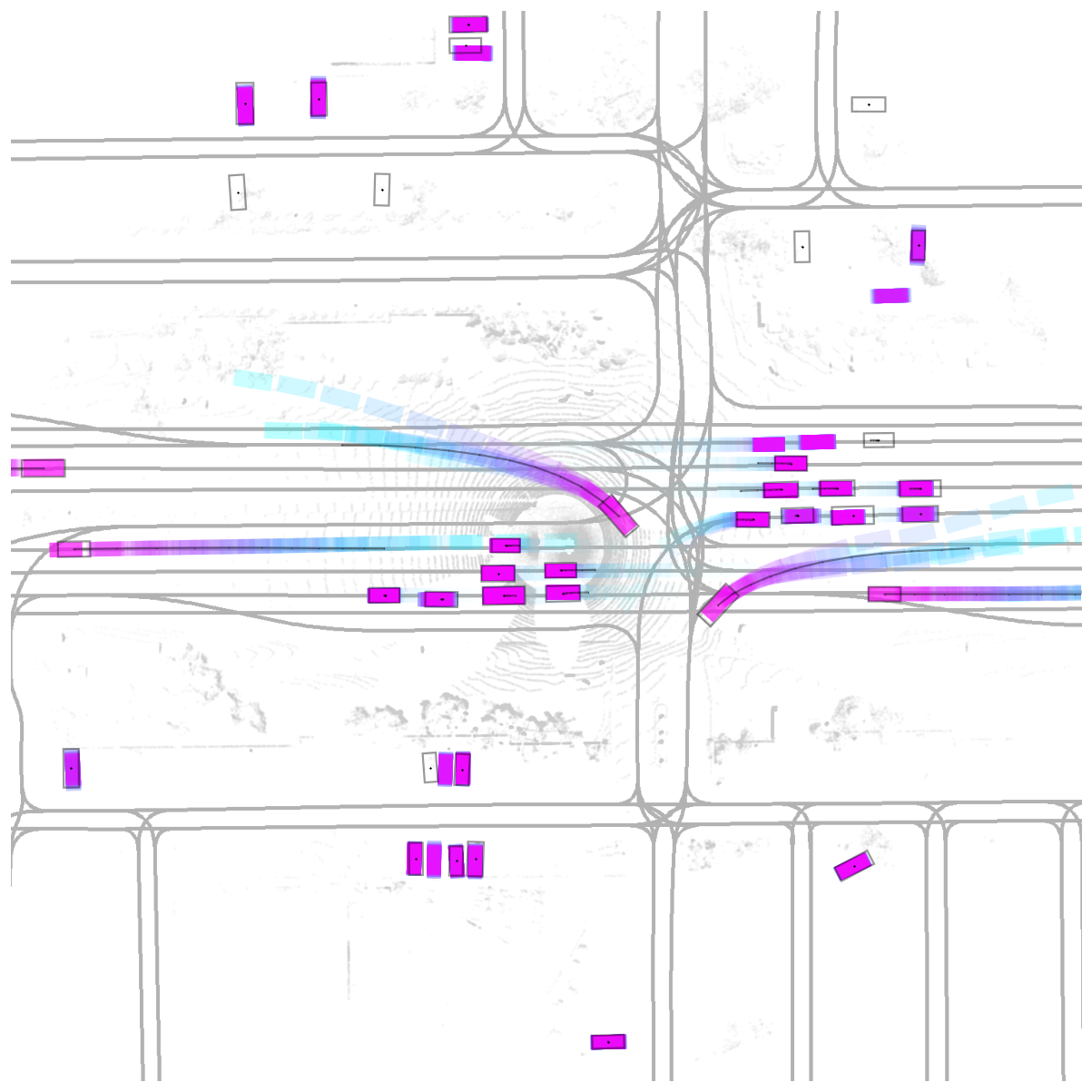}};
        \node[inner sep=0pt, outer sep=0, anchor=west]  (scene1lvl2)    at (scene1lvl1.east)     {\includegraphics[width=\imw]{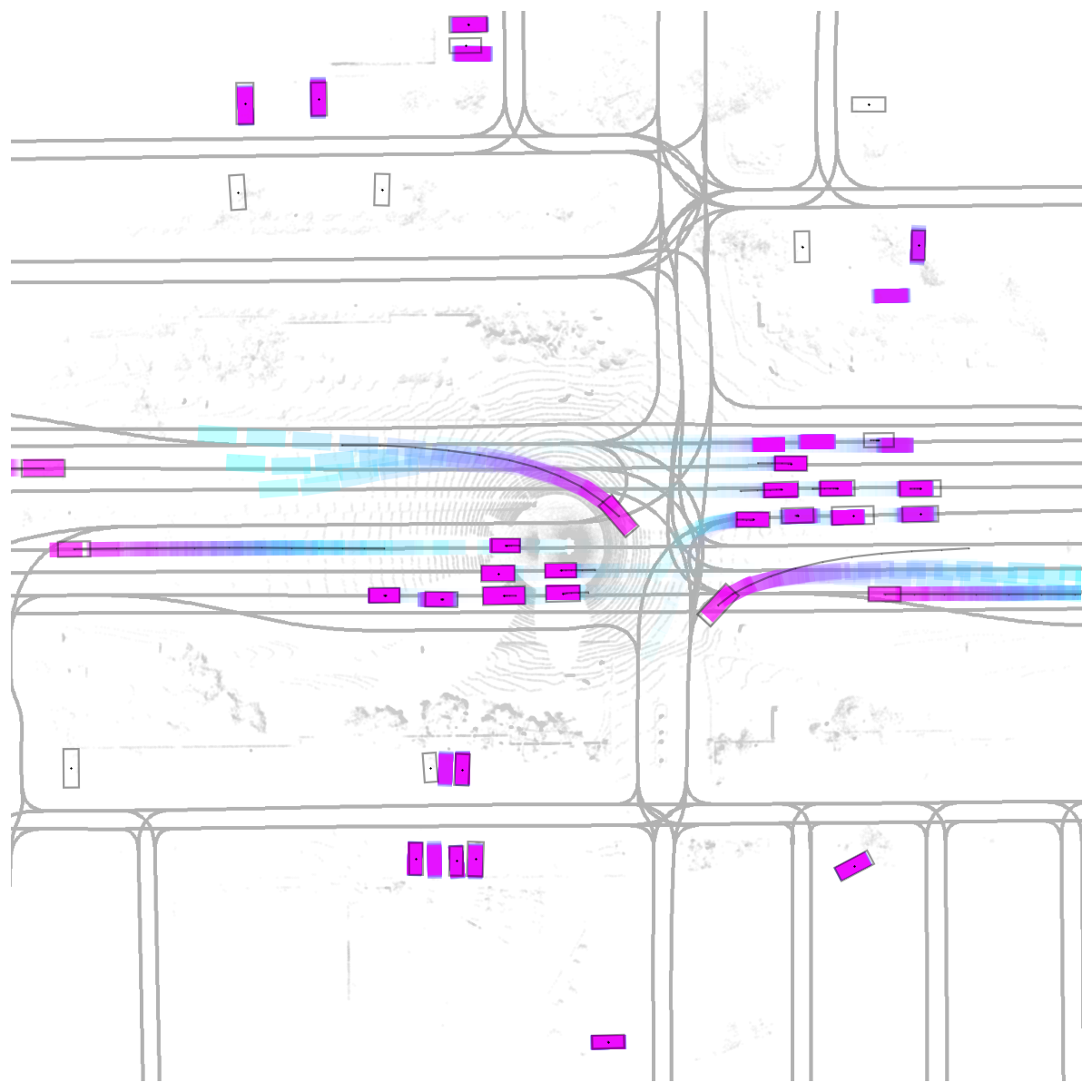}};
        \node[inner sep=0pt, outer sep=0, anchor=west]  (scene1lvl3)    at (scene1lvl2.east)     {\includegraphics[width=\imw]{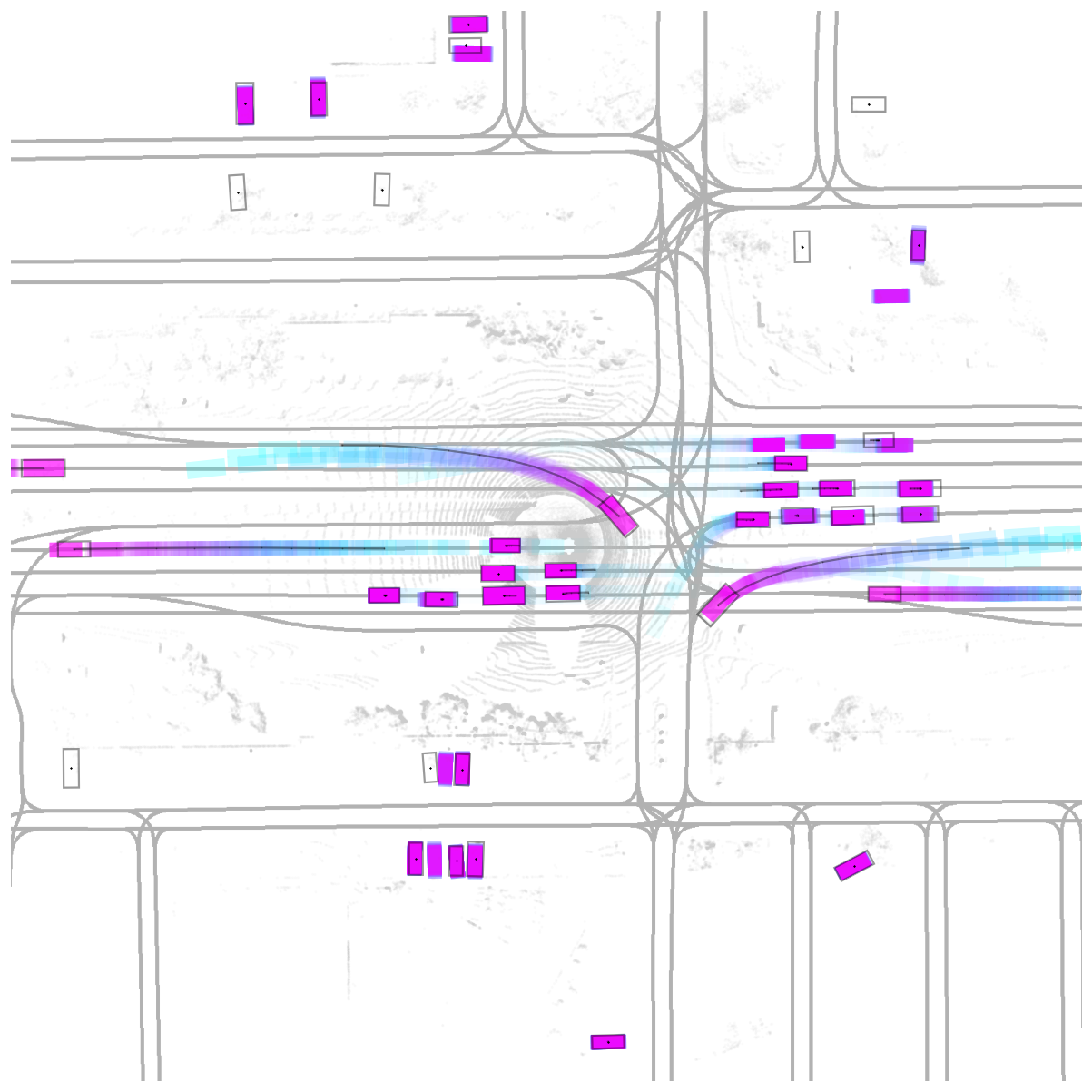}};
        \node[anchor=south] at (scene1lvl0.north) {\ourmodel{} $i=0$};
        \node[anchor=south] at (scene1lvl1.north) {\ourmodel{} $i=1$};
        \node[anchor=south] at (scene1lvl2.north) {\ourmodel{} $i=2$};
        \node[anchor=south] at (scene1lvl3.north) {\ourmodel{} $i=3$};

        \node[inner sep=0pt, outer sep=0, anchor=north]  (scene2lvl0)    at (scene1lvl0.south)    {\includegraphics[width=\imw]{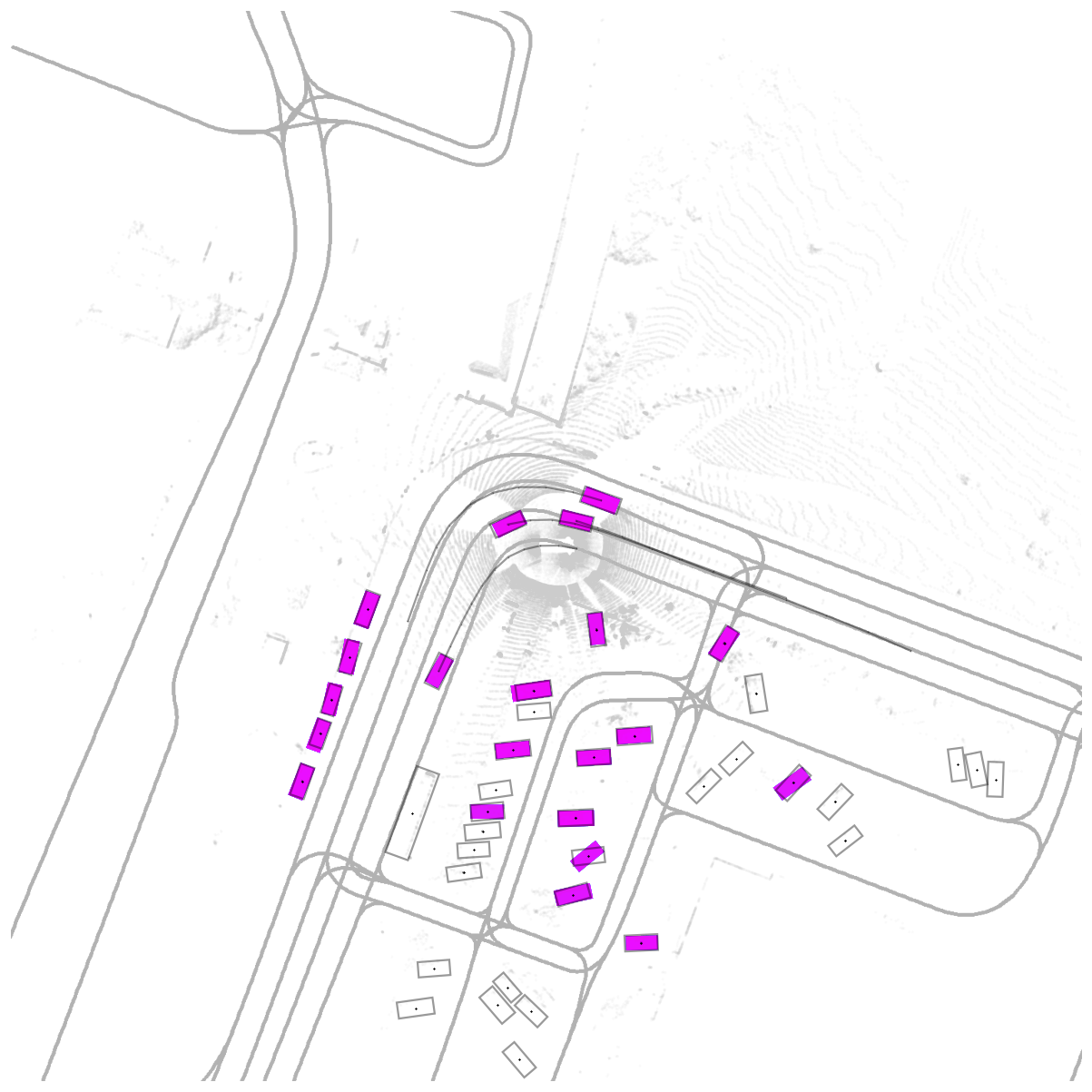}};
        \node[inner sep=0pt, outer sep=0, anchor=west]   (scene2lvl1)    at (scene2lvl0.east)     {\includegraphics[width=\imw]{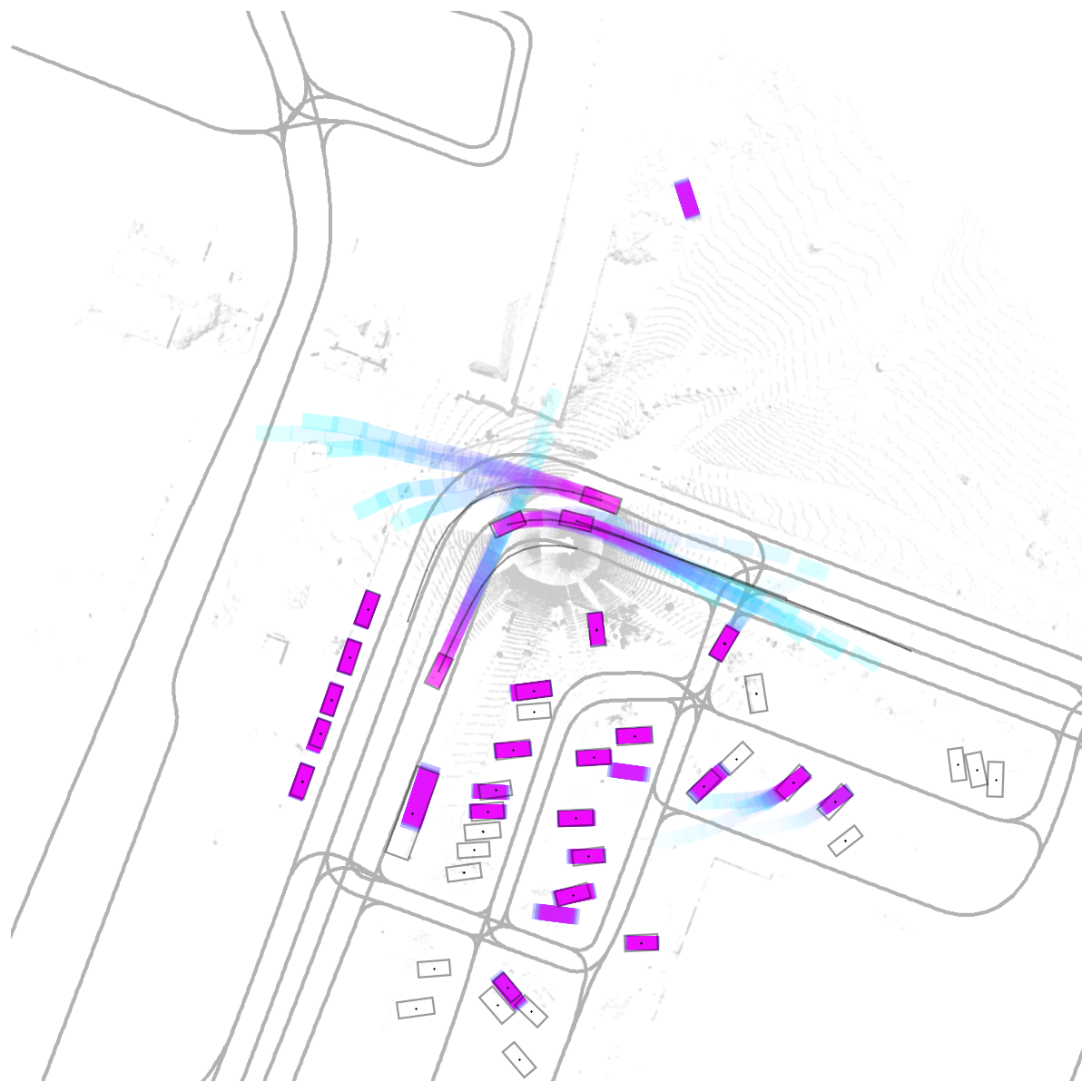}};
        \node[inner sep=0pt, outer sep=0, anchor=west]   (scene2lvl2)    at (scene2lvl1.east)     {\includegraphics[width=\imw]{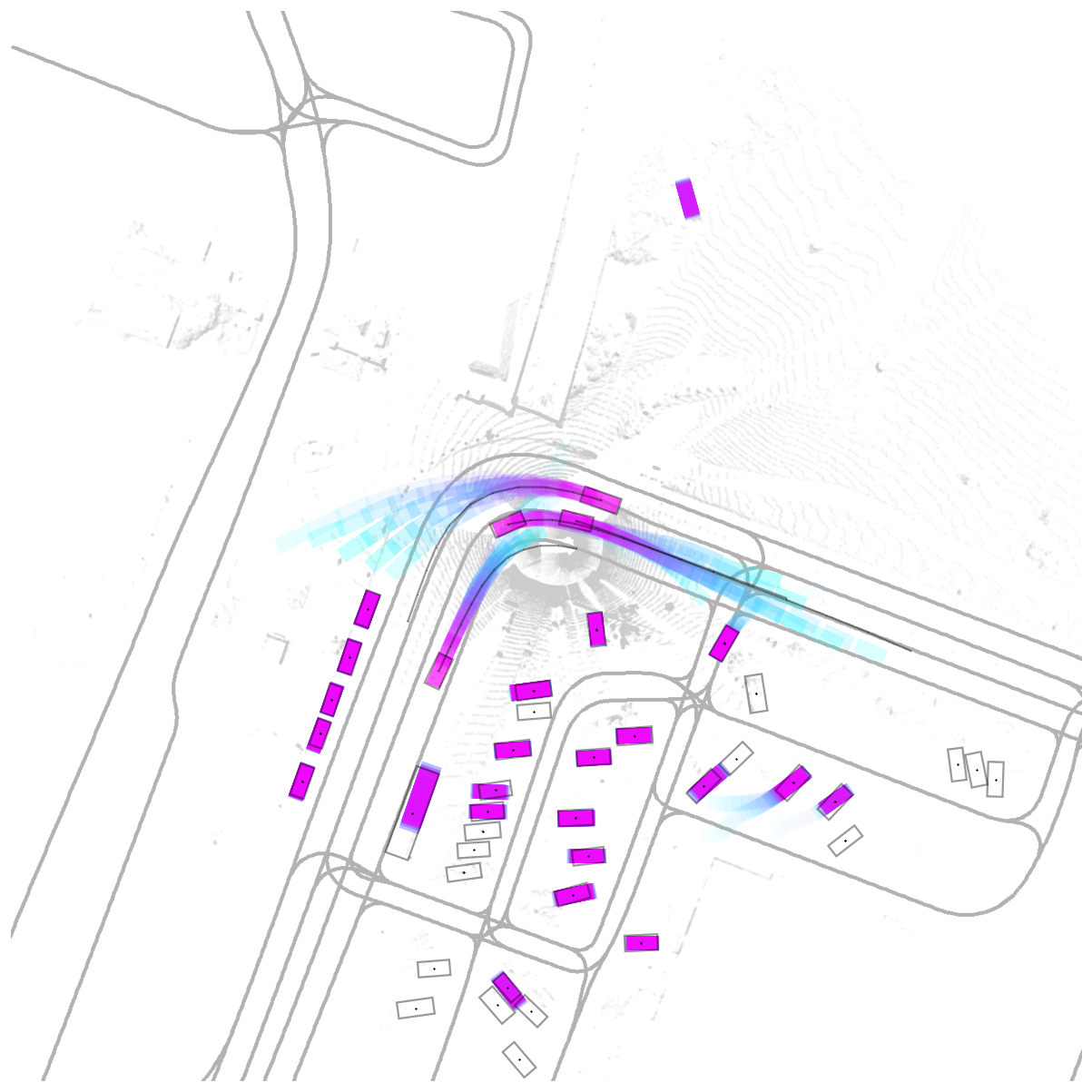}};
        \node[inner sep=0pt, outer sep=0, anchor=west]   (scene2lvl3)    at (scene2lvl2.east)     {\includegraphics[width=\imw]{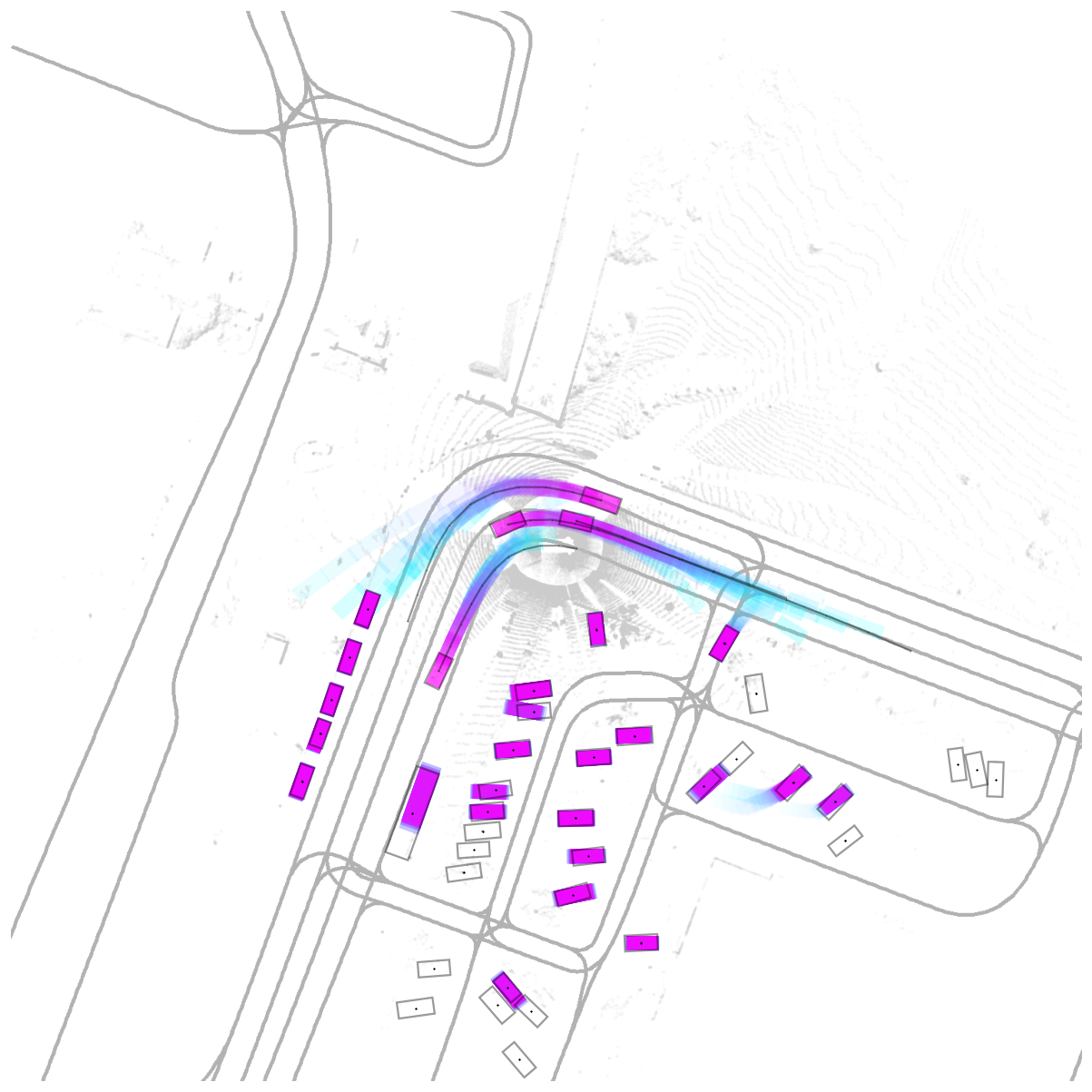}};

        \node[inner sep=0pt, outer sep=0, anchor=north]  (scene3lvl0)    at (scene2lvl0.south)    {\includegraphics[width=\imw]{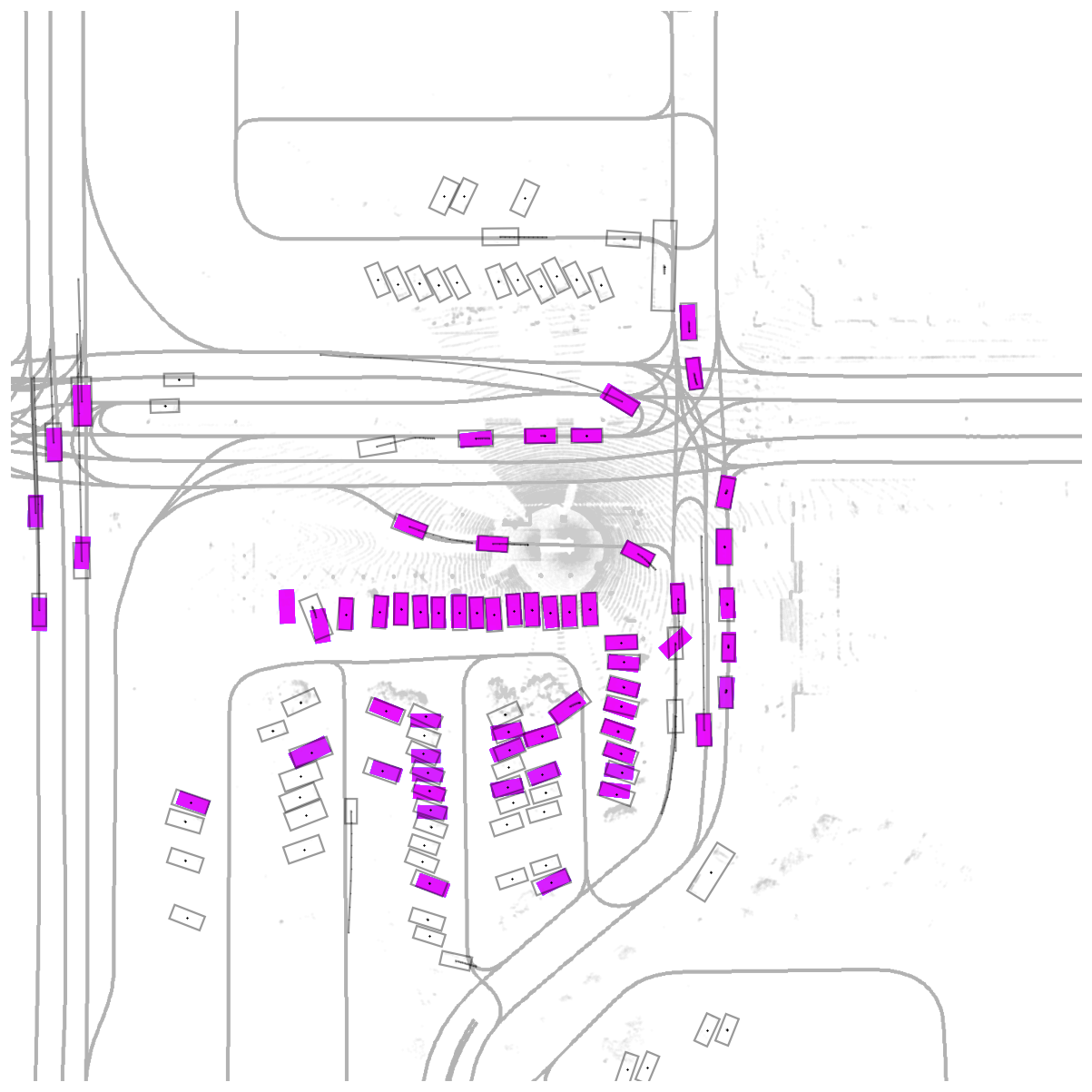}};
        \node[inner sep=0pt, outer sep=0, anchor=west]   (scene3lvl1)    at (scene3lvl0.east)     {\includegraphics[width=\imw]{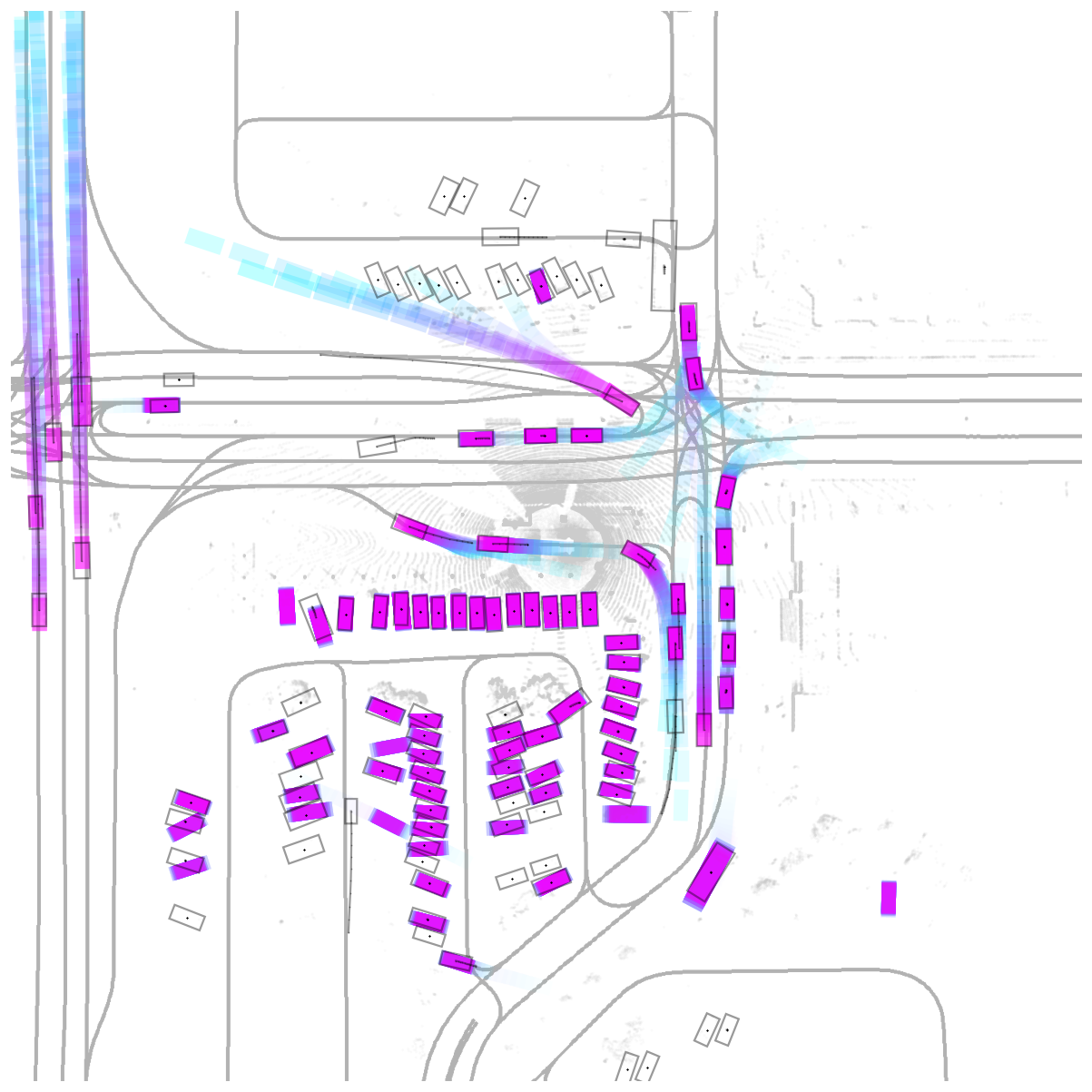}};
        \node[inner sep=0pt, outer sep=0, anchor=west]   (scene3lvl2)    at (scene3lvl1.east)     {\includegraphics[width=\imw]{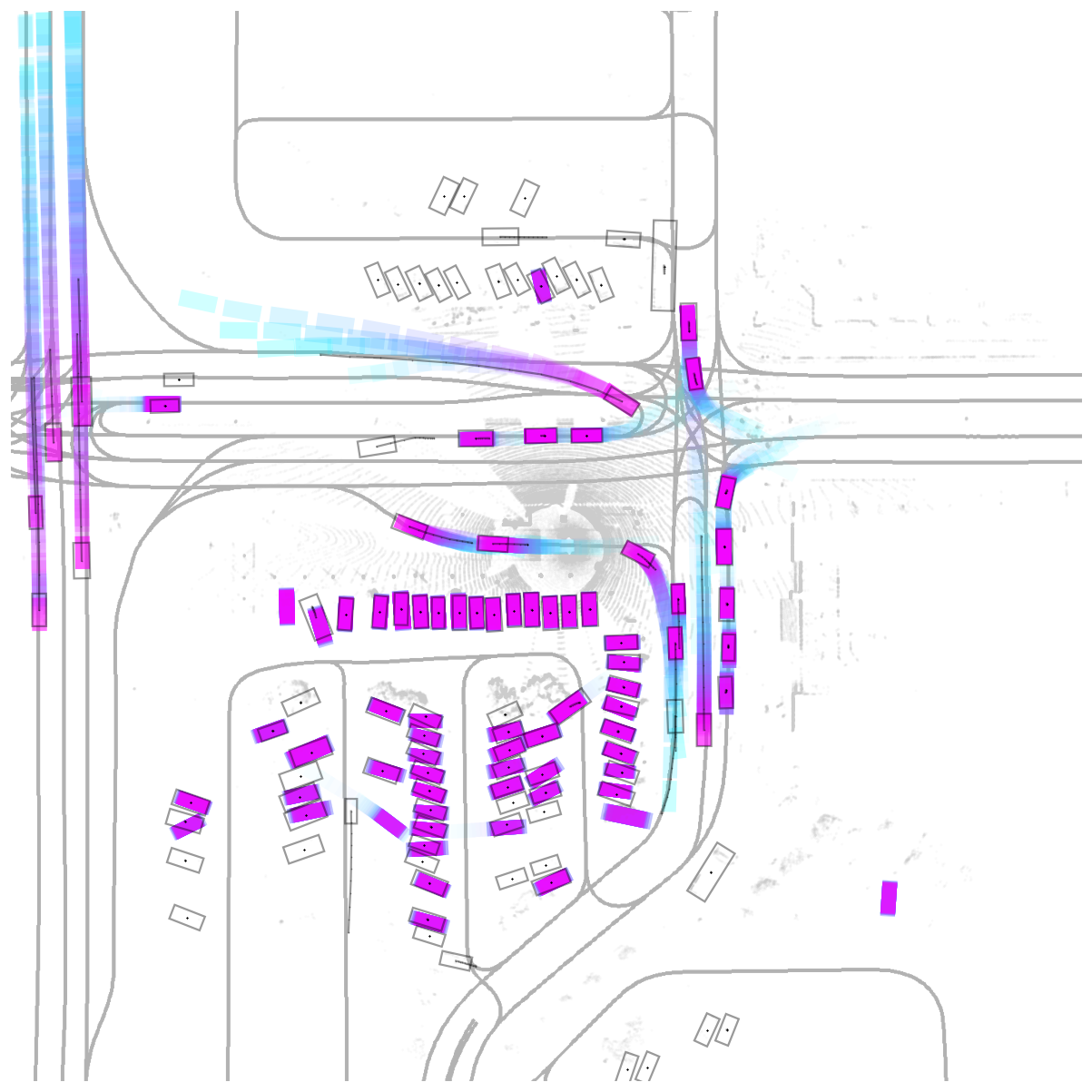}};
        \node[inner sep=0pt, outer sep=0, anchor=west]   (scene3lvl3)    at (scene3lvl2.east)     {\includegraphics[width=\imw]{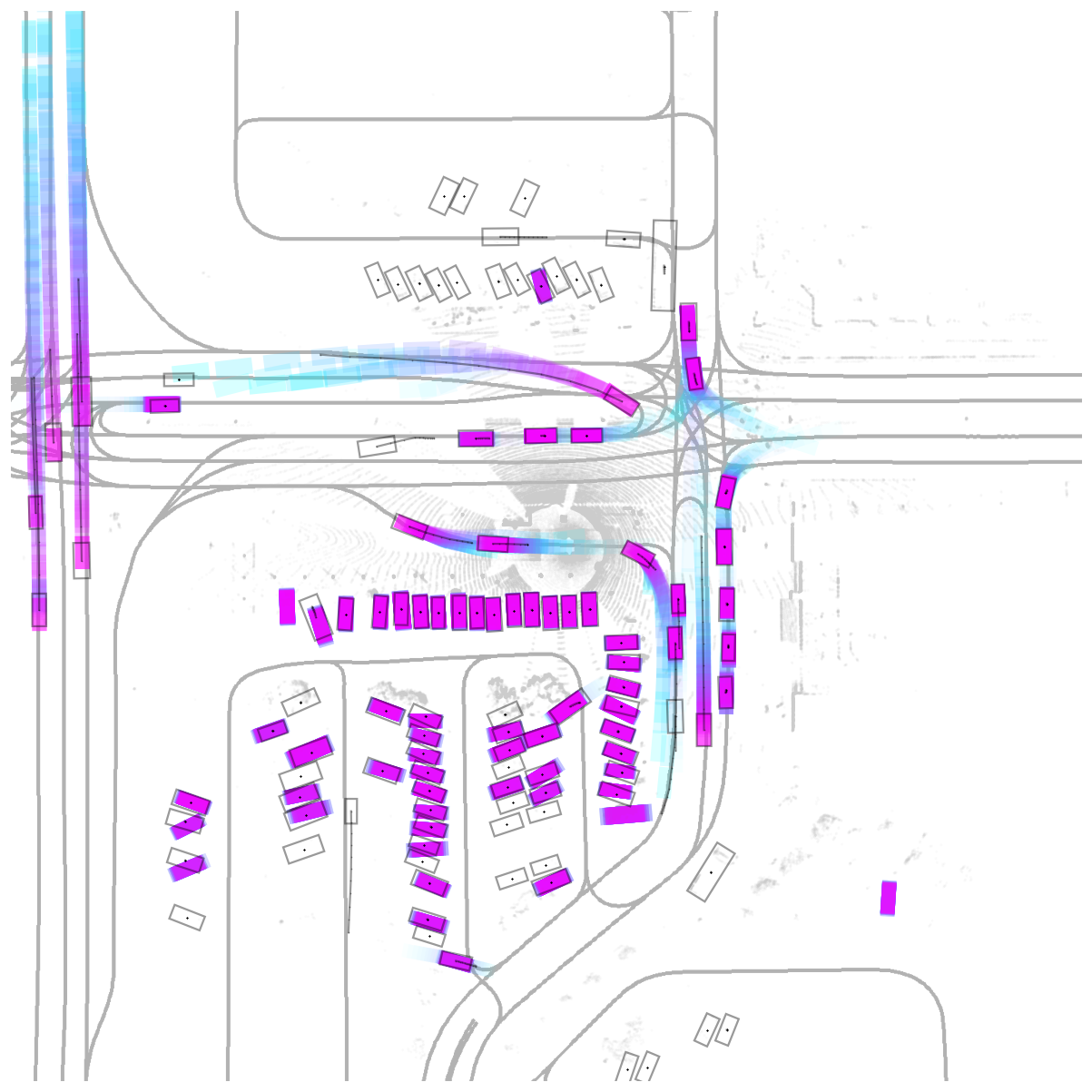}};

        \node[inner sep=0pt, outer sep=0, anchor=north]  (scene4lvl0)    at (scene3lvl0.south)    {\includegraphics[width=\imw]{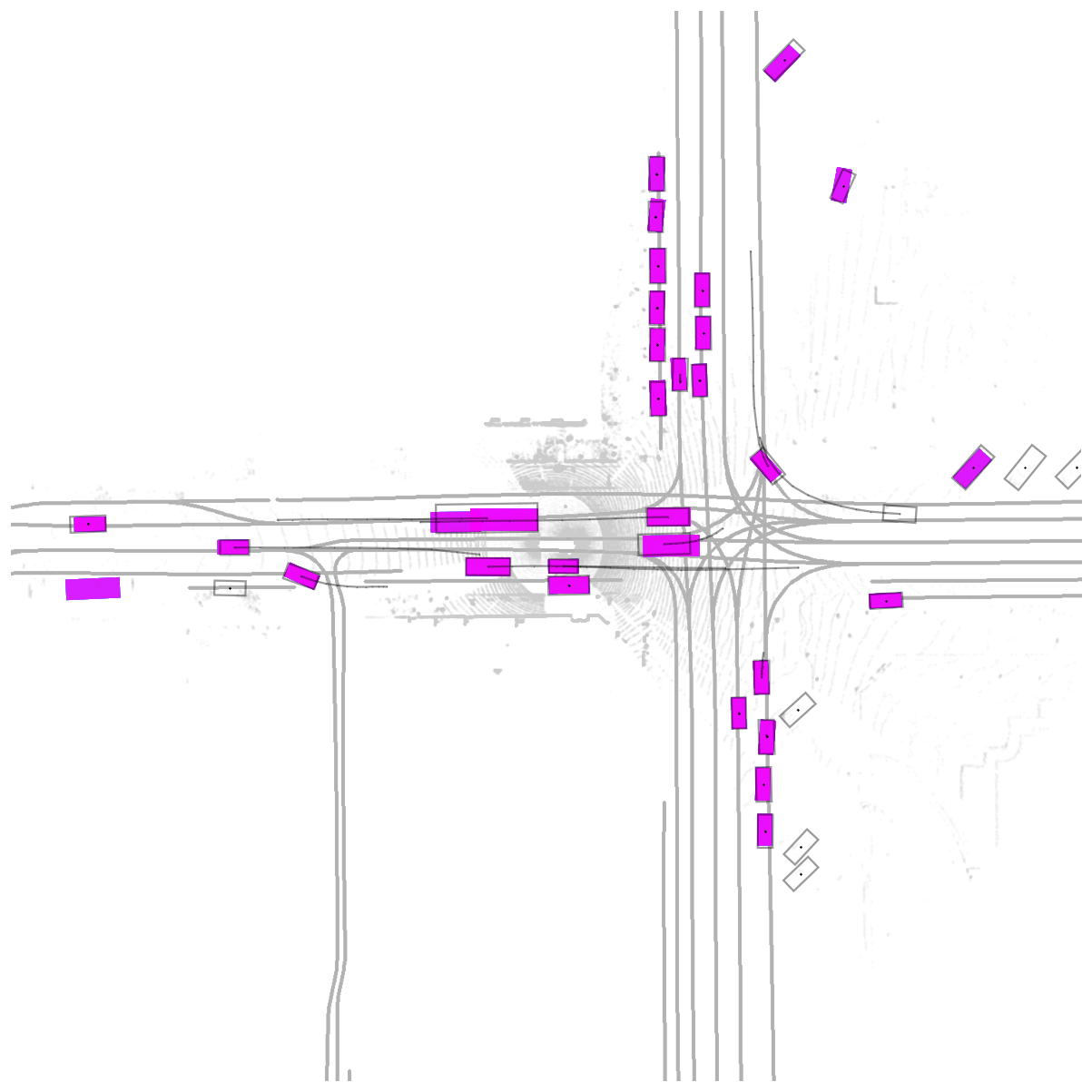}};
        \node[inner sep=0pt, outer sep=0, anchor=west]   (scene4lvl1)    at (scene4lvl0.east)     {\includegraphics[width=\imw]{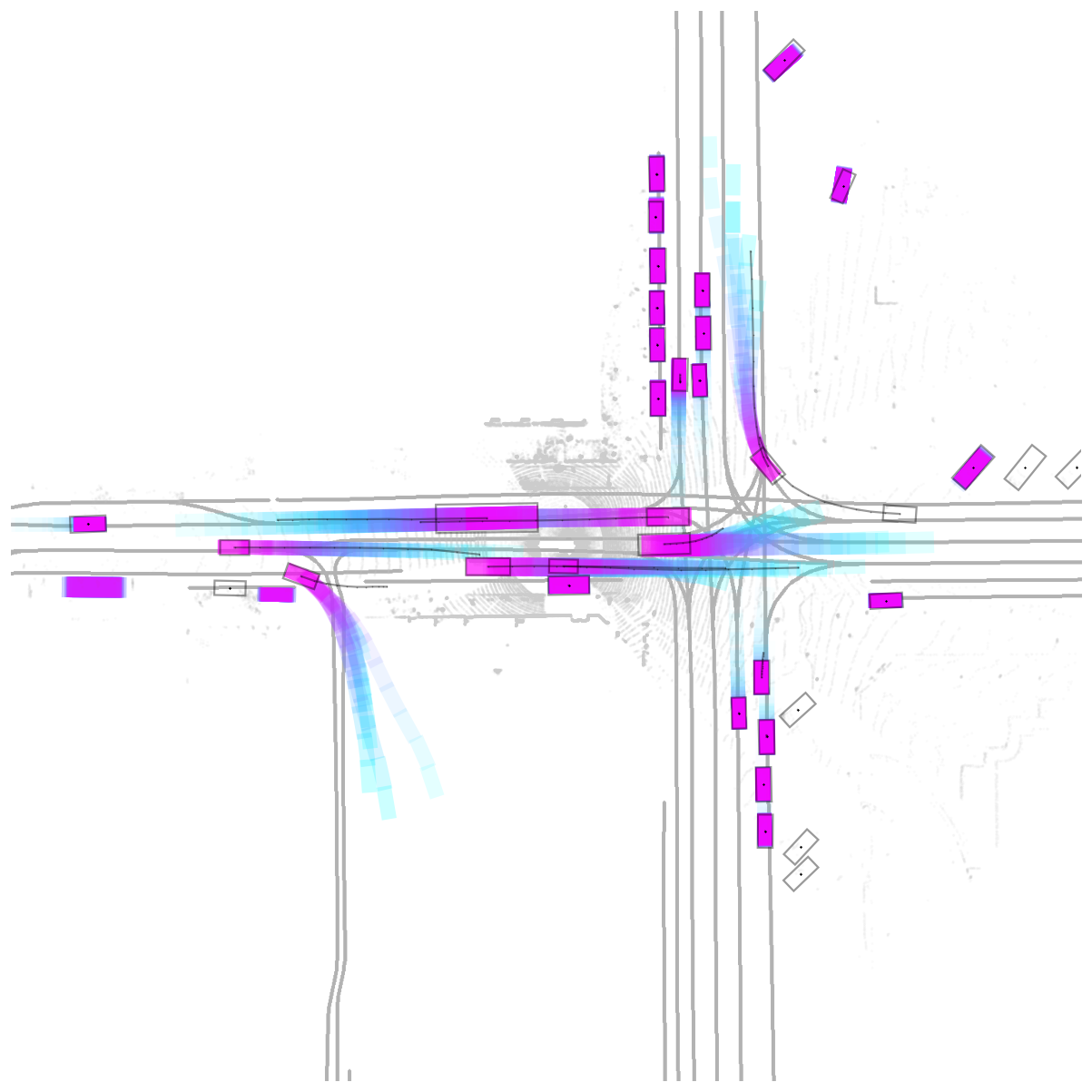}};
        \node[inner sep=0pt, outer sep=0, anchor=west]   (scene4lvl2)    at (scene4lvl1.east)     {\includegraphics[width=\imw]{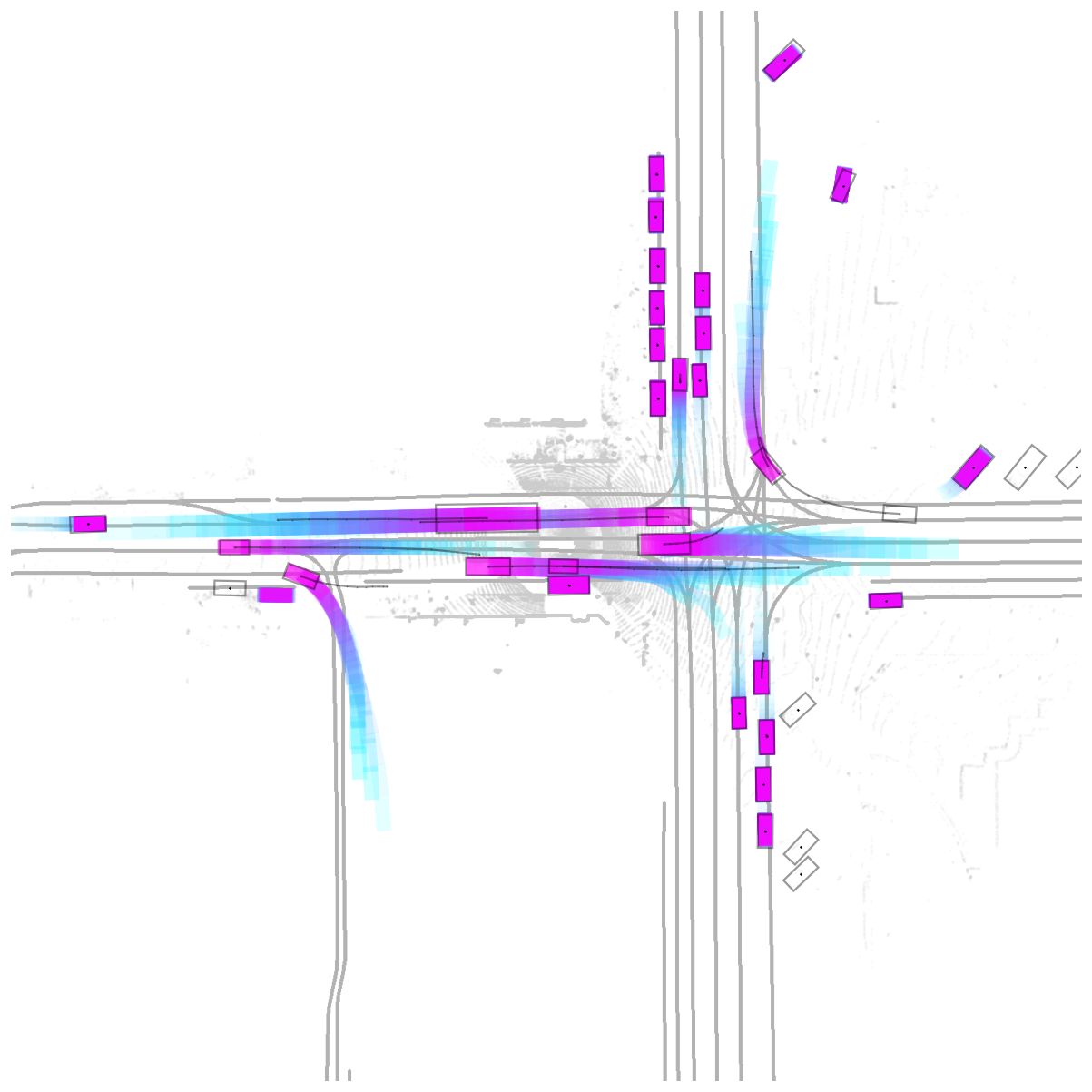}};
        \node[inner sep=0pt, outer sep=0, anchor=west]   (scene4lvl3)    at (scene4lvl2.east)     {\includegraphics[width=\imw]{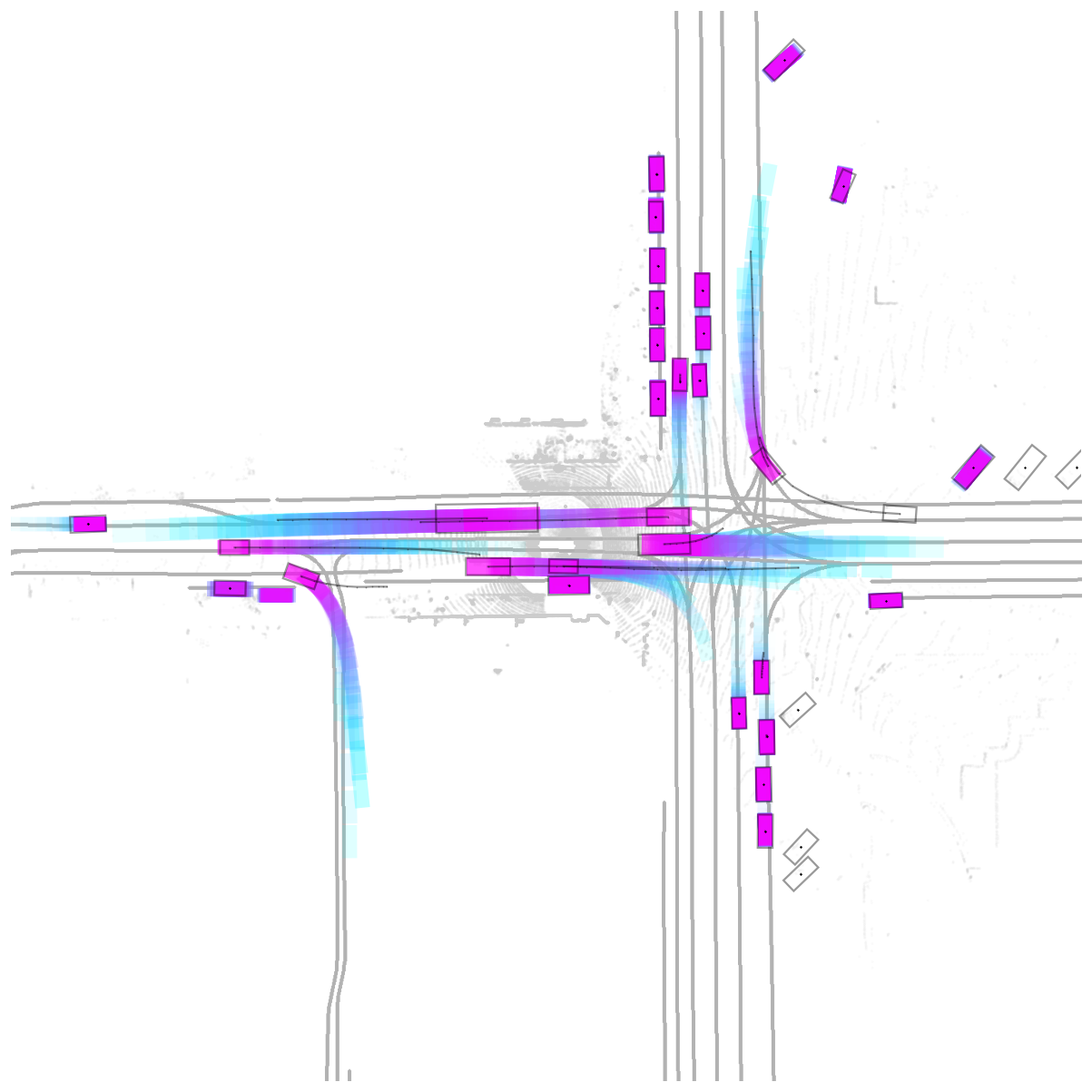}};

    \end{tikzpicture}
    \caption{Visualizing the \ourmodel{}'s self improvement over refinement blocks on WOD.}
    \label{fig:intermediate-qualitative-wod}
\end{figure*}

\Cref{fig:intermediate-qualitative-wod} shows the self-improvement of \ourmodel{}
over transformer refinement blocks on WOD. Again, we see that the trajectories follow
the map more accurately over transformer refinement blocks, and the detections become more accurate, with limited gains between
the second-last and final refinement blocks.

\section{Limitations and Future Work} \label{sec:limitations}
While our work provides a simpler approach to object detection and trajectory forecasting based on trajectory refinement that achieves better results than previous paradigms, it has several limitations.

\ourmodel{} does not consider uncertainty on its detections beyond the confidence score. This design follows the standard in detection literature, but there are likely better methods than this. Instead, future works could consider multiple detection hypotheses similar to how multiple modes are considered for motion forecasting. Our method is very suitable for this since we already have a query and pose volume with a mode dimension, but right now, all the poses at $t=0$ are updated to be the same. The main challenge to extend detection to multi-hypothesis remains in developing strong objective functions for training and metrics for evaluation.

Another limitation is that in the current experiments, we did not consider the task of joint future forecasting across agents \cite{ilvm,scenetransformer}, only marginal forecasting for each agent. This task would be a good benchmark to evaluate the object-to-object interaction understanding of \ourmodel{}, but it is out of scope for this paper.

Finally, we did not tackle the problem of temporal consistency of
detection and forecasts over multiple forward passes at consecutive frames. Traditional approaches track the detections and forecasting outputs, which could
be applied to \ourmodel{} \cite{pnpnet}. Future work may explore how to learn better temporal consistency in the \ourmodel{} framework.

Despite its limitations, we hope \ourmodel{} can lay a strong foundation for future work on end-to-end detection and forecasting.

\end{document}